\pdfoutput=1
\documentclass[lettersize,journal]{IEEEtran}
\usepackage{amsmath,amsfonts}
\usepackage{algorithmic}
\usepackage{array}
\usepackage[caption=false,font=normalsize,labelfont=sf,textfont=sf]{subfig}
\usepackage{textcomp}
\usepackage{stfloats}
\usepackage{url}
\usepackage{verbatim}
\usepackage{booktabs}
\usepackage{graphicx}
\usepackage[table]{xcolor}
\usepackage{hyperref}

\graphicspath{{./Figures/}} 
\def\BibTeX{{\rm B\kern-.05em{\sc i\kern-.025em b}\kern-.08em
    T\kern-.1667em\lower.7ex\hbox{E}\kern-.125emX}}
\usepackage{balance}

\newtheorem{theorem}{Theorem}
\newtheorem{definition}[theorem]{Definition}

\begin{document}
\title{\textbf{CCDSReFormer}: Traffic Flow Prediction with a Criss-Crossed Dual-Stream Enhanced Rectified Transformer Model}
\author{Zhiqi Shao, 
        Michael G.H. Bell, 
        Ze Wang, 
        D. Glenn Geers, 
        Xusheng Yao, 
        and Junbin Gao
        
\thanks{Zhiqi Shao and Junbin Gao are with the Business Analytics Discipline, The University of Sydney, Camperdown, NSW 2006, Australia. (e-mail: zhiqi.shao@sydney.edu.au; junbin.gao@sydney.edu.au)} 
\thanks{Michael G.H. Bell, Ze Wang, and D. Glenn Geers are with the Institute of Transport and Logistics, The University of Sydney, Camperdown, NSW 2006, Australia. (e-mail: michael.bell@sydney.edu.au; 
ze.wang@sydney.edu.au; glenn.geers@sydney.edu.au;)} 
\thanks{Xusheng Yao (Correspond Author) is with the College of Management and Economics, Tianjin University, Tianjin University, 300072, China. (e-mail: xsyao@tju.edu.cn)} 
}


\markboth{Submisstion to xxx,~Vol.~xx, No.~x, Month~Year}%
{CCDSReFormer: Traffic Flow Prediction with a Criss-Crossed Dual-Stream Enhanced Rectified Transformer Model}

\maketitle

\begin{abstract}

Accurate, effective, and rapid traffic forecasting is crucial for intelligent traffic systems, playing a significant role in urban traffic planning, management, and control. Despite the effectiveness of existing Spatio-Temporal Transformer models in predicting traffic flow, they face challenges in balancing computational efficiency with accuracy, a tendency to favor global over local time series information, and a segregated approach to spatial and temporal data, which limits understanding of complex spatio-temporal interactions. To address these limitations, we introduce the Criss-Crossed Dual-Stream Enhanced Rectified Transformer model (CCDSReFormer), featuring three novel modules: Enhanced Rectified Spatial Self-attention (ReSSA), Enhanced Rectified Delay Aware Self-attention (ReDASA), and Enhanced Rectified Temporal Self-attention (ReTSA). These modules collectively reduce computational demands through sparse matrix attention, prioritize local information for a comprehensive capture of traffic dynamics, and integrate spatial and temporal data through a criss-crossed learning approach. Our extensive experiments on six real-world datasets demonstrate the superior performance of CCDSReFormer. Additionally, an ablation study was conducted to evaluate the contribution of each component to the model's predictive accuracy. The results confirm the effectiveness of our proposed model in accurately forecasting traffic flow.
\end{abstract}

\begin{IEEEkeywords}
Transformer;  Traffic Flow; Prediction;  Spatio-Temporal Data;  Time-Series Prediction
\end{IEEEkeywords}

\section{Introduction}

At the core of advancing intelligent transportation systems (ITS) is the intricate challenge of traffic flow prediction, crucial for enhancing route efficiency, mitigating congestion, and reducing accident rates in urban settings. The rapid evolution of data acquisition technologies, coupled with the exponential increase in vehicle numbers, has facilitated a more comprehensive collection of traffic data, propelling traffic flow prediction to the forefront of urban transportation management research \cite{ITS2011}.

The quintessential hurdle in this domain is the accurate modeling and understanding of spatiotemporal dependencies within traffic data. These dependencies, reflecting the dynamic nature of traffic flow, involve complex patterns of spatial and temporal interplay that are constantly evolving. Capturing these patterns is fundamental for accurate prediction and poses a significant computational challenge, necessitating a delicate balance between prediction accuracy and computational efficiency \cite{jiang2023pdformer, GOMES2023200268}.

Traditional methodologies, such as autoregressive integrated moving average (ARIMA) models \cite{ahmed1979analysis, zeng2008short, chen2011short, tong2008highway} and Kalman filters \cite{KUMAR2017582, EMAMI2020102025, xu2017real, zhou2019hybrid, gao2013application}, have made strides in traffic prediction. Yet, their effectiveness is often limited by inherent assumptions of data stationarity and their inadequacy in capturing the complex spatiotemporal dependencies characteristic of traffic flow. Spatial regression methods \cite{Zheng_Ni_2013}, while introducing spatial awareness, remain constrained by the dynamic intricacies of traffic patterns and the computational burdens they impose.

With the increasing computational power, deep learning has become the leading methodology in traffic flow prediction, excelling in handling the complex and nonlinear patterns of traffic data. In the past, emerging studies have focused on employing recurrent neural networks (RNNs) and its variants, such as long short-term memory (LSTM)~\cite{oliveira2021forecasting, luo2019spatiotemporal, ma2021short}, which are particularly suited to grid-based data \cite{fu2016using, hochreiter1997long}. These methods are crucial in modeling temporal aspects of traffic. Convolutional Neural Networks (CNNs), which  are also suitable for grid data, have been widely used due to their efficacy in extracting temporal features \cite{zhang2017stresnet, zhang2019short, ProSTformer}. 

More recently, graph convolutional networks (GCNs), aligning with the graph-structured nature of traffic data, have shown promise in identifying intricate spatial characteristics and dependencies in road networks~\cite{ZHENG2022128274, CHEN2022522, XING2023102513, tgcn2020}. However, GNNs in traffic prediction encounter specific challenges: they struggle with dynamic spatial dependencies that change over time, have a limited capacity for long-range dependency modeling due to local interaction focus, and often overlook the time-delay impact in spatial information transfer \cite{hochreiter1997long}. While the attention mechanism has demonstrated effectiveness in capturing both spatial and temporal information by adapting to the input changes \cite{jiang2023pdformer, lin2020adsan, yan2021tformer}, there remain some limitations such as:

\begin{enumerate}
    \item Lack of integration of both spatial and temporal information for learning the complex dynamics;
    \item Less adept at making short-term predictions (or lack of local feature focus); and
    \item Their use of dense matrices may cause high computational cost.
\end{enumerate}

To fill these gaps, we propose a state-of-the-art technique call the Criss-Cross Dual-Stream Enhanced Rectified Transformer (\textbf{CCDSReFormer}) model which consists of a novel Criss-Crossed Dual-Stream for enriched spatial and temporal learning, an Enhanced Convolution (\textbf{EnCov}) method focusing on local traffic pattern nuances, and a rectified linear self attention (\textbf{ReLSA}) mechanism for dynamic and computationally efficient attention allocation in traffic flow prediction. We show that our new method out-performs extant techniques used for analyzing spatial and temporal data in traffic networks. 

Our main contributions are summarized as follows:
\begin{itemize}
    \item Our model introduces a criss-crossed dual stream, enabling simultaneous learning of spatial and temporal information to enhance performance. This dual approach effectively captures the complexities of traffic flow, offering a thorough understanding of spatial and temporal dynamics.
    \item We design a locally enhanced convolution within our attention mechanism, to make the model focus on local spatio-temporal features and capture the nuanced dynamics of traffic patterns influenced by localized conditions.
    \item To address the limitations of traditional softmax-based attention, our model is the first to employ Rectified Linear Self Attention (\textbf{ReLSA}) \cite{ReLABiao2021} in the traffic flow prediction field, offering a dynamic, adaptable approach that responds to the unique spatial-temporal characteristics of traffic data while reducing computational complexity.
    \item We conducted extensive experiments across six real-world datasets, demonstrating that our model outperforms the existing state-of-the-art in both performance and computational efficiency, with robust parameter tuning capabilities.
\end{itemize}

The rest of this paper is structured as follows: Section \ref{Literature} provides a summary of previous related studies. Section \ref{Preliminaries} outlines the relevant definitions and problem statement. The proposed \textbf{CCDSReFormer} model is elaborated in Section \ref{Method}. Section \ref{Experiment} discusses the experiments conducted and their results. The paper concludes with a summary and future research directions in Section \ref{Conclusion}.

\section{Literature Review}
\label{Literature}

In previous studies, parameter-based approaches such as ARIMA models~\cite{ahmed1979analysis,zeng2008short,chen2011short,tong2008highway}, Kalman filters \cite{KUMAR2017582,EMAMI2020102025,xu2017real,zhou2019hybrid,gao2013application}, and other regression techniques~\cite{alam2019prediction,priambodo2017predicting} have been used for traffic flow prediction. However, they must be trained on large data sets to achieve higher precision  and they can only extract the time features from traffic flow data effectively ignoring the spatial information which is important for predicting traffic flow~\cite{EMAMI2020102025}. Non-parametric machine learning methods such as artificial neural networks (ANNs)~\cite{topuz2010hourly,zeng2008short,sharma2018ann,cohen2019pedestrian}, k-nearest neighbors (KNN)~\cite{luo2019spatiotemporal,rahman2020short,cai2020sample,yang2019k} and support vector machines (SVM) \cite{rahman2020short} have also been used to predict traffic flow, with varying degrees of success.

Deep learning models have notably improved traffic prediction methodologies. This section reviews various methods that utilize deep learning for traffic prediction. Generally, applications of deep learning in this domain can be classified into four major categories RNNs, CNNs, GNNs and the attention mechanism, each with its unique advantages.

RNNs including LSTM  are commonly applied to
sequence data because of their memorization capability, which can learn both long and short-term dependencies between parts of a data sequence. Previously they have been used for traffic flow prediction as exemplified in~\cite{oliveira2021forecasting,luo2019spatiotemporal,ma2021short,LI2021102977,YANG2021103228}.  These papers demonstrated the ability of these methods to capture the temporal patterns in traffic data. However, by their  ability to take in very long data sequences, these methods suffer from the vanishing gradient problem~\cite{hochreiterVanishingGradientProblem1998}. 

CNNs have shown proficiency in processing grid-based spatial data, particularly in capturing spatial dependencies within traffic data, making them a go-to choice for early traffic prediction tasks as evidenced by studies such as~\cite{zhang2017stresnet,KE2017591,DUAN2016168}. More recently, \cite{WU2018166, kong2023variational} proposed a hybrid model that combines CNNs for spatial feature extraction and gated recurrent units (GRUs) for temporal feature analysis in traffic flow data. The research in \cite{zhang2019short} further harnesses CNNs, employing a spatio-temporal feature selection algorithm to optimize input data and extract pivotal traffic features, enhancing the predictive model.

Recently, the spotlight has turned to Graph Neural Networks (GNNs) for their adeptness in handling traffic data, effectively represented as graphs. Unlike the rigid structure of CNNs, GNNs embrace a flexible, graph-based approach suitable for the complex, non-Euclidean topology of traffic networks~\cite{asifGraphNeuralNetwork2021}. They've been increasingly utilized for traffic flow prediction, with innovations such as integrating learnable positional attention in GNNs for capturing spatial-temporal patterns~\cite{wang2020traffic}, the Progressive Graph Convolutional Network (PGCN) enhancing adaptability to both training and testing phases~\cite{shin2024pgcn}, and models like the Dynamics Extractor and node connection strength index enhancing traffic flow characteristics analysis~\cite{chen2023node}. Other notable developments include DCRNN's use of bidirectional random walks for complex spatial-temporal dynamics~\cite{li2018dcrnn}, STGCN's efficient convolutional graph structures~\cite{yu2018stgcn}, GWNET's adaptive graph modeling~\cite{wu2019gwnet}, and MTGNN's novel graph learning techniques~\cite{wu2020mtgnn}. Additionally, STFGNN and STGNCDE offer advanced spatio-temporal forecasting by integrating differential equations with neural networks~\cite{li2021stfgnn, choi2022stgncde}. The paper \cite{guo2020optimized} uses an optimized GCN to preserve the spatial structure of road networks through a graph representation. AST-InceptionNet introduces multi-scale feature extraction and adaptive graph convolutions to address spatial heterogeneity and unknown adjacencies~\cite{wang2023adaptive}. Despite their advancements, these models confront challenges with dynamic spatial dependencies, long-range dependency modeling, over-smoothing, and time-delay impacts, indicating ongoing areas for refinement.

Therefore, the growing adoption of attention mechanisms in traffic forecasting is becoming increasingly prevalent. The self-attention mechanism developed by~\cite{Vaswani2017attention}, has significantly impacted data processing for complex tasks, including language translation, image recognition, and sequence prediction, by dynamically focusing on the most pertinent input data sections. This mechanism is increasingly acknowledged for its ability to concurrently process spatial and temporal information. In traffic forecasting, models like the spatial-temporal transformer network model (STTN)\cite{xu2020sttn} uses spatial transformers and long-range temporal dependencies that dynamically model directed spatial dependencies, a graph multi-attention network model (GMAN)\cite{zheng2020gman} utilizing an encoder-decoder architecture with spatio-temporal attention blocks to model the impact of spatio-temporal factors on traffic conditions, and featuring an attention layer that effectively links historical and future time steps for long-term traffic prediction, an attention-based spatial-temporal graph neural network (ASTGNN)\cite{guo2021astgnn} integrating a graph convolutional recurrent module (GCRN) with a global attention module, designed to effectively model both long-term and short-term temporal correlations in traffic data, and propagation delay-aware dynamic long-range transformer for traffic flow prediction (PDFormer)\cite{jiang2023pdformer} design multi-self-attention modules to capture the dynamic relations, a progressive space-time self-attention (ProSTformer)\cite{ProSTformer} focuses on spatial dependencies from local to global regions and a novel traffic flow prediction approach combining Vision Transformers (VTs) and Convolutional Neural Networks (CNN)\cite{ramana2023vision} is used to accurately forecast urban congestion. These models collectively demonstrate the growing trend of incorporating advanced attention mechanisms to improve the accuracy and efficiency of spatio-temporal traffic forecasting. However, the complexity of these attention mechanisms can lead to increased computational costs, presenting a challenge that needs to be addressed. Finding a balance between computational efficiency and model accuracy remains a critical issue to resolve.

In summary, traffic flow prediction has progressed from traditional parameter-based methods to sophisticated deep learning approaches, offering a nuanced understanding of the complex spatio-temporal dynamics within transportation networks. While RNNs, CNNs, GNNs, and attention mechanisms have improved prediction accuracy, they continue to face challenges in balancing computational cost and efficiency. Additionally, achieving an optimal mix of local and global information is pivotal, as long-term forecasts tend to prioritize broader trends, potentially overlooking critical local insights. Addressing these issues is key to refining accuracy and computational practicality, steering future research toward more adept and streamlined forecasting techniques.

\section{Preliminaries}
\label{Preliminaries}
In this section, we provide background information and introduce fundamental ideas that are pertinent to the traffic flow prediction scheme presented in Section~\ref{prelim} and the vanilla attention mechanism discussed in Section~\ref{Attention_Mechanism}.

\subsection{Essential Notation and Definitions}\label{prelim}
To better understand our work,  we first present several preliminaries and define our problem formally in this subsection.

\begin{definition}[Graphical Representation of a Road Network]
A \textit{Road Network} is modeled as a graph \( \mathcal G = (\mathcal V, \mathcal E, \mathcal A) \), where $ 
\mathcal V = \{v_1, \ldots, v_N\} $ is a set of $N (=|\mathcal{V}|)$ nodes 
composed from points of interest in the network such as traffic sensors, intersections, or subway stations, \( \mathcal E \subseteq \mathcal V \times \mathcal V \) is a set of edges typically determined by roads that connect the nodes, and \( \mathcal A \) signifies the (square) adjacency matrix associated with the network graph. 
\end{definition}

\begin{figure}[t]
    \centering
\includegraphics[scale = 0.16]{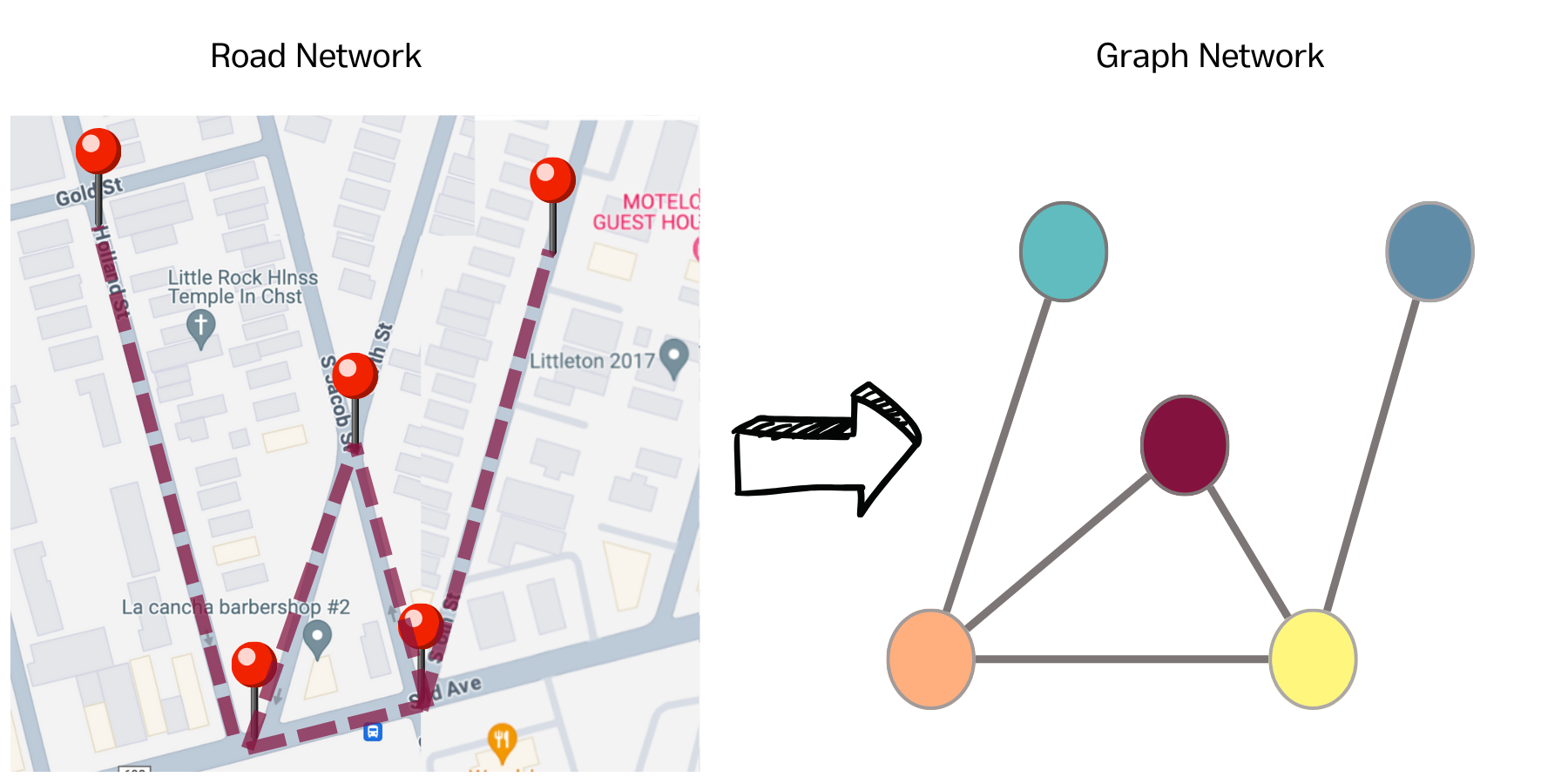}
    \caption{The graph network is structured to reflect the local road network topology. The figure on the left illustrates the geographical locations of a subset of traffic sensor nodes as marked on the chosen benchmark dataset. On the right, the corresponding graph network is depicted, composed of these sensor nodes interconnected according to their locations and the road network links they monitor.}
    \label{fig:roadgraph}
\end{figure}

\begin{figure}[t]
    \centering
    \includegraphics[width=0.5\textwidth]{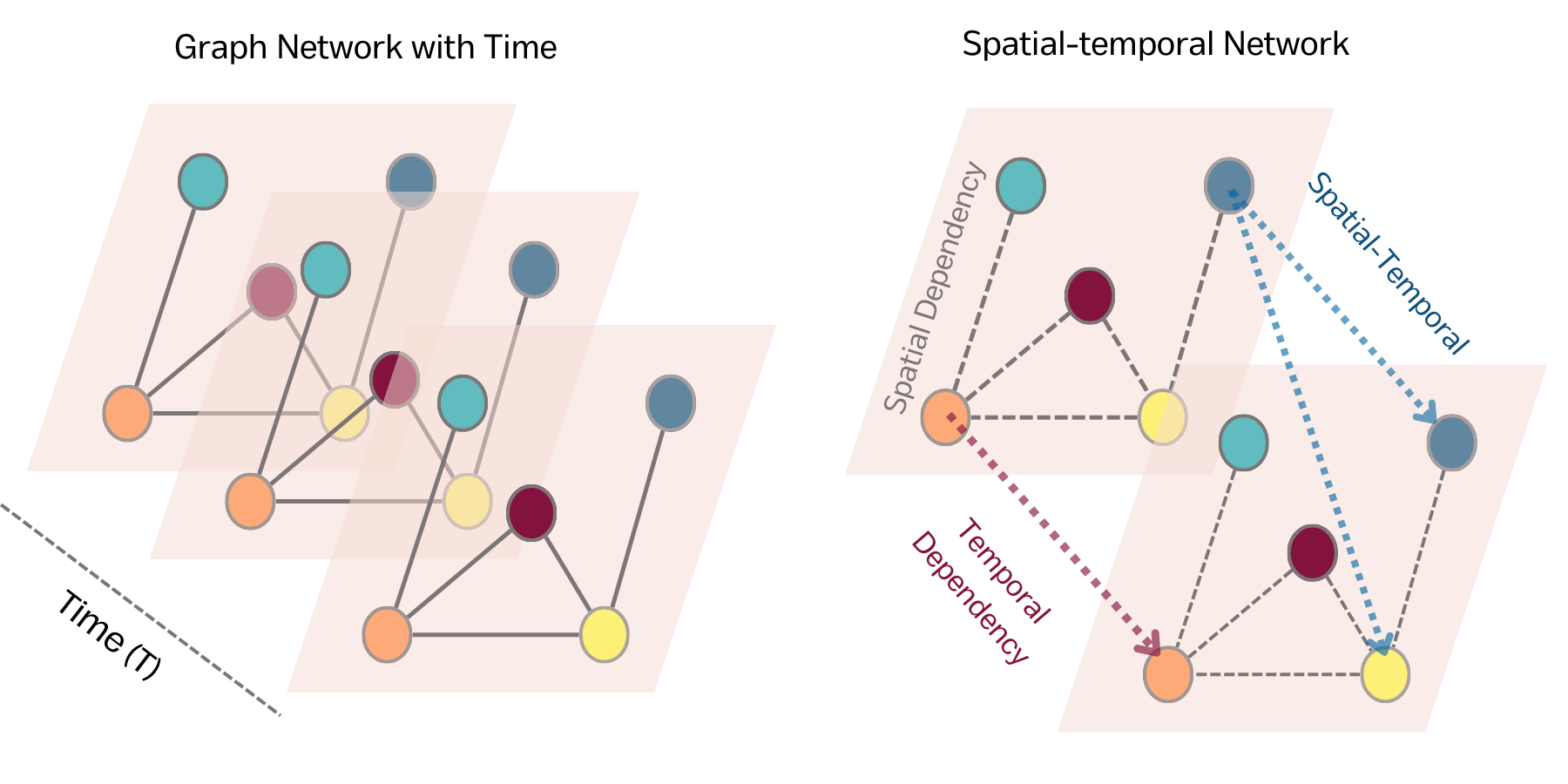}
    \caption{The figure on the left represents the structure of graph data, and that on the right illustrates the spatio-temporal network of the blue node. Grey lines represent the network's connections, showing spatial influence. The red arrow reflects the self-influence of the blue node over time, while the blue arrows demonstrate the combined spatial and temporal influence between the blue node and its neighboring nodes at the subsequent time step. It is similar applies to the grid data.}
    \label{fig:spatial-temp}
\end{figure}

Figure~\ref{fig:roadgraph} illustrates an example of a traffic network depicted as a graph network. Although the graph is not changing,
the node information may change during the time when data is collected at the sensor nodes. Thus, the network can be regarded as a spatio-temporal graph network as shown in Figure~\ref{fig:spatial-temp}, showcasing the dynamic nature of traffic data over time. This gives rise to the following:
\begin{definition}[Tensor Representation of Traffic Flow] Given a road network $\mathcal{G}$ with $N$ nodes, 
the traffic flow information on the road network at timestamp $t$ is denoted by \(\mathbf X_t \in \mathbb{R}^{N \times d}\) where $d$ is the dimensionality of the flow features. For any given $T>0$, the overall traffic flow can be represented by a rank-3 tensor, denoted by
\begin{equation*}
\mathbf X = ( \mathbf X_1,  \mathbf X_2, \ldots,  \mathbf X_T) \in \mathbb{R}^{T \times N \times d},
\end{equation*}
where the first mode (or dimension) of the tensor corresponds to the timestamps, the second mode to the network nodes and the third mode corresponds to the flow channels (or dimensions).
\end{definition}

Given a traffic flow tensor  \( \mathbf X \in \mathbb{R}^{T \times N \times d}\) over a road network $\mathcal{G}$, the \(t\)-slice over the first mode is denoted by \( \mathbf X_{t::} \in \mathbb{R}^{N \times d}\) which contains the spatial information for all nodes at time $t$, and the \(n\)-slice over the second mode is denoted by \( \mathbf X_{:n:} \in \mathbb{R}^{T \times d}\) which is the temporal data at node $n$ for all timestamps $[1, 
 \dots, T]$.


\paragraph{Problem Statement}
The objective of traffic flow forecasting is to predict future traffic flow based on historical flow data. Specifically, for a given section of traffic flow tensor \(\mathbf X\) recorded on the road
network $\mathcal{G}$, the aim is to compute a function \(f\) that takes traffic flow values from the past \(m\) timestamps as input and outputs an estimate of the traffic flow for the following \(n\) timestamps: 
\begin{equation*}
f\left( [ \mathbf X_{t-h+1}, \ldots,  \mathbf X_{t}] ; \mathcal{G} \right) \mapsto [ \mathbf X_{t+1}, \ldots,  \mathbf X_{t+h'}]
\end{equation*}
for any given initial timestamp $t$.

\subsection{The Attention Mechanism}
\label{Attention_Mechanism}

In order to introduce the new attention modules used in our proposed model we first describe the attention mechanism in detail by explaining scaled dot-product or `vanilla' self-attention~\cite{Vaswani2017attention}.

\paragraph{Vanilla Self-Attention}  
The principle of the attention mechanism, originating in natural language processing (NLP)~\cite{BahdanauChoBengio2015}, is to map query terms to a weighted sum of known keys and output the best matched values. A commonly used attention mechanism in various
machine learning models is the so-called self-attention framework. 

The primary objective of self-attention is to seek a more information-rich representation $\mathbf Y$ of a given set of feature data $\bar{\mathbf X}\in \mathbb{R}^{N\times d}$ by exploiting so-called `self-attention' information.  To realize this goal, we first construct 
three components: query ($\mathbf Q$), key ($\mathbf K$), and value ($\mathbf V$), from $\bar{\mathbf X}$ by, 
\begin{equation}
\mathbf Q =  \mathbf {\bar X}  \mathbf W_Q, \quad \mathbf K =  \mathbf {\bar X}   \mathbf W_K, \quad \mathbf  V =  \mathbf {\bar X}   \mathbf W_V
\label{Eq:QKV}
\end{equation}
where 
\( \mathbf W_Q \in \mathbb{R}^{N\times d_q}\), \( \mathbf W_K \in \mathbb{R}^{N\times d_k}\), and \( \mathbf W_V \in \mathbb{R}^{N\times d_v}\)  are weight matrices that must be learned from the input data; and $d_q$, $d_k$ and $d_v$ are the dimension of the query, key and value vectors respectively. For convenience it is usual to set $d_q=d_k=d_v$.

The new representation $\mathbf Y$ is calculated in the following way based on the attention mechanism principle. 
Once we have $\mathbf Q\in\mathbb{R}^{N\times d_q}$, $\mathbf K\in\mathbb{R}^{N\times d_k}$, and $\mathbf V\in\mathbb{R}^{N\times d_v}$ according to \eqref{Eq:QKV}, we are able to determine how the features in $\bar{\mathbf X}$ pay influence or ``attention'' to all other features through the query features $\mathbf Q$ and key features $\mathbf K$ by the following 
scaled attention score $\mathbf A$ given by 
\begin{equation}
\mathbf A = \frac{\mathbf{Q}\mathbf{K}^\top}{\sqrt{d_k}} \in \mathbb{R}^{N\times N}, \label{Eq:Score}
\end{equation}
where we have assumed $d_k=d_q$. The atttention score is scaled by the square root of the dimension of the key vectors to stabilize  numerical calculations during the training process.

Clearly the $ij$th element of $\mathbf A$ is the scaled dot-product of the $i$th and $j$th feature of $\bar{\mathbf X}$ because they correspond to the the query and key vectors respectively. Thus the $i$th row of $\bar{\mathbf X}$ contains the attention score of the $i$th feature with respect to all other features including itself. 

The attention score matrix  is then passed through a $\text{softmax}$ function (a generalization of the logisitic function to dimensions greater than one) to obtain the attention weight $\textbf{O}$, in which each element is positive and every row sums to unity,
\begin{equation}
\textbf{O} = \text{softmax}(\mathbf{A}). \label{Eq:Output}
\end{equation}
The new representation $\mathbf Y$ is then computed by, 
\begin{equation}
\mathbf Y = \textbf{O}\mathbf V, \label{Eq:Y}
\end{equation}
where the dimensions of $\mathbf Y$ are the same as the dimensions of the value matrix $\mathbf V$. 

We call the implementation \eqref{Eq:QKV} -- \eqref{Eq:Y} one-head attention, offering one way of exploration. In multi-head attention,  we can utilize multiple sets (or ``heads'') of \(\mathbf Q\), \(\mathbf K\), and \(\mathbf V\) by learning different weight matrices $\mathbf W_Q, \mathbf W_K, \mathbf W_V$, producing multiple outputs $\mathbf Y$ that can be concatenated and linearly transformed to produce the final output.

\textit{Remark 1:} In this paper we set $d_q = d_k = d_v = d_0$ for a given $d_0$.

\section{The Method}
\label{Method}
\begin{figure}[h]
    \centering
    \includegraphics[width=0.4\textwidth]{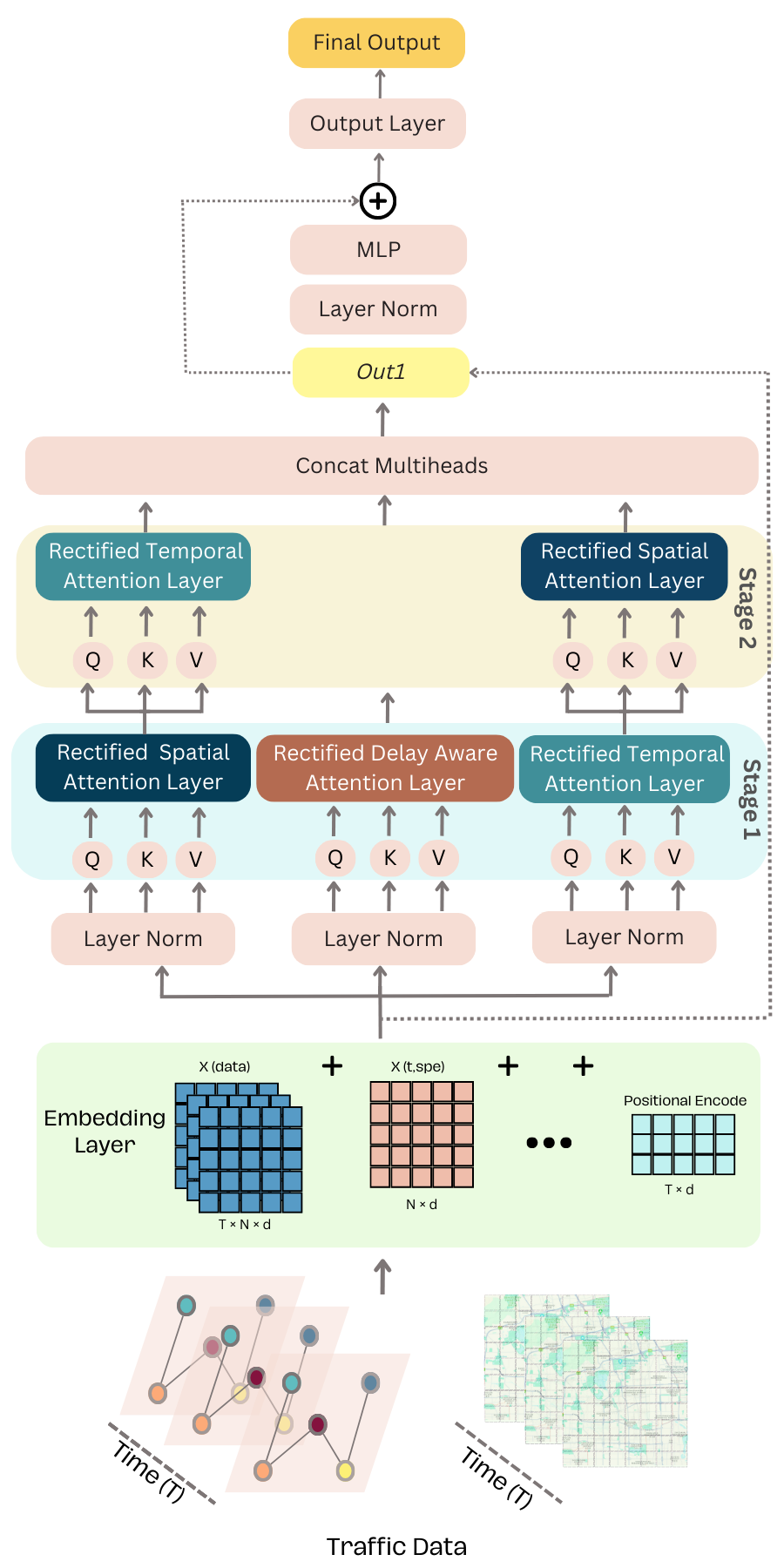}
    \caption{The figure shows the process of \textbf{CCDSReFormer} which begins with the input traffic data being directed to the embedding layers. Subsequently, a layer normalization (Lay Norm) is applied. Following this, the model engages in various attention mechanisms including \textbf{\textbf{ReSSA}}, \textbf{ReTSA}, and \textbf{ReDASA}, facilitating cross-learning between \textbf{\textbf{ReSSA}}, \textbf{ReTSA} to assimilate information from various sources. The subsequent concatenation of these results yields the initial output, denoted as \textbf{$Out_1$}. This is followed by another layer normalization and a multi-layer perceptron (MLP) process. The output thus obtained is combined with the solution by concatenating the output from the data embedding layer and \textbf{$Out_1$}. Finally, this composite output is fed into the output layer to generate the final output. }
    \label{fig:reformer}
\end{figure}

In this section, we introduce the proposed \textbf{CCDSReFormer} in detail. The framework of \textbf{CCDSReFormer} is shown as Figure ~\ref{fig:reformer}.

\subsection{Data Embedding Layer}
Similar to the study \cite{jiang2023pdformer}, the data embedding layer morphs the input into a higher-dimensional representation, \( \mathbf X_{\text{data}} \in \mathbb{R}^{T \times N \times d}\), via a fully connected layer where \(d\) 
is the embedding dimension. 

To further explore and incorporate spatio-temporal network dynamics, the output of the data embedding layer 
is used as input for computing a spatial embedding of the road network structure as described below. 

In similar vein, a temporal periodic embedding is used to capture recurrent variations in traffic flow such as the morning and afternoon rush hours.

\paragraph{Spatial Embedding} For each timestamp $t$, the spatial embedding \( \mathbf X_{t,\text{spe}} \in \mathbb{R}^{N \times d}\) is derived from the graph Laplacian eigenvectors. The process begins with the computation of the normalized Laplacian matrix \(\mathbf \Delta\) which is calculated using the adjacency matrix \(\mathbf{\mathcal A}\) and the diagonal degree matrix \(\mathbf D\) where \( \mathbf D (i, i) = \sum_{j=1}^{N} \mathcal A(i, j) \). The symmetrically  normalized Laplacian matrix is given by \(\mathbf \Delta = \mathbf I - \mathbf D^{-1/2}\mathcal A \mathbf D^{-1/2}\), where \( I \) is the identity matrix of appropriate dimension. Secondly an eigenvalue decomposition of \( \mathbf \Delta \) is performed, resulting in \( \mathbf \Delta = \mathbf U^{\top} \mathbf \Lambda \mathbf U \) where \( \mathbf \Lambda = \text{diag}(\lambda_0, \lambda_1, \ldots, \lambda_{N-1}) \) is the ordered matrix of eigenvalues, satisfying \( 0 = \lambda_0 \leq \lambda_1 \leq \ldots \leq \lambda_{N-1} \), and \( \mathbf U = (u_0, u_1, \ldots, u_{N-1}) \) is the corresponding matrix of eigenvectors. 
We then select the \( k \) eigenvectors from $\mathbf U$ corresponding to the  $k$ smallest nontrivial eigenvalues to construct the $k$-dimensional embedding for all nodes \( V \) at time \(t\), denoted by \( \mathbf{U}_{t,{\text{spe}}} = [u_1, u_2, \ldots, u_k] \in \mathbb{R}^{N \times k} \). In the following the parameter \( k = 8 \) as used in the baseline model~\cite{jiang2023pdformer}. Subsequently, \( \mathbf{{U}}_{t,{\text{spe}}} \) is subjected to a linear transformation, mapping it into a new \( d \)-dimensional space. This process culminates in the formation of the spatial graph Laplacian embedding \( \mathbf{X}_{t,\text{spe}} \in \mathbb{R}^{N \times d} \) at time \( t \), effectively embedding the graph in a Euclidean space and thus preserving its global structure
\cite{jiang2023pdformer}.


\paragraph{Temporal Embedding} 
Each timestamp $t$, can be converted to either a weekly index $w(t)$ or a daily index $d(t)$. Concretely, the weekly index \(w(t)\) translates the timestamp  \(t\) into a day-of-week representation (1 to 7), while the daily index \(d(t)\) maps it to the specific minute of the day (1 to 1440). All the indices are then converted into trainable temporal embeddings by a way of the embedding layers. 
The weekly and daily embeddings for all the timestamps are denoted by \( \mathbf X_w \in \mathbb{R}^{T \times d}\) and \(\mathbf X_d \in \mathbb{R}^{T \times d}\), respectively, where $d$ is the same as the spatial embedding dimension. 

\paragraph{Temporal Positional Encoding}
For generating the temporal positional encoding \( \mathbf{X}_{tpe} \), we draw inspiration from the study by Vaswani et al. (2017)  \cite{Vaswani2017attention}. 
In the context of traffic network prediction, it's crucial to account for another dynamic attribute: the temporal position relative to the input. To address this, we define the temporal input positions as \(t = \{1, 2, \dots T\}\), representing the sequence of timestamps.
To capture this temporal aspect effectively, we define the temporal positional encoding
as \( \mathbf{X}_{t,\text{tpe}} \) for each timestamp \( t \) as a $d$-dimensional vector in \(\mathbb{R}^d \). 
The components of \( \mathbf{X}_{t,\text{tpe}} \) are computed as follows:
\begin{equation*}
\mathbf{X}_{t,\text{tpe}}(i) =
\begin{cases}
\sin\left(\frac{t}{10000^{2i/d}}\right) & \text{if } i \text{ is even}, \\
\cos\left(\frac{t}{10000^{2(i-1)/d}}\right) & \text{if } i \text{ is odd}.
\end{cases}
\end{equation*}
Finally we use \( \mathbf{X}_{\text{tpe}} \in\mathbb{R}^{T\times d}\) to collect all the temporal positional encoding. 

This ensures that each dimension of the positional encoding varies according to a sinusoidal function of a different wavelength, providing a unique and continuous encoding for each time step \( t \). Such an encoding is instrumental in enabling the model to capture the nuances of the temporal dynamics inherent in road network data. 

\paragraph{Final Output}
The final output from the data embedding layer, represented as \(\mathbf X_{\text{emb}}\),  is simply the element-wise sum of the various components:
\begin{equation*} 
\mathbf X_{\text{emb}} =  \mathbf X_{\text{data}} \oplus_1  \mathbf X_{\text{spe}} \oplus_2  \mathbf X_w \oplus_2 \mathbf X_d \oplus_2  \mathbf X_{\text{tpe}},
\end{equation*}
where $\oplus_k$ denotes the broadcasting summation along the $k$th mode to ensure dimensional consistency. This concept of broadcasting summation is derived from the functionality provided by Python's NumPy library \footnote{Broadcasting summation refers to the capability in Python's NumPy library \href{https://numpy.org/doc/stable/user/basics.broadcasting.html} to perform element-wise binary operations on arrays of different sizes }.
The resulting  \(\mathbf X_{\text{emb}}\) is then passed to the spatio-temporal encoders. To simplify the notation the subscript $\text{emb}$ will be dropped in the following sections, i.e., $\mathbf{X} \equiv \mathbf{X}_{\text{emb}}$.

\subsection{Workflow of Enhanced Rectified Linear Self-Attention}
\label{generalrelsa}

To provide a clear and logical introduction to the \textbf{CCDSReFormer}, we begin with an overview of our newly designed attention which is named Enhanced Rectified Linear Self-Attention (\textbf{EnReLSA}), as depicted in Figure~\ref{fig:reatt}. Understanding this concept is crucial for grasping the spatial, temporal, and delay-aware modules that are central to the architecture of \textbf{CCDSReFormer}.

The \textbf{EnReLSA} approach adapts the standard self-attention mechanism by 
initially determining the query, key, and value components through specific matrix operations. It modifies these matrices with learnable parameters to tailor the model's focus. Further, to enhance the handling of the computational demand posed by the attention mechanism, \textbf{EnReLSA} incorporates a summation of Rectified Self-Attention \textbf{(ReLSA)} and enhanced convolution \textbf{EnCov}. 

The \textbf{(ReLSA)} approach introduces sparsity into the attention matrix, effectively reducing its complexity. Specifically, \textbf{ReLSA} employs a rectified linear unit (ReLU) for selecting positive values. Additionally, to further stabilize the attention distribution, \textbf{ReLSA} leverages a layer normalization technique known as \textbf{RMSNorm}. Then, an enhanced convolutional step (\textbf{EnCov}) is designed to enhance the localizes attention, allowing the model to hone in on adjacent features, which enhances its ability to discern dynamic traits in the data. Thus, the final out of \textbf{EnReLSA} is able to capture both local and global dependencies in the data.

The subsequent sections of the text promise a deeper exploration of the components of the \textbf{CCDSReFormer}, offering the workings of \textbf{EnReLSA} with different inputs and its contribution to the model's performance.

\subsection{Rectified Spatial Self-Attention Module (\textbf{ReSSA})}
\begin{figure}[t]
    \centering
    \includegraphics[width=0.35\textwidth]{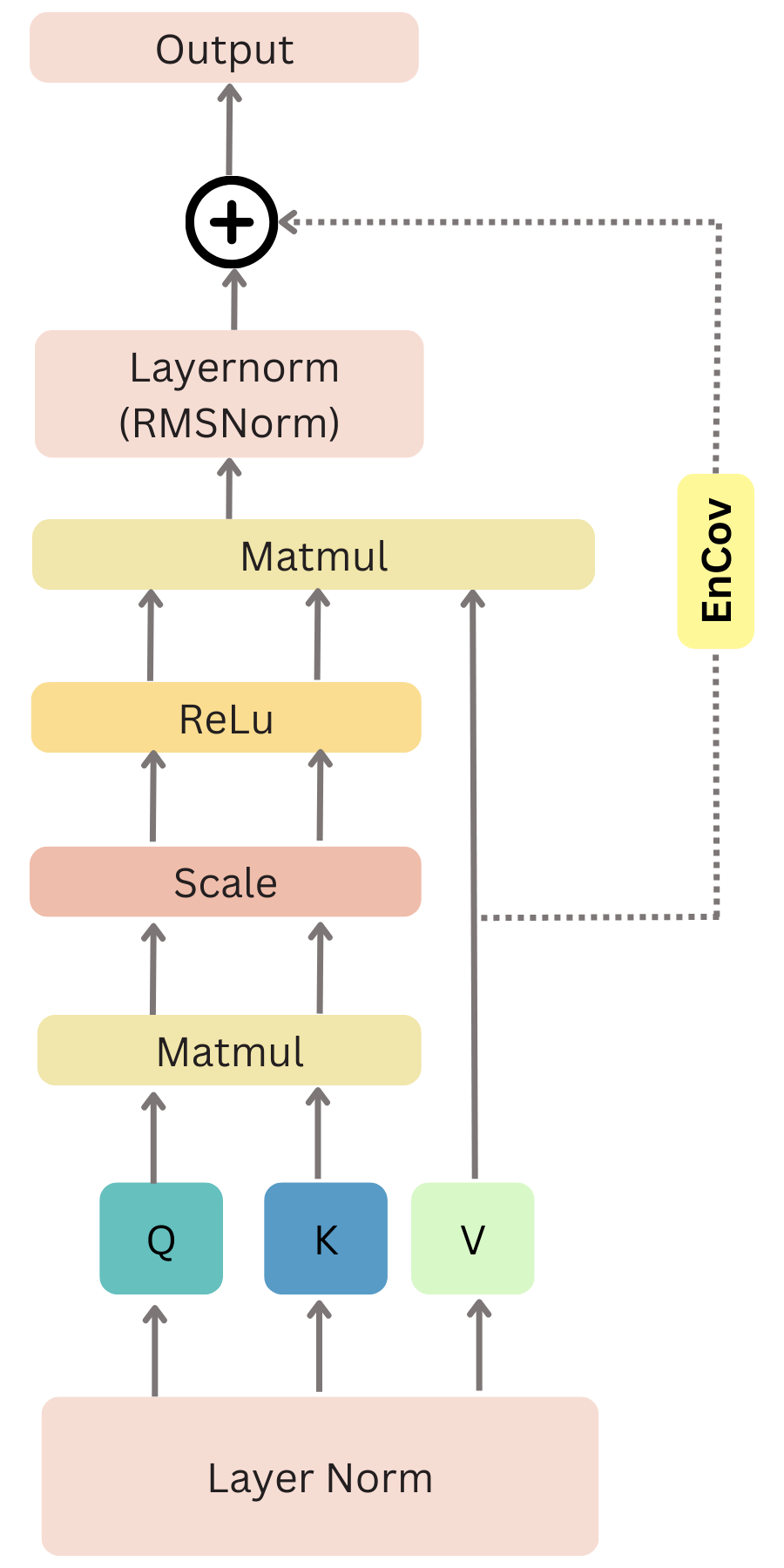}
    \caption{The figure shows the proposed the Enhance Rectified Linear Self Attention(ReLSA) with \textbf{EnCov} workflow.}
    \label{fig:reatt}
\end{figure}
Rectified spatial self-attention (\textbf{ReSSA}) is used to capture the dynamic spatial dependence of traffic data at reduced computational cost. As previously introduced the attention mechanisms in Section \ref{Attention_Mechanism}, at each time $t$, the query, key, and value of self-attention in the rectified spatial self-attention module can be written as, referring from Figure~\ref{fig:reatt},
\begin{equation*}
\textbf{Q}^{(S)}_t = \mathbf{X}_{t::} \cdot \mathbf{W}^S_Q\text{,} \; 
\textbf{K}^{(S)}_t = \mathbf{X}_{t::} \cdot \mathbf{W}^S_K\;\text{and} \;
\textbf{V}^{(S)}_t = \mathbf{X}_{t::} \cdot \mathbf{W}^S_V
\end{equation*}
where $\mathbf{W}^S_Q$, $\mathbf{W}^S_K$ and $\mathbf{W}^S_V \in \mathbb{R}^{d \times d_0}$ are learnable parameters and $d_0$ is the dimension of the query, key, and value matrix in this work. Then, we apply self-attention operations in the spatial dimension to model the interactions between nodes and obtain the spatial dependencies (attention scores) among all nodes at time $t$ as:
\begin{equation}\label{ReSSA-att-1}
\mathbf{A}^{(S)}_t = \frac{(\mathbf{Q}^{(S)}_t)(\mathbf{K}^{(S)}_t)^{\top}}{\sqrt{d_0}}.
\end{equation}
It can be seen that the spatial dependencies $\mathbf{A}^{(S)}_t \in \mathbb{R}^{N \times N}$ between nodes are different in different time slices, i.e., dynamic. This variability necessitates a self-attention mechanism capable of adapting to these dynamic spatial dependencies. Traditional self-attention models, which assume fully connected graphs, do not always align with the more complex relationships found in real-world scenarios. In particular, the interactions between nodes that are geographically proximate or share certain functional similarity, regardless of their physical distance, are crucial.

To address this issue, our model incorporates a binary geographic masking matrix $\mathbf{M}_{\text{geo}}$, as proposed in study \cite{jiang2023pdformer}. This matrix is specifically designed to account for geographic contiguity, giving priority to nodes within a predefined distance threshold. Such a focus allows for a more nuanced representation of spatial dependencies that are geographically grounded.

Additionally, to enhance computational efficiency, our model integrates the concept of rectified attention as detailed in the work \cite{ReLABiao2021}. The introduction of the Rectified Linear Spatial Self-Attention module (ReLSA) circumvents the re-centering constraint, resulting in a methodology that is not only more flexible but also computationally less demanding.

The final output of the ReLSA module is obtained by multiplying the attention scores with the value matrix. The formulation for this output is as follows:
\begin{align}\label{ReSSA_relsa}
& \textbf{ReLSA}_{\text{geo}}(\mathbf{Q}^{(S)}_t, \mathbf{K}^{(S)}_t, \mathbf{V}^{(S)}_t) \nonumber\\
&= \textbf{LN}( \textbf{ReLU}(\mathbf{A}^{(S)}_t \odot \mathbf{M}_{\text{geo}})\mathbf{V}^{(S)}_t), 
\end{align}
where $\mathbf{M}_{\text{geo}}$ is binary geographic masking matrix as we mentioned earlier, the $\odot$ indicates the Hadamard product. 
$\textbf{RELU}(\cdot) = \max(0, \cdot)$ is the rectified linear unit and \textbf{LN} means `layer normalization' \cite{zhang2019root} which in this work is chosen to be 
$\textbf{RMSNorm}$. So
\begin{eqnarray} \label{RMSNorm}
\textbf{LN}\Big(\textbf{RELU}(\cdot)\Big) & = & \textbf{RMSNorm} \Big(\textbf{RELU}(\cdot)\Big) \\ 
                           & = & \frac{\textbf{RELU}(\cdot)}{\textbf{RMS}\Big(\textbf{RELU}(\cdot)\Big)}\odot g\text{,}
\end{eqnarray}
where $\odot$ signifies the Hadamard (element-wise) product; \( g \), the gain parameter, is an appropriately sized matrix typically initialized with all elements set to unity and 
\textbf{RMS}($\cdot$) calculates the root mean square statistic, contributing to the stabilization of attention.


In the self-attention mechanism, the matrices $\mathbf{Q}^{(S)}_t$, $\mathbf{K}^{(S)}_t$, and $\mathbf{V}^{(S)}_t$ correspond to the query, key, and value components, respectively. These components are essential in computing the pairwise similarities between the queries and keys, where the value matrix $\mathbf{V}^{(S)}_t$ represents a 2D feature map encapsulating spatial and temporal features. To further refine the focus on spatial characteristics and to enhance local feature representation, we implement an Enhanced Convolution (\textbf{EnCov}) with a 3x3 2D convolution, to the value matrix $\mathbf{V}^{(S)}_t$. The \textbf{EnCov} is specifically designed to amplify the spatial information within the self-attention framework. Consequently, the output of the \textbf{ReSSA}, which integrates the benefits of rectified attention and enhanced local feature processing, is given as:
\begin{align*}
        &\text{\textbf{ReSSA}}(\mathbf{Q}^{(S)}_t, \mathbf{K}^{(S)}_t, \mathbf{V}^{(S)}_t) \nonumber\\  
        &= \textbf{EnCov}(\mathbf{V}^{(S)}_t) + \textbf{ReLSA}_{\text{geo}}(\mathbf{Q}^{(S)}_t, \mathbf{K}^{(S)}_t, \mathbf{V}^{(S)}_t) .
\end{align*}

\subsection{Rectified Temporal Self-Attention Module (ReTSA)}

In traffic data, various dependencies exist, such as periodic or trending patterns, among traffic conditions observed in different time slices. The nature of these dependencies can vary depending on the specific situation. Therefore, we have incorporated a Temporal Self-Attention module to effectively identify and capture dynamic temporal patterns. More formally, for a given node, we initially derive the query, key, and value matrices as follows:
\begin{equation}\label{temp_qkv}
    \mathbf Q^{(T)}_n = \mathbf X_{:n:} \cdot \mathbf W^{\top}_Q, 
    \quad\; \mathbf K^{(T)}_n = \mathbf X_{:n:} \cdot \mathbf W^{\top}_K, 
    \quad\; \mathbf V^{(T)}_n = \mathbf X_{:n:} \cdot \mathbf W^{\top}_V,
\end{equation}
where $\mathbf Q^{(T)}_n$, $\mathbf K^{(T)}_n$, and $\mathbf V^{(T)}_n$ represent the query, key, and value matrices for node \(n\), respectively. Here $\mathbf{W}^{\top}_Q$, $\mathbf{W}^{\top}_K$ and $\mathbf{W}^{\top}_V \in \mathbb{R}^{d \times d_0}$ are learnable parameters and $d_0$ is the dimension of the query, key, and value matrices. Then, we apply self-attention operations in the spatial dimension to model the interactions between nodes and obtain the spatial dependencies (attention scores) among all nodes at time $t$ as:
\begin{align}\label{ReSSA-att-2}
\mathbf{A}^{(T)}_n = \frac{(\mathbf{Q}^{(T)}_n)(\mathbf{K}^{(T)}_n)^{\top}}{\sqrt{d_0}}. 
\end{align}
It can be seen that temporal self-attention can discover the dynamic temporal patterns in traffic data that are different for each node. Since the temporal self-attention has a global receptive to model the long-range temporal dependencies among all time slices. Thus, the usage of \textbf{EnCov} can increase the attention to local features. 
This locality ensures that even if two different queries have the same weight under self-attention, they can obtain different outputs from different local features (time, space), thereby better capturing dynamic characteristics. Hence, the rectified function \textbf{LN(RELU($\cdot$))} also can reduce the computational cost. Finally, we can obtain the
the output of the temporal self-attention module as follows:
\begin{align}\label{ReTSA}
   &\textbf{ReTSA}(\mathbf{Q}^{(T)}_n, \mathbf{K}^{(T)}_n, \mathbf{V}^{(T)}_n) \nonumber\\
   &= \textbf{EnCov}(\mathbf{V}^{(T)}_n) + \textbf{LN}( \textbf{ReLU}(\mathbf{A}^{(T)}_n)\mathbf{V}^{(T)}_n).
\end{align} 

\subsection{Rectified Delay Aware Self Attention (ReDASA)}

In the real world, when an accident happening in one area, it might take a few minutes before the traffic in adjacent areas is affected. To model this aspect effectively, we draw inspiration from the concept of Delay Aware Self Attention, as elaborated in \cite{jiang2023pdformer}. This approach is adept at integrating delay-related information into the key matrix, thereby capturing the essence of temporal lags in the propagation of spatial information. For a detailed exposition of this concept, the reader is referred to \cite{jiang2023pdformer}.

Our implementation extends this idea by amalgamating Delay Aware Self Attention with Rectified Self Attention, as depicted in Figure~\ref{fig:reatt}. In this hybrid model, normal Delay Aware Self Attention is used to enrich the key matrix with temporal information. We denote this modified key matrix as $\hat{\mathbf K}_t^{(S)}$. The primary operation in this model involves the computation of the product of the query matrix and $\hat{\mathbf K}_t^{(S)}$, which leads to the derivation of spatial dependencies at the specific time slice $t$. The equation for this computation is as follows:
\begin{equation}\label{redasa_att}
\hat{\mathbf{A}}^{(S)}_t = \frac{(\mathbf{Q}^{(S)}_t)(\hat{\mathbf K}^{(S)}_t)^{\top}}{\sqrt{d_0}}.
\end{equation}

As we mentioned, the self-attention is assumed to be fully connected attention graph. Hence here we employed a graph mask matrix $\mathbf{M}_{\text{sem}}$ alongside the mask matrix  $\mathbf{M}_{\text{geo}}$ as utilized in Eq.~\eqref{ReSSA_relsa}. These matrices enable the simultaneous capture of both short-range and long-range spatial dependencies in traffic data. Then, the ReLSA can be written as:

\begin{align}\label{ReSSA_sem}
&\textbf{ReLSA}_{\text{sem}}(\mathbf{Q}^{(S)}_t, \mathbf{K}^{(S)}_t, \mathbf{V}^{(S)}_t) \nonumber\\
& = \textbf{LN}( \textbf{ReLU}(\mathbf{A}^{(S)}_t \odot \mathbf{M}_{\text{sem}})\mathbf{V}^{(S)}_t).
\end{align}
where the $\odot$ indicates the Hadamard product. 
.

With the sum of \textbf{EnCov}, the final \textbf{ReDASA} can be formulate as:
\begin{align}
           & \textbf{ReDASA}(\mathbf{Q}^{(S)}_t, \mathbf{K}^{(S)}_t, \mathbf{V}^{(S)}_t) \nonumber\\
           & = \textbf{EnCov}(\mathbf{V}^{(S)}_t) + \textbf{ReLSA}_{\text{sem}}(\mathbf{Q}^{(S)}_t, \mathbf{K}^{(S)}_t, \mathbf{V}^{(S)}_t), \label{Eq:21}
\end{align}
In this manner, the rectified spatial self-attention module seamlessly integrates and enhances both short-range geographic proximity by \textbf{ReSSA} and long-range semantic context with \textbf{ReDASA} simultaneously, with the bonus of rectified attention \cite{ReLABiao2021}, the computation complexity is greatly reduced.


\subsection{Criss-Crossed Dual-Stream Learning (CCDS)}
Criss-Crossed learning is designed as crossed-learning for temporal and spatial attention which is depicted in Figure~\ref{fig:reformer}. In the proposed framework, a novel criss-crossed learning approach is employed to harness both spatial and temporal dynamics from time series data. Initially, the input of spatial information is directed through the spatial attention module which is in Stage 1, then the output is fed into the temporal attention module which is in Stage 2. The output, which follows the spatial-temporal attention sequence, is denoted as \( \mathcal{O}^{ReSSA} \) and is defined as:
\begin{align}
    \mathcal{O}^{ReSSA} = \textbf{ReTSA}(\textbf{ReSSA}(\mathbf{Q}^{(S)}_t, \mathbf{K}^{(S)}_t, \mathbf{V}^{(S)}_t))
\end{align}

Simultaneously, the input on time-related information traverses through Stage 1 via the temporal attention module, with its output channeled into Stage 2 with the spatial attention module. The result for transitions from temporal to spatial processing, is represented as \( \mathcal{O}^{ReTSA} \), formulated as:
\begin{align}
        \mathcal{O}^{ReTSA} = \text{\textbf{ReSSA}}(\text{\textbf{ReTSA}} (\mathbf{Q}^{(T)}_n, \mathbf{K}^{(T)}_n, \mathbf{V}^{(T)}_n))
\end{align}

This integrated learning architecture ensures a comprehensive understanding of spatial-temporal relationships, thereby enhancing the model's learning capability by allowing it to capture complex patterns and dependencies inherent in the data, fostering a more robust representation.


\subsection{Attention Mixer and Output layer}
\paragraph{Attention Mixer} In our model, we integrate three types of attention- \textbf{ReSSA}, \textbf{ReDASA}, and \textbf{ReTSA} using a multi-head self-attention block. The Rectified Spatial-Temporal Self-Attention block (\textbf{ReSTSA}) simultaneously processes spatial and temporal information. This integration is achieved by concatenating outputs from each attention head \(h_{ReSSA}\), \(h_{ReDASA}\) and \(h_{ReTSA}\),  then projecting these concatenated results to obtain the final output. For simplicity, we denote the output of \textbf{ReSSA} (two stages), \textbf{ReDASA}, and \textbf{ReTSA} (two stages) with \(\mathcal{O}^{ReSSA}\), \(\mathcal{O}^{ReDASA}\) and \(\mathcal{O}^{ReTSA}\). Then, the \textbf{ReSTSA} block is formally defined as: 
\begin{align*}
    \textbf{ReSTSA} & = \bigoplus(\mathcal{O}^{ReSSA}_1, \ldots, \mathcal{O}^{ReSSA}_{h_{ReSSA}}, \nonumber\\
    & \mathcal{O}^{ReDASA}_1, \ldots, \mathcal{O}^{ReDASA}_{h_{ReDASA}}, \mathcal{O}^{ReTSA}_1, \ldots, \mathcal{O}^{ReTSA}_{h_{ReTSA}}) \mathbf{\widehat W}.
\end{align*}
Here, $\bigoplus$ signifies the concatenation operation, and $\mathcal{O}^{ReSSA}_i$, $\mathcal{O}^{ReDASA}_i$, and $\mathcal{O}^{ReTSA}_i$ represent the outputs from the geographical, semantic, and temporal heads, respectively. 
$\mathbf{\widehat W} \in \mathbb{R}^{d \times d}$ is the projection matrix. We define the dimension \( d_0 \) as a function of the total number of heads in our enhanced rectified spatial-temporal self-attention model, calculated by dividing the original dimension \( d \) by the sum of the \textbf{ReSSA}, \textbf{ReDASA}, and \textbf{ReTSA} heads:
\begin{equation*}
    d_0 = \frac{d}{h_{ReSSA} + h_{ReDASA} + h_{ReTSA}}.
\end{equation*}
Furthermore, a position-wise fully connected feed-forward network is employed on the output of the multi-head self-attention block, resulting in outputs denoted by: 
\begin{equation*}
    \mathcal{Y} \in \mathbb{R}^{T \times N \times d}.
\end{equation*}

\paragraph{Output Layer:} For the final output layer, a skip connection with 1x1 convolutions is utilized after each spatial-temporal encoder layer. This process transforms the outputs \( \mathcal{Y}\) into a skip dimension \( \mathcal{Y}_{sk} \) within the space \( \mathbb{R}^{T \times N \times d_{sk}} \), where \( d_{sk} \) represents the skip dimension. The final hidden state \( \mathcal{Y}_{hid} \in \mathbb{R}^{T \times N \times d_{sk}} \) is then derived by aggregating the outputs from each skip connection layer. For multi-step forecasting, the output layer linearly transforms the final hidden state \( \mathbf{X}_{hid} \) into the required dimensions as:
\begin{equation*}
    \widehat{\mathcal{Y}} = \text{Conv2}(\text{Conv1}(\mathcal{Y}_{hid})),
\end{equation*}
where \( \widehat{\mathcal{Y}} \) in \( \mathbb{R}^{T_0 \times N \times d} \) represents the prediction results for \( h'\) steps, and both Conv1 and Conv2 are \(1 \times 1\) convolutions. This direct approach is preferred over recursive methods for multi-step prediction to minimize cumulative errors and enhance model efficiency.

\section{Experiments}
\label{Experiment}
Before the experimental description, we first introduce the datasets to be used in Section~\ref{Dataset}. This is followed by a description of the baseline models in Section~\ref{Base}. Detailed information about the experimental settings is provided in Section~\ref{Experim_set}. Subsequently, the experiment results are presented in Section~\ref{exp_results} and the evaluation metrics are described in Section~\ref{Evaluationmetrixc}. The results of the ablation study \footnote{An ablation study is a method of evaluating a system's performance by sequentially removing its components to identify their individual impacts on the overall effectiveness.} are discussed in Section~\ref{Ablation}, followed by a comprehensive discussion in Section~\ref{Discussion}.

\subsection{Dataset description}
\label{Dataset}
In our study, we evaluate \textbf{CCDSReFormer} using six real-world datasets in both graph-based and grid-based data structures. Specifically, the graph-based datasets include \texttt{PeMS04}, \texttt{PeMS07}, and \texttt{PeMS08}, while the grid-based datasets comprise \texttt{CHIBike}, \texttt{TDrive}, and \texttt{NYTaxi}. These datasets are publicly accessible and can be found in the LibCity repository, as referenced in Wang et al. (2023)\cite{wang2023libcity}. Detailed information about each of these datasets is provided in Table \ref{tab:datasets}.

\begin{table*}[t]
\centering
\caption{Dataset Information}
\footnotesize
\begin{tabular}{|c|c|c|c|c|c|}
\hline
Dataset & \#Nodes & \!\#Edges & \#Timestamps & Time Interval & Time Range \\
\hline
\texttt{PeMS04} & 307 & 340 & 16992 & 5 min & 01/01/2018 - 02/28/2018 \\
\texttt{PeMS07} & 883 & 866 & 28224 & 5 min & 05/01/2017 - 08/31/2017 \\
\texttt{PeMS08} & 170 & 295 & 17856 & 5 min & 07/01/2016 - 08/31/2016 \\
\texttt{NYCTaxi} & 75 (15x5) & 484 & 17520 & 30 min & 01/01/2014 - 12/31/2014 \\
\texttt{CHIBike} & 270 (15x18) & 1966 & 4416 & 30 min & 07/01/2020 - 09/30/2020 \\
\texttt{T-Drive} & 1024 (32x32) & 7812 & 3600 & 60 min & 02/01/2015 - 06/30/2015 \\
\hline
\end{tabular}
\label{tab:datasets}
\end{table*}

\textbf{\texttt{PeMS04}} \cite{song2020spatial}: Representing traffic data from the San Francisco Bay Area, this dataset was accumulated by the Caltrans Performance Measurement Systems (PeMS). Data from one sensor is condensed into 5-minute intervals, incorporating traffic flow, average speed, and average occupancy. It encompasses records from 307 sensors, spanning from Jan 1, 2018, to Feb 28, 2018.

\textbf{\texttt{PeMS07}} \cite{song2020spatial}: Representing traffic data from the San Francisco Bay Area, this dataset was accumulated by the Caltrans Performance Measurement Systems (PeMS). Data from one sensor is condensed into 5-minute intervals, incorporating traffic flow, average speed, and average occupancy. It encompasses records from 883 sensors, spanning from May 1, 2017, to Aug 31, 2017.

\textbf{\texttt{PeMS08} }\cite{song2020spatial}: This is the highway traffic flow data collected by the California Department of Transportation (Caltrans). The
data range is from Jul 1, 2016 to Aug 31, 2016. The flow data is sampled every 5 minutes. The number of sensors is 170. 

\textbf{\texttt{NYCtaxi}} \cite{liu2020dynamic}: The data was made available by the \texttt{NYCtaxi} \& Limousine Commission (TLC) and built on data from ride-hailing companies such as Uber and Lyft. The records cover New York from Jan 1, 2014 to Dec 31, 2014. For each demand record, the data provides information such as the pick-up time, drop-off time, pick-up zone, drop-off zone, etc. The traffic zones are predetermined by
the TLC. 

\textbf{\texttt{CHIBike}} \cite{wang2023libcity}: The \texttt{CHIBike} dataset comprises bicycle-sharing data from Chicago, capturing the period from Jul 1, 2020, to Sep 30, 2019. This dataset includes detailed records of bike rentals and returns across various stations in Chicago. Each record encompasses information such as rental and return times, originating and destination stations, and trip duration.

\textbf{\texttt{T-Drive}} \cite{pan2019urban}: The \texttt{T-Drive} dataset is derived from a comprehensive collection of taxi trajectory data in Beijing, spanning one week of continuous operation. It contains over 15 million GPS records from thousands of taxis, providing detailed insights into urban traffic flow. The data encapsulates information such as GPS coordinates, timestamps, and taxi operation statuses (e.g., occupied, vacant).

\subsection{Tested frameworks and baseline methods}\label{Base}
We compare \textbf{CCDSReFormer} with the following 11 baselines in three categories. 

\noindent(1) Graph Neural Network based Models:
\begin{itemize}
\item DCRNN \cite{li2018dcrnn}: Is a deep learning framework for traffic forecasting that addresses the complexities of spatial dependencies on road networks and non-linear temporal dynamics, using bidirectional random walks and an encoder-decoder architecture.
    \item STGCN \cite{yu2018stgcn}: Is a novel deep learning framework for traffic time series prediction, utilizing complete convolutional structures on graphs for faster training and fewer parameters, effectively capturing spatio-temporal correlations and outperforming baselines on various real-world traffic datasets.
    \item GWNET \cite{wu2019gwnet}: Is a novel graph neural network for spatial-temporal graph modeling, which overcomes the limitations of fixed graph structures and ineffective temporal trend capture in existing methods by using an adaptive dependency matrix and stacked dilated 1D convolutions.
    \item MTGNN \cite{wu2020mtgnn}: Introduces a general graph neural network framework tailored for multivariate time series data, which automatically extracts uni-directed relations among variables through a graph learning module and incorporates novel mix-hop propagation and dilated inception layers.
    \item STFGNN \cite{li2021stfgnn}: Is a model that effectively learns hidden spatial-temporal dependencies through a novel fusion of various spatial and temporal graphs, and integrates a gated convolution module for handling long sequences.
    \item STGNCDE \cite{choi2022stgncde}: Is a breakthrough method  in traffic forecasting, combining graph convolutional networks and recurrent neural networks with neural controlled differential equations for spatial and temporal processing.
\end{itemize}

\noindent(2) Self-Attention based Models: 
\begin{itemize}
    \item STTN \cite{xu2020sttn}: Is a novel approach for long-term traffic forecasting that dynamically models directed spatial dependencies using a spatial transformer and long-range temporal dependencies using a temporal transformer, offering enhanced accuracy and scalable training. 
    \item GMAN \cite{zheng2020gman}: Is a long-term traffic prediction model, utilizing an encoder-decoder architecture with spatio-temporal attention blocks to model the impact of spatio-temporal factors on traffic conditions, featuring a transform attention layer that effectively links historical and future time steps, and demonstrating superior performance in real-world traffic volume and speed prediction tasks.
    \item ASTGNN \cite{guo2021astgnn}: Is a novel spatial-temporal neural network framework integrating a graph convolutional recurrent module (GCRN) with a global attention module, designed to effectively model both long-term and short-term temporal correlations in traffic data, and demonstrating superior predictive performance on five real traffic datasets compared to baseline methods. 
    \item PDFormer \cite{jiang2023pdformer}:  Uses self-attention to capture dynamic spatial dependencies and explicitly model time delays in traffic systems, demonstrating state-of-the-art performance.
    \item STAEformer \cite{liu2023staeformer}: A state of art transformer model that leveraging Spatio-Temporal Adaptive Embedding, achieves top performance in traffic forecasting by effectively capturing complex spatio-temporal patterns, marking a significant advance over previous models.
    
\end{itemize} 

\noindent(3) Other based Models: 
 \begin{itemize}
    \item VAR \cite{Hamilton1994}: Vector Auto-Regression used in traffic flow prediction for its ability to capture the linear interdependencies among multiple time series, making it suitable for forecasting traffic conditions based on historical data.
     \item SVR \cite{Drucker1997}: Support Vector Regression utilizes historical data to predict future traffic conditions by employing a linear kernel function, offering reliable forecasts even with high-dimensional data.
 \end{itemize}
\subsection{Experiment Settings}
\label{Experim_set}
\paragraph{Data Processing} In line with contemporary practices of all the baselines \cite{jiang2023pdformer}, we partition the three graph-based datasets into training, validation, and test sets using a 6:2:2 split. For these datasets, we predict traffic flow over the next hour (12 time steps) based on the data from the preceding hour (12 time steps), thus employing a multi-step prediction approach. For the grid-based datasets, we adopt a 7:1:2 split ratio which is aligned with the baseline models. In this case, the prediction model uses data from the previous six time steps to forecast the next step's traffic inflow and outflow. Before training, we standardize the inputs across all datasets by applying Z-score normalization. The code is available in: 
 \textcolor{blue}{\textit{https://github.com/superca729/CCDSReFormer}}.

\paragraph{Parameter Settings}
All experiments are conducted on a machine with the GPU RTX 3090(24GB) CPU 15 vCPU Intel(R) Xeon(R) Platinum 8358P CPU @ 2.60GHz memory
150GB. We use the same configurations with \cite{jiang2023pdformer}  for the hidden dimension $d$, tuning its values within\{16, 32, 64\}, and assess different depths for the encoder layers $L$ with an option set \{2, 4, 6, 8\}. Unlike the baselines, we fix the same number of attention heads to \{2, 3\} for \(h_{ReSSA}\) and \(h_{ReTSA}\), respectively, to ensure the Criss-Cross performance well. We set the number of \(h_{ReDASA}\) in 4. The selection of the optimal model configuration was based on their performance on the validation set. For training a model, we employ the same    Adam optimizer as \cite{jiang2023pdformer} with a learning rate of 0.001. The models were trained with a batch size of 16 over 200 epochs.

\subsection{Evaluation Metrics}
\label{Evaluationmetrixc}
Three common metrics, the mean absolute error (MAE), the mean absolute percentage error (MAPE) and the root mean square error (RMSE); are used to measure the traffic forecasting performance of the tested methods. 
\begin{equation*}\text{MAE} = \frac{1}{n} \sum_{i=1}^{n} |\hat{y}_i - y_i|,\end{equation*}
\begin{equation*}\text{MAPE} = \frac{1}{n} \sum_{i=1}^{n} \left| \frac{\hat{y}_i - y_i}{y_i} \right| \times 100\%, 
\end{equation*}
\begin{equation*}\text{RMSE} = \sqrt{\frac{1}{n} \sum_{i=1}^{n} (\hat{y}_i - y_i)^2} 
\end{equation*} 
where $y = \{y_1, y_2, \ldots, y_n\}$ is the ground-truth value, $\hat{y} = \{\hat{y}_1, \hat{y}_2, \ldots, \hat{y}_{n}\}$ is the prediction value. 

In our evaluations, we exclude missing values when calculating metrics. For grid-based datasets, samples with flow values below 10 were filtered out, as the method described in \cite{yao2018dmvstnet,jiang2023pdformer}. Additionally, for these datasets, we compute the final result by taking the average of the inflow and outflow evaluation metrics which is the same as work \cite{jiang2023pdformer}. 

\subsection{Experiment Results}
\label{exp_results}

\begin{figure}[t]
    \centering
    \includegraphics[width=0.45\textwidth]{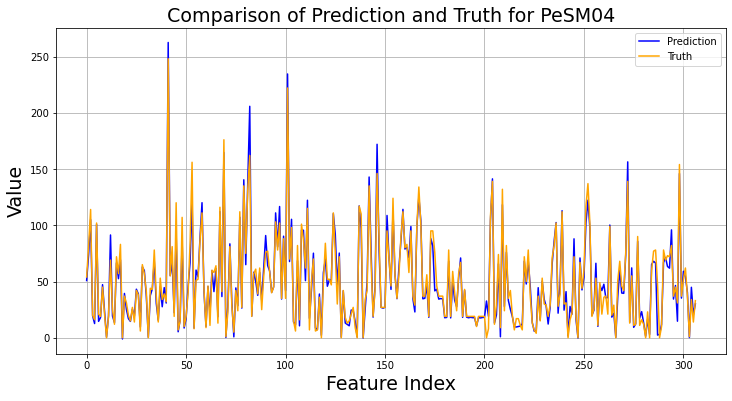}
    \caption{The plot illustrates a comparison between the actual data (ground truth) and the predicted values for the \texttt{PeMS04} dataset using the \textbf{CCDSReFormer} model. The data points are indexed along the x-axis, which represents as time.}
    \label{PeMS04_true}
\end{figure}

\begin{figure}[t]
    \centering
    \includegraphics[width=0.4\textwidth]{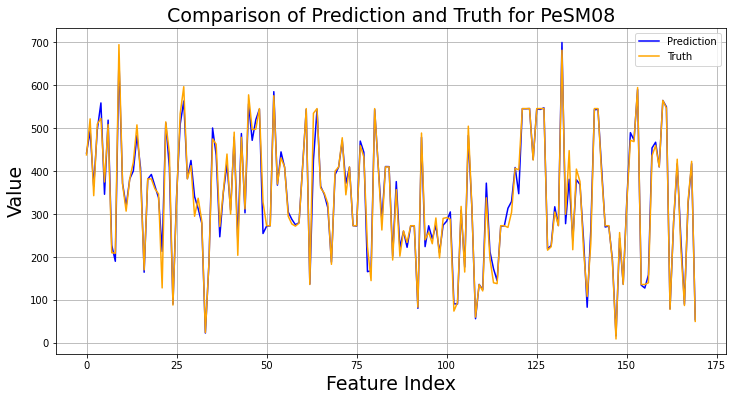}
    \caption{The plot illustrates a comparison between the actual data (ground truth) and the predicted values for the \texttt{PeMS08} dataset using the \textbf{CCDSReFormer} model.  The data points are indexed along the x-axis, which represents as time.}
    \label{PeMS08_true}
\end{figure}

\begin{figure}[t]
    \centering
    \includegraphics[width=0.4\textwidth]{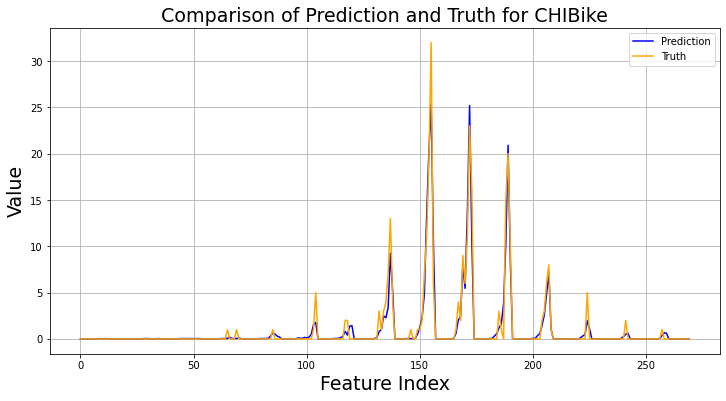}
    \caption{The plot illustrates a comparison between the actual data (ground truth) and the predicted values for the \texttt{CHIBike} dataset using the \textbf{CCDSReFormer} model.}
    \label{CHIBike_true}
\end{figure}

\begin{figure}[t]
    \centering
    \includegraphics[width=0.4\textwidth]{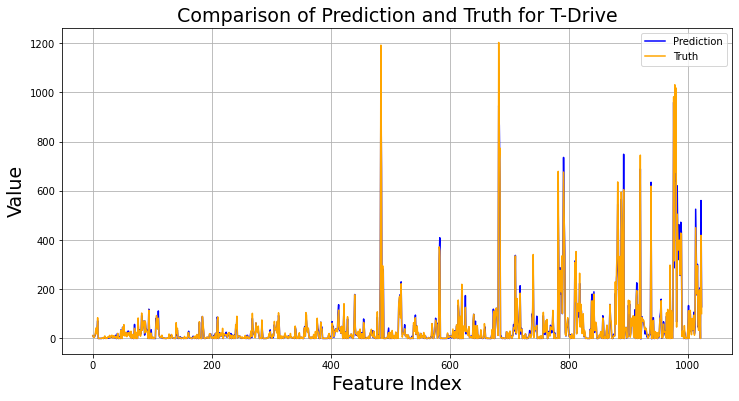}
    \caption{The plot illustrates a comparison between the actual data (ground truth) and the predicted values for the \texttt{TDrive} dataset using the \textbf{CCDSReFormer} model.  The data points are indexed along the x-axis, which represents as time.}
    \label{TDrive_true}
\end{figure}

\begin{figure}[t]
    \centering
    \includegraphics[width=0.4\textwidth]{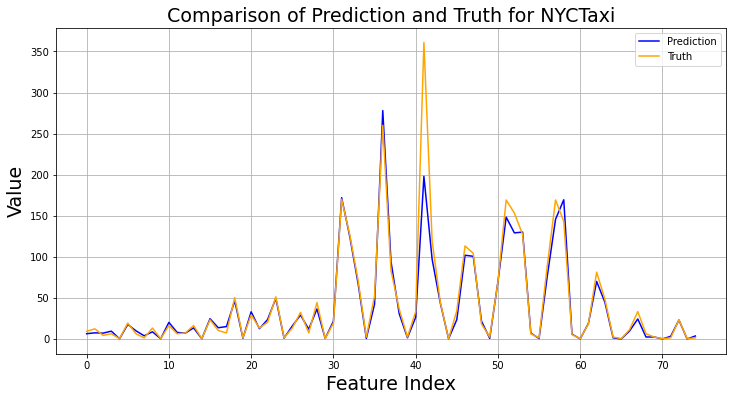}
    \caption{The plot illustrates a comparison between the actual data (ground truth) and the predicted values for the \texttt{NYTaxi} dataset using the \textbf{CCDSReFormer} model.  The data points are indexed along the x-axis, which represents as time.}
    \label{NYTaxi_true}
\end{figure}

The performance results on two types of datasets are presented in Table \ref{tab:graph_based_results} and \ref{tab:grid_based_results}, where the best results are marked in \textcolor{red}{\textbf{First}}, \textcolor{blue}{\textbf{Second}} and \textcolor{purple}{\textbf{Third}}. To facilitate a straightforward comparison with the baseline models, we have presented the results in three decimal places, consistent with the precision used for the baseline models. We also further present a visualization comparing the prediction and ground truth of \textbf{CCDSReFormer} showing as in Figures~\ref{PeMS04_true}, \ref{PeMS08_true}, \ref{CHIBike_true}, \ref{TDrive_true} and \ref{NYTaxi_true}.

\paragraph{Performance Results on Various Datasets} 
To ensure an equitable assessment of model efficacy, we have chosen our baseline models from among the leading contenders in Traffic Flow prediction as identified at the time of drafting this paper. We include comparisons with the current state-of-the-art (SOTA) models (eg. STAEFormer\cite{liu2023staeformer}) with the time when we worked on the paper.

Based on the results presented in Tables \ref{tab:graph_based_results} and \ref{tab:grid_based_results}, our \textbf{CCDSReFormer} demonstrates superior performance over all baseline models in terms of MAE (Mean Absolute Error) and RMSE (Root Mean Squared Error) across all datasets, as confirmed by the Student’s t-test at a significance level of 0.01. For the \texttt{PeMS04} dataset, \textbf{CCDSReFormer} outperforms other models with the lowest MAE of 18.176, the lowest MAPE of 12.096\%, and the lowest RMSE of 29.844. For the \texttt{PeMS07} dataset, \textbf{CCDSReFormer} again shows superior performance with the lowest MAE of 19.475, although it had a higher MAPE (32.473\%) compared to PDFormer.  While \textbf{CCDSReFormer} still achieves the best MAPE of 8.529\%. The RMSE was recorded as the lowest for \textbf{CCDSReFormer} at 32.499. For the \texttt{PeMS08} dataset, both \textbf{CCDSReFormer} and PDFormer are comparable 
in MAE and MAPE metrics. Specifically, \textbf{CCDSReFormer} achieves the best MAE of 13.564 and the best RMSE of 23.250, while PDFormer records the best MAPE of 9.046\% . Consistent with previous observations, \textbf{CCDSReFormer} outperforms in MAE and RMSE metrics. However, its performance in MAPE was not strong enough to compete with PDFormer. This pattern suggests that while \textbf{CCDSReFormer} is generally reliable in terms of absolute error, it might struggle with relative error, especially in datasets where small actual values are more prevalent, thus leading to higher MAPE scores. 

\paragraph{Visualization of CCDSReFormer Performance} We further present visualization to compare the ground truth and the prediction based on the \textbf{CCDSReFormer} model on the datasets in Figure~\ref{PeMS04_true}, \ref{PeMS08_true}, \ref{CHIBike_true}, \ref{TDrive_true} and \ref{NYTaxi_true}, respectively. The comparison between the ground truth (depicted in tan) and predicted values (shown in blue) reveals key insights into the model's performance. Notably, the predicted values generally aligned well with the actual data, indicating the model's capability to capture overall trends. However, specific deviations were observed: an overestimation in the \texttt{PeMS04} (Figure~\ref{PeMS04_true}), and \texttt{TDrive} (Figure~\ref{TDrive_true}) datasets, where the blue lines rose above the tan ground truth lines, and a slight underestimation in the \texttt{PeSM08} (Figure~\ref{PeMS08_true}) \texttt{CHIBike} (Figure~\ref{CHIBike_true}) and \texttt{NYTaxi} (Figure~\ref{NYTaxi_true}) dataset, as indicated by the blue line falling just slightly below the actual values. 

\begin{table*}[t]
\centering
\caption{Performance metrics for different models across graph datasets, with ranking highlights.}
\footnotesize
\label{tab:graph_based_results}
\begin{tabular}{@{}lp{1.1cm}p{1.1cm}p{1.1cm}|p{1.1cm}p{1.1cm}p{1.1cm}|p{1.1cm}p{1.1cm}p{1.1cm}@{}}
\toprule
\textbf{Models} & \multicolumn{3}{c}{\textbf{\texttt{PeMS04}}} & \multicolumn{3}{c}{\textbf{\texttt{PeMS07}}} & \multicolumn{3}{c}{\textbf{\texttt{PeMS08}}}\\
\midrule
 & MAE & MAPE & RMSE & MAE & MAPE & RMSE & MAE & MAPE & RMSE \\
\midrule
VAR & 23.750 & 18.090 & 36.660 & 101.200 & 39.690 & 155.140 & 22.320 & 14.470 & 33.830 \\
SVR & 28.660 & 19.150 & 44.590 & 32.970 & 15.430 & 50.150 & 23.250 & 14.710 & 36.150 \\
\hline
DCRNN         & 22.737 & 14.751 & 36.575 & 23.634 & 12.281 & 36.514 & 18.185 & 11.235 & 28.176 \\
STGCN         & 21.758 & 13.874 & 34.769 & 22.898 & 11.983 & 35.440 & 17.838 & 11.211 & 27.122 \\
STFGNN        & 19.830 & 13.021 & 31.870 & 22.072 & 9.212  & 35.805 & 16.636 & 10.547 & 26.206 \\
STGNCDE       & 19.211 & 12.772 & 31.088 & 20.620 & 8.864  & 34.036 & 15.455 & 9.921  & 24.813 \\
STTN          & 19.478 & 13.631 & 31.910 & 21.344 & 9.932  & 34.588 & 15.482& 10.341 & 24.965 \\
GMAN          & 19.139 & 13.192 & 31.601 & 20.967 & 9.052  & 34.097 & 15.307 & 10.134 & 24.915 \\
ASTGNN        & 18.601 & 12.630 & 31.028 & 20.616 & \textbf{\textcolor{purple}{8.861}}  & 34.017 & 14.974 & 9.489  & 24.710 \\
PDFormer      & \textbf{\textcolor{purple}{18.321}} & \textbf{\textcolor{blue}{12.103}} & \textbf{\textcolor{blue}{29.965}} & \textbf{\textcolor{purple}{19.832}}& \textbf{\textcolor{blue}{8.012}}  & \textbf{\textcolor{purple}{32.870}} & \textbf{\textcolor{purple}{13.583}} & \textbf{\textcolor{red}{9.046}} & \textbf{\textcolor{purple}{23.505}} \\
STAEFormer & \textbf{\textcolor{blue}{18.224}} & \textbf{\textcolor{purple}{12.301}}  & \textbf{\textcolor{purple}{30.832}}& \textbf{\textcolor{blue}{19.343}} & \textbf{\textcolor{red}{8.012}}& \textbf{\textcolor{blue}{32.603}} &  \textbf{\textcolor{blue}{13.462}} & \textbf{\textcolor{blue}{8.889}} & \textbf{\textcolor{blue}{23.254}} \\
\hline
CCDSReFormer  & \textbf{\textcolor{red}{18.176}} & \textbf{\textcolor{red}{12.096}} & \textbf{\textcolor{red}{29.544}} & \textbf{\textcolor{red}{19.305}} & 32.473 & \textbf{\textcolor{red}{32.499}} & \textbf{\textcolor{red}{13.424}} & \textbf{\textcolor{purple}{9.067}} & \textbf{\textcolor{red}{23.250}} \\
\bottomrule
\end{tabular}
\end{table*}

\begin{table*}[t]
\centering
\caption{Performance metrics for different models across grid datasets.}
\footnotesize
\label{tab:grid_based_results}
\begin{tabular}{@{}lp{1.1cm}p{1.1cm}p{1.1cm}|p{1.1cm}p{1.1cm}p{1.1cm}|p{1.1cm}p{1.1cm}p{1.1cm}@{}}
\toprule
\textbf{Models} & \multicolumn{3}{c}{\textbf{\texttt{CHIBike}}} & \multicolumn{3}{c}{\textbf{\texttt{TDrive}}} & \multicolumn{3}{c}{\textbf{\texttt{NYTaxi}}} \\
\midrule
 & MAE & MAPE & RMSE & MAE & MAPE & RMSE & MAE & MAPE & RMSE \\
\midrule
DCRNN & 22.737 & 14.751 & 36.575 & 23.634 & 12.281 & 36.514 & 18.185 & \textcolor{purple}{\textbf{11.235}} & 28.176  \\
STGCN & 21.758 & 13.874 & 34.769 & 22.898 & 11.983 & 35.440 & 17.838 & \textcolor{blue}{\textbf{11.211}} & 27.122 \\
STFGNN & 21.938 & 17.566 & 38.411 & 21.143 & 17.261 & 37.836 & 19.553 & 16.560 & 36.179  \\
STGNCDE & 19.211 & 18.601 & 31.088 & 19.478 & 12.772 & 31.910 & 19.139 & 13.631 & 31.601 \\
STTN & 20.620 & \textcolor{red}{\textbf{8.864}} & 34.036 & 21.344 & \textcolor{red}{\textbf{9.932}} & 34.588 & 20.967 & \textcolor{red}{\textbf{10.134}} & 34.097 \\
GMAN & \textcolor{purple}{\textbf{15.455}} & \textcolor{blue}{\textbf{9.921}} & \textcolor{purple}{\textbf{24.813}} & \textcolor{purple}{\textbf{15.482}}& \textcolor{blue}{\textbf{10.341}} & \textcolor{blue}{\textbf{24.965}} & \textcolor{purple}{\textbf{15.307}} & \textcolor{red}{\textbf{10.134}} & \textcolor{purple}{\textbf{24.915}}  \\
ASTGNN & \textcolor{blue}{\textbf{13.279}} & 13.926 & \textcolor{blue}{\textbf{21.675}} & \textcolor{blue}{\textbf{13.366}}& \textcolor{purple}{\textbf{13.984}}& \textcolor{red}{\textbf{21.834}} & \textcolor{blue}{\textbf{13.270}} & 13.893 & \textcolor{blue}{\textbf{21.661}}  \\
PDFormer & 19.289 & 16.504 & 36.118 & 20.513 & 16.659 & 37.143 & 19.104 & 16.449 & 36.053  \\

\hline
\textbf{CCDSReFormer} &  \textcolor{red}{\textbf{11.572}}& \textcolor{purple}{\textbf{12.751}} & \textcolor{red}{\textbf{18.359}} & \textcolor{red}{\textbf{12.180}} & 20.89 & \textcolor{purple}{\textbf{23.668}} & \textcolor{red}{\textbf{3.786}} & 29.693 & \textcolor{red}{\textbf{5.328}}\\

\bottomrule
\end{tabular}
\end{table*}

\subsection{Ablation Study}
\label{Ablation}
To further investigate the efficacy of \textbf{CCDS}, \textbf{EnCov}, and \textbf{ReLSA} components in the \textbf{CCDSReFormer}, we conduct an ablation study on the \texttt{PeMS04} dataset as showing in Table~\ref{table:ablation}, examining different variants of the proposed model. We initially reproduced the model and with adding of each component, we obtain a better performance of the model 

This study primarily focuses on key performance metrics: MAE, MAPE, and RMSE. The original PDFormer, as detailed in the paper\cite{jiang2023pdformer}, is used as a baseline. Our reproduced version exhibites slightly higher values in MAE, MAPE, and RMSE. Variants incorporating additional components specifically \textbf{CCDS}, \textbf{EnCov}, and \textbf{ReLSA} show diverse improvements. Notably, the \textbf{CCDSReFormer}, integrating all three of these components, achieves the most superior performance across all metrics. In particular, the \textbf{CCDSReFormer} variant excelled, recording the lowest MAE (18.176), MAPE (12.096\%), and RMSE (29.844). These results underscore the synergistic effect of combining \textbf{CCDS}, \textbf{EnCov}, and \textbf{ReLSA}, significantly enhancing the model's predictive accuracy on the \texttt{PeMS04} dataset.

\begin{table}[ht]
    \centering
    \caption{Ablation Study for \texttt{PeMS04} Dataset}
    \label{table:ablation}
    \begin{tabular}{l|c|c|c}
    \hline
    Models & MAE & MAPE & RMSE \\
    \hline
\textbf{CCDSReFormer} & \textbf{18.176} & \textbf{12.096} & \textbf{29.844} \\
    \textbf{w/o} ReLSA & 18.332 & 12.472 & 29.948 \\
    \textbf{w/o} EnCov & 18.269 & 12.121 & 30.022 \\
    \textbf{w/o} CCDS & 18.355 & 12.217 & 30.055 \\
    \hline
    \end{tabular}
\end{table}


To verify the effectiveness on \textbf{ReLSA}, we further test the average training time on \textbf{CCDSReFormer}, and ensure all models operated on the same device. Based on the result shown in Table~\ref{fig:runingtime}, our model showcases a shorter running time compared to PDFormer, ASTGNN, and GMAN, indicating superior performance efficiency. It also remains competitive with STTN, underscoring its effectiveness. Given that the STAEFormer has fewer parameters, it's reasonable to expect better performance in terms of both training and inference times. 

\begin{table}[htbp]
\centering
\caption{Model performance metrics on the PeMS04 dataset}
\label{fig:runingtime}
\begin{tabular}{@{}lcccc@{}}
\toprule
Model & \multicolumn{2}{c}{Training} & \multicolumn{2}{c}{Inference} \\ 
\cmidrule(r){2-3} \cmidrule(l){4-5}
 & Time (s) & & Time (s) & \\
\midrule
GMAN     & 493.578 & & 39.824 & \\
ASTGNN   & 205.223 & & 50.112 & \\
PDFormer & 131.871 & & 8.420 & \\
CCDSReFormer & 112.73 & & 7.871 \\
STTN     & 100.398 & & 12.596 & \\
STAEFormer  & 83.099  & & 7.156 & \\
\bottomrule
\end{tabular}
\end{table}

\newpage
\subsection{Discussion/Case Study}
\label{Discussion}
\captionsetup[subfloat]{font=scriptsize}   
\begin{figure*}[htbp]
  \centering
  \caption{Attention scores in \textbf{ReSSA}: horizontal axis for stage and heads growth, vertical axis for layer depth. Note: H$n$ = Head $n$, L$n$ = Layer $n$}\label{fig:attentionmap_ressa}
  \subfloat[Stage 1: H1, L1]{%
\includegraphics[width=0.24\textwidth]{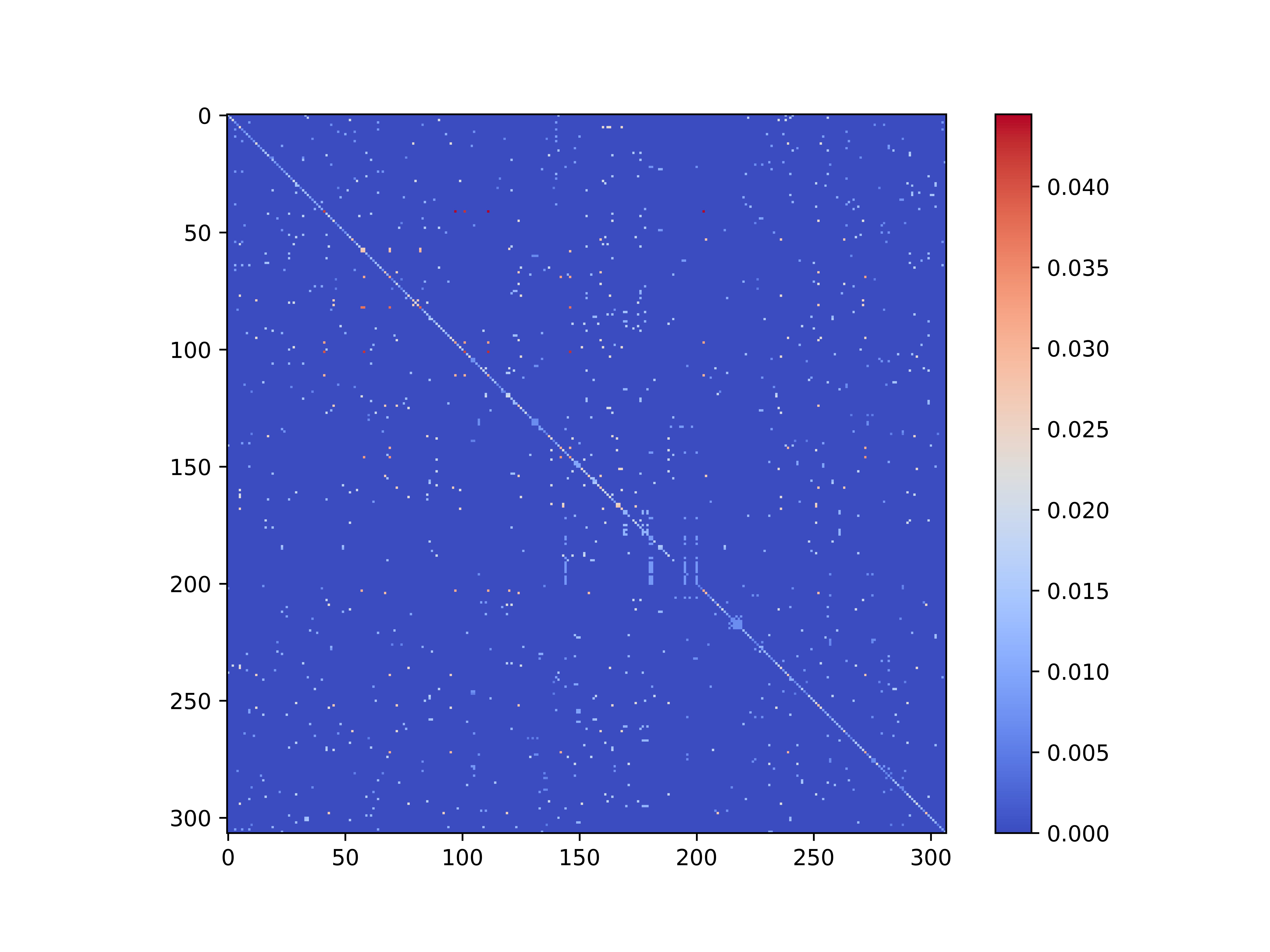}
    \label{fig:sub1}
  }
  \hfill
  \subfloat[Stage 1: H2, L1]{%
  \includegraphics[width=0.24\textwidth]{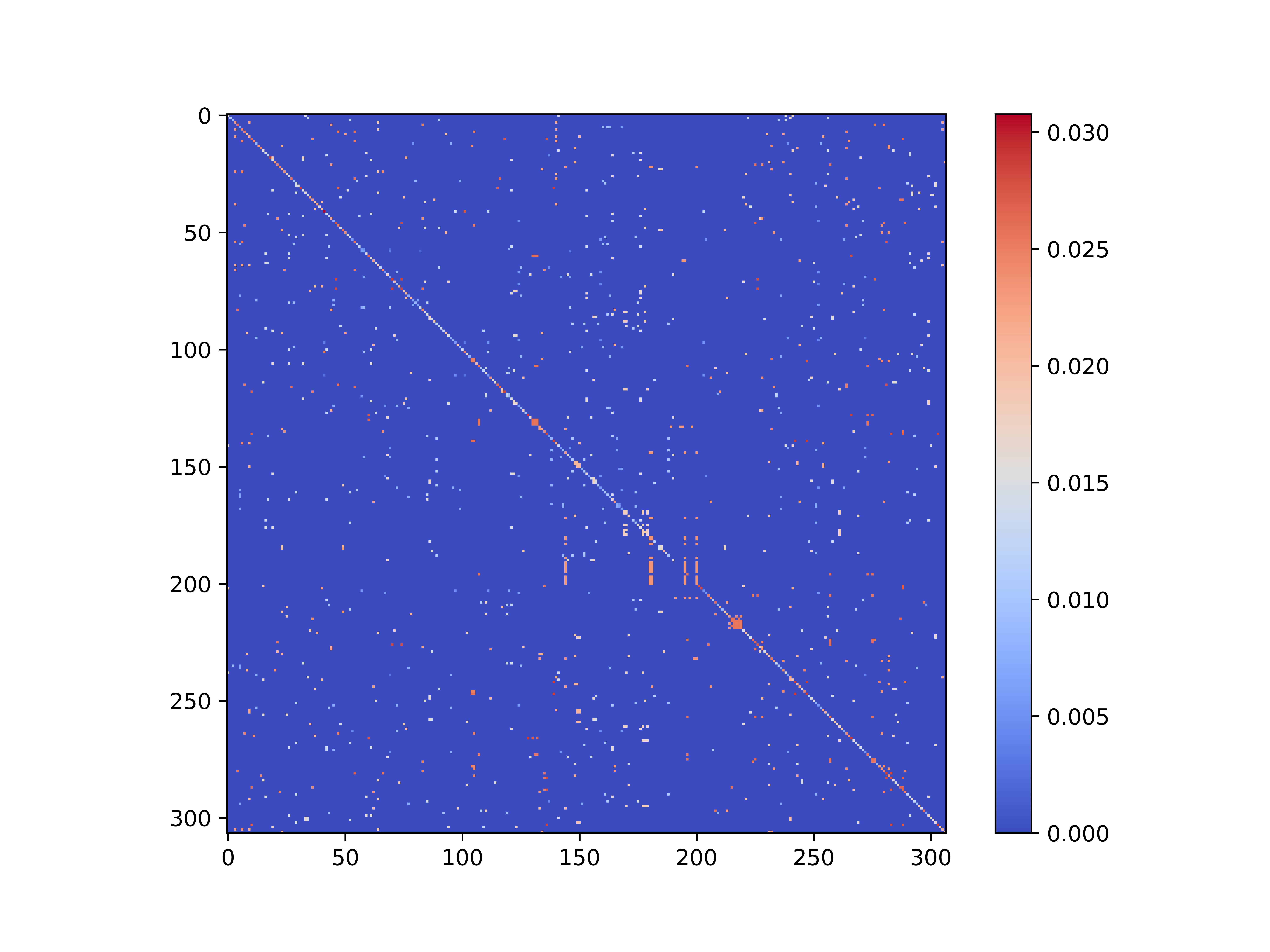}
    \label{fig:sub2}
  }
  \hfill
  \subfloat[Stage 2: H1, L1]{%
    \includegraphics[width=0.24\textwidth]{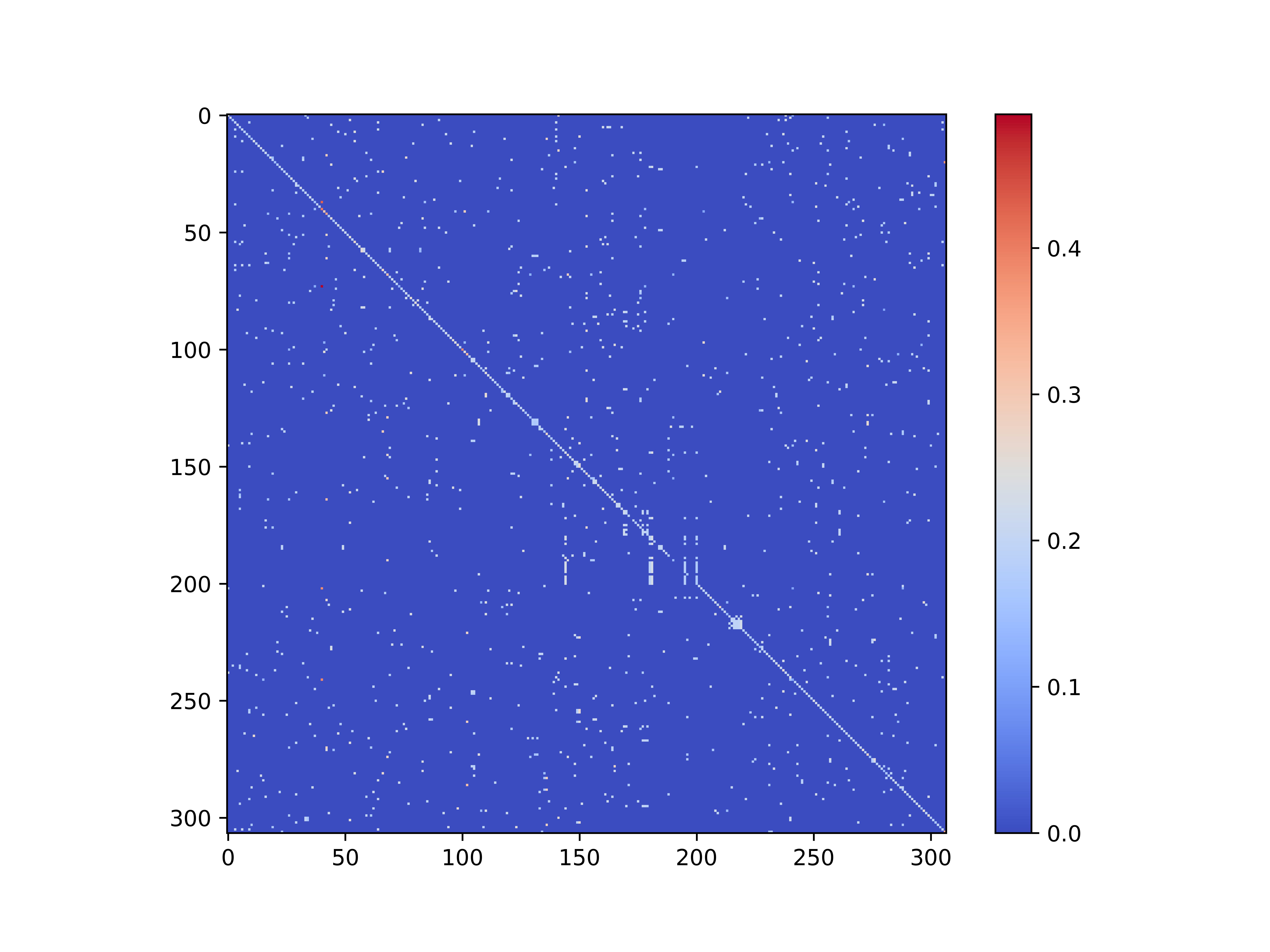}
    \label{fig:sub4}
  }
  \hfill
  \subfloat[Stage 2: H2, L1]{%
    \includegraphics[width=0.24\textwidth]{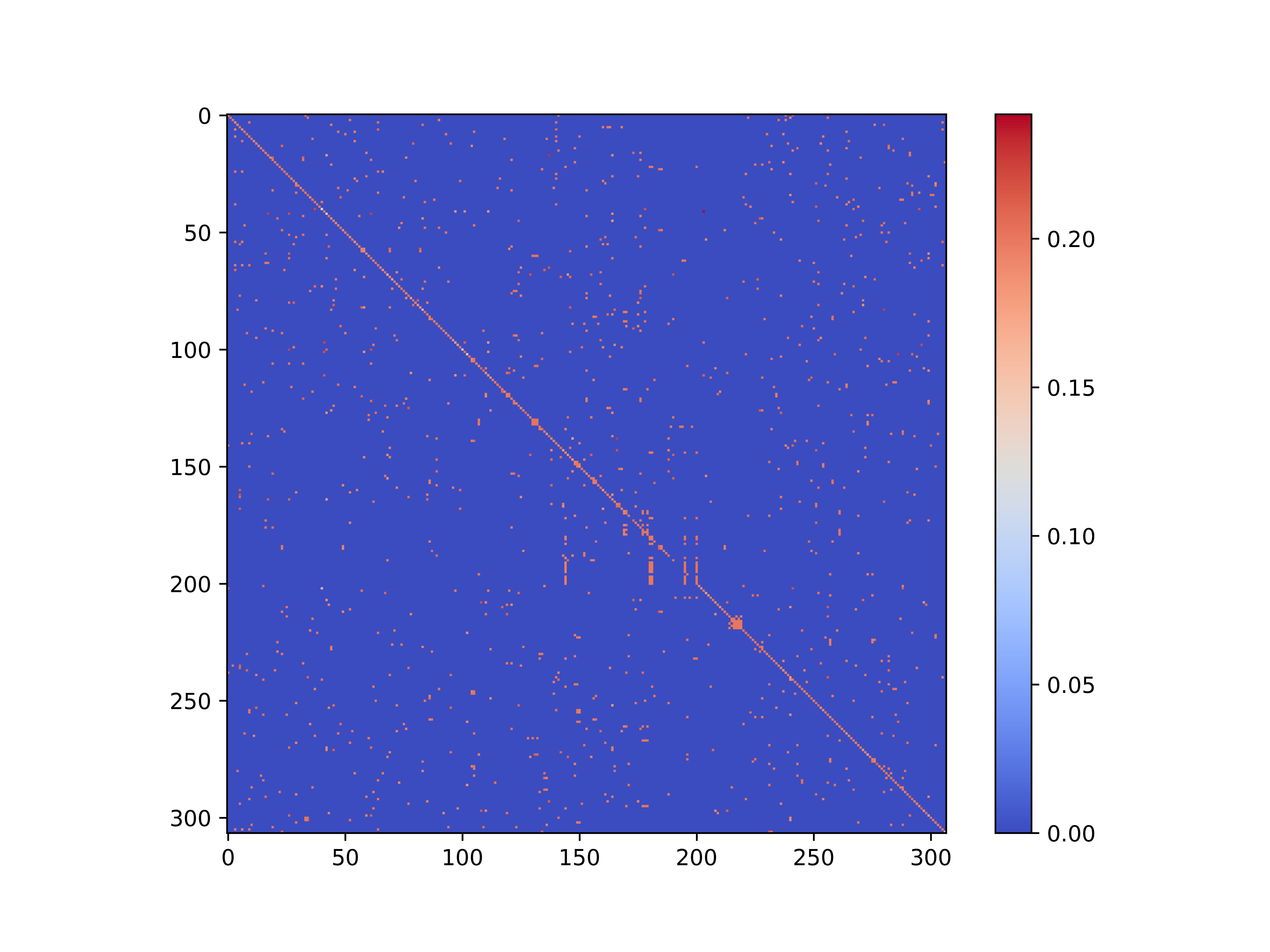}
    \label{fig:sub5}
  } \\[-6pt]
  \subfloat[Stage 1: H1, L2]{%
    \includegraphics[width=0.24\textwidth]{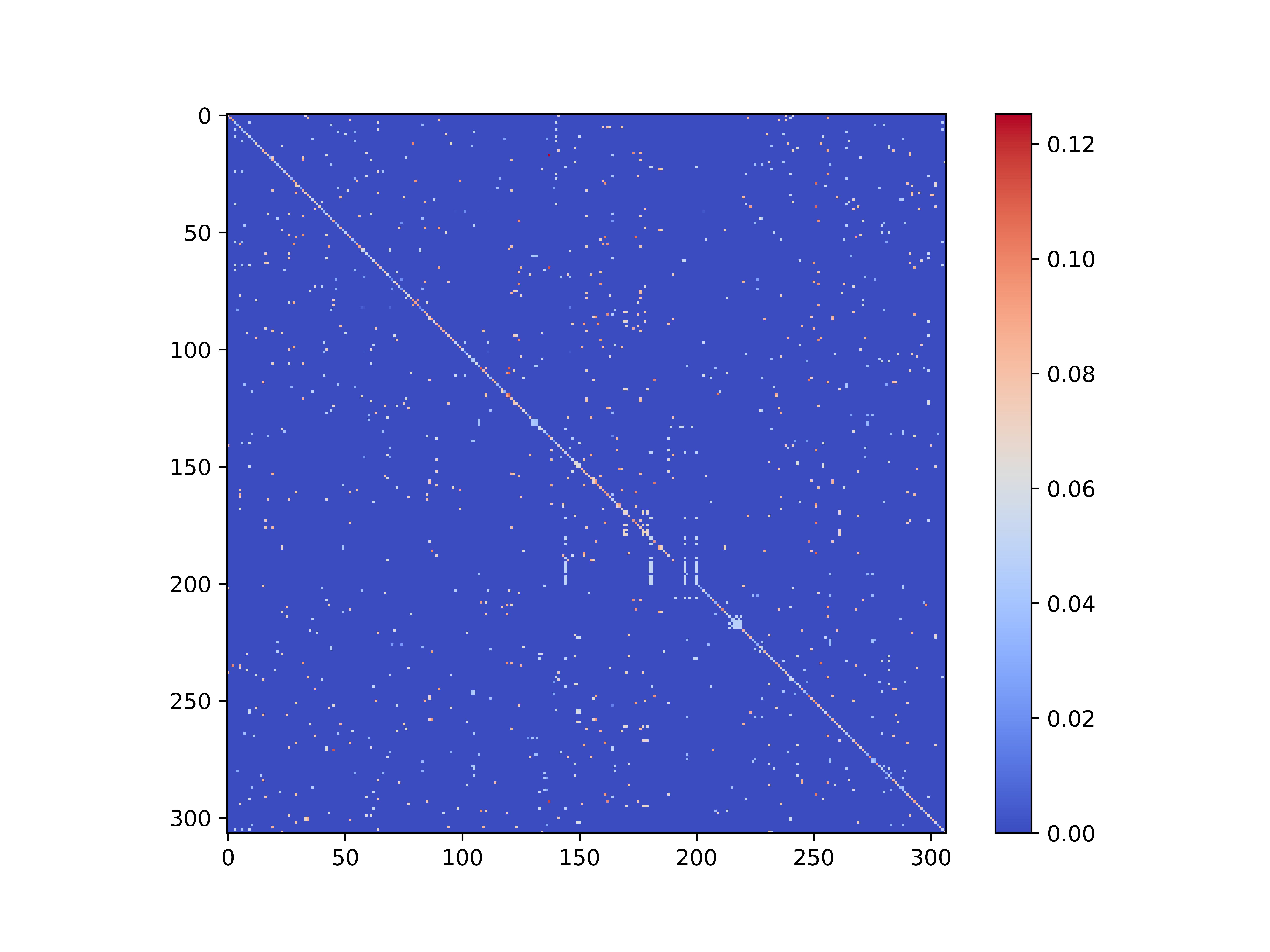}
    \label{fig:sub6}
  }
  \hfill
  \subfloat[Stage 1: H2, L2]{%
    \includegraphics[width=0.24\textwidth]{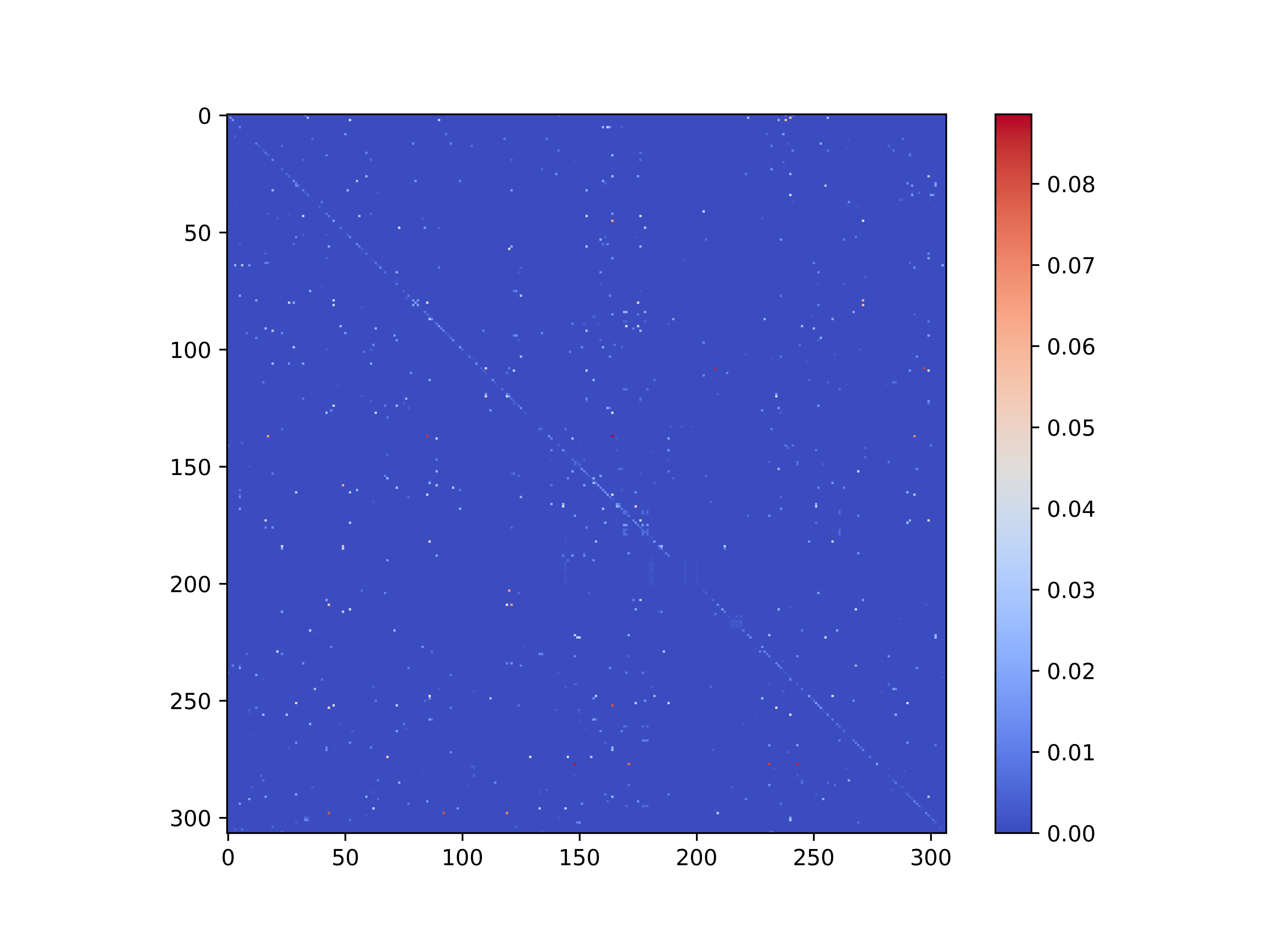}
    \label{fig:sub7}
  }
  \hfill
  \subfloat[Stage 2: H1, L2]{%
    \includegraphics[width=0.24\textwidth]{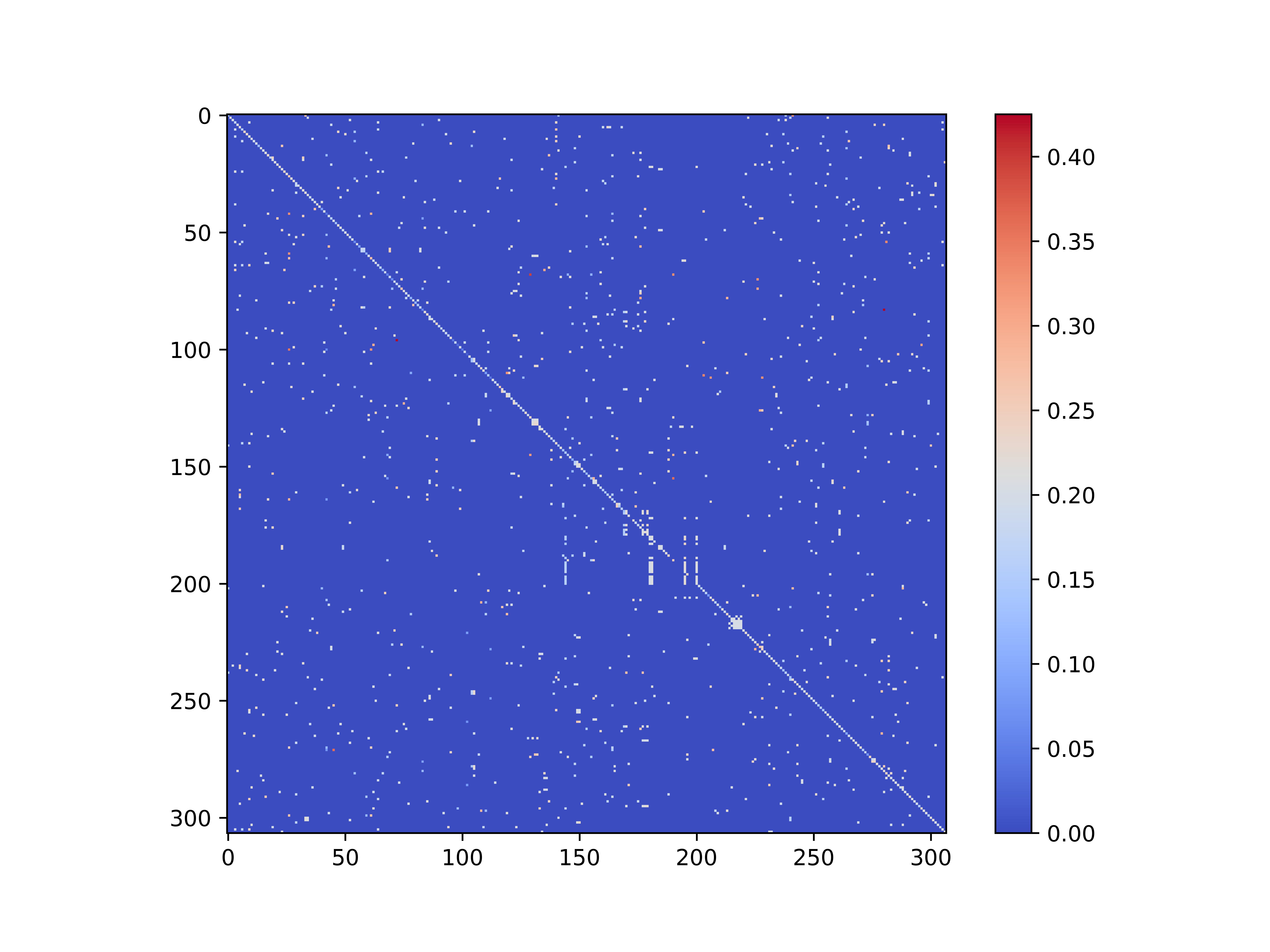}
    \label{fig:sub8}
  }
  \hfill
  \subfloat[Stage 2: H2, L2]{%
    \includegraphics[width=0.24\textwidth]{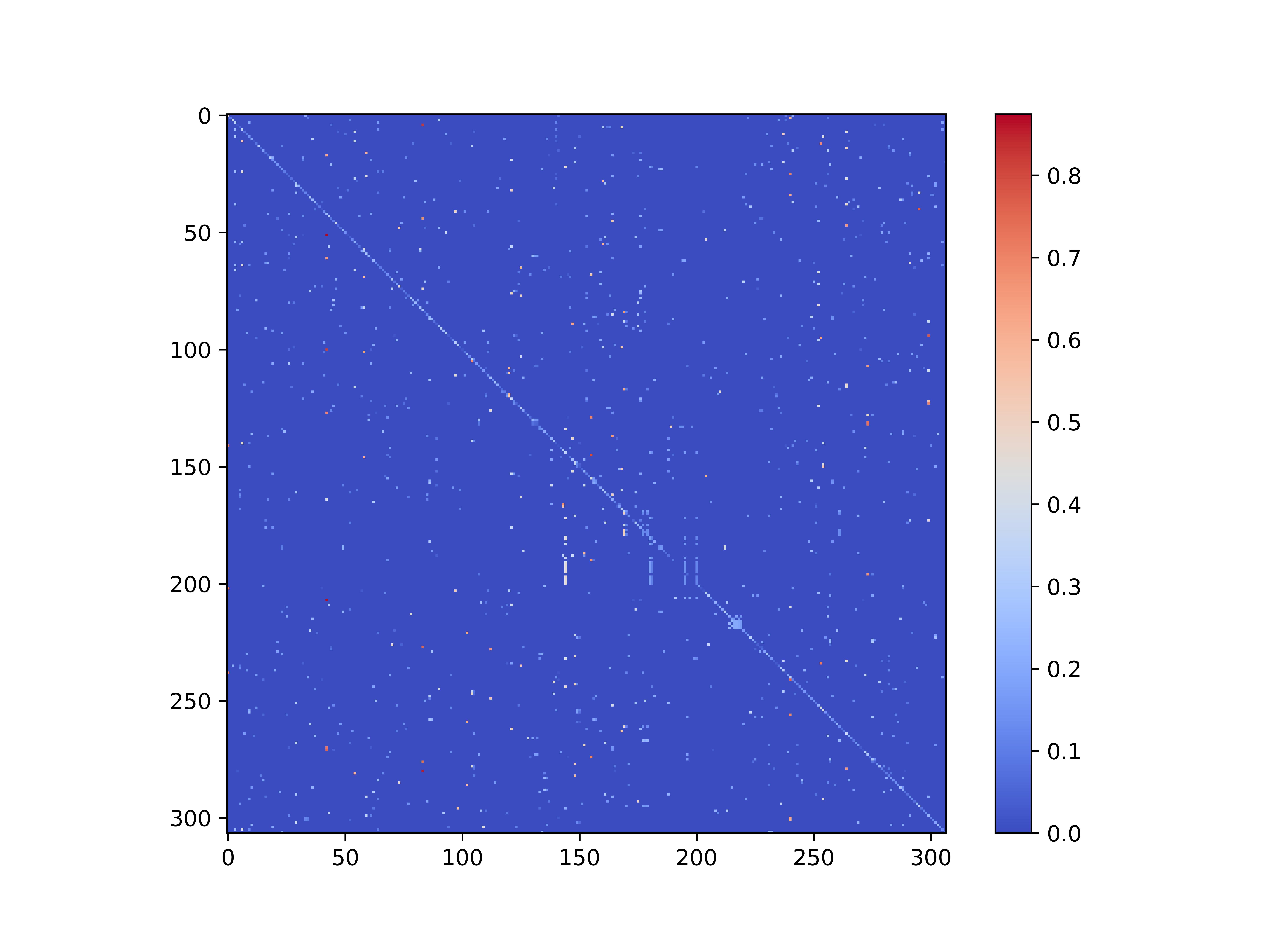}
    \label{fig:sub9}
  }\\[-6pt]
  \subfloat[Stage 1: H1, L3]{%
    \includegraphics[width=0.24\textwidth]{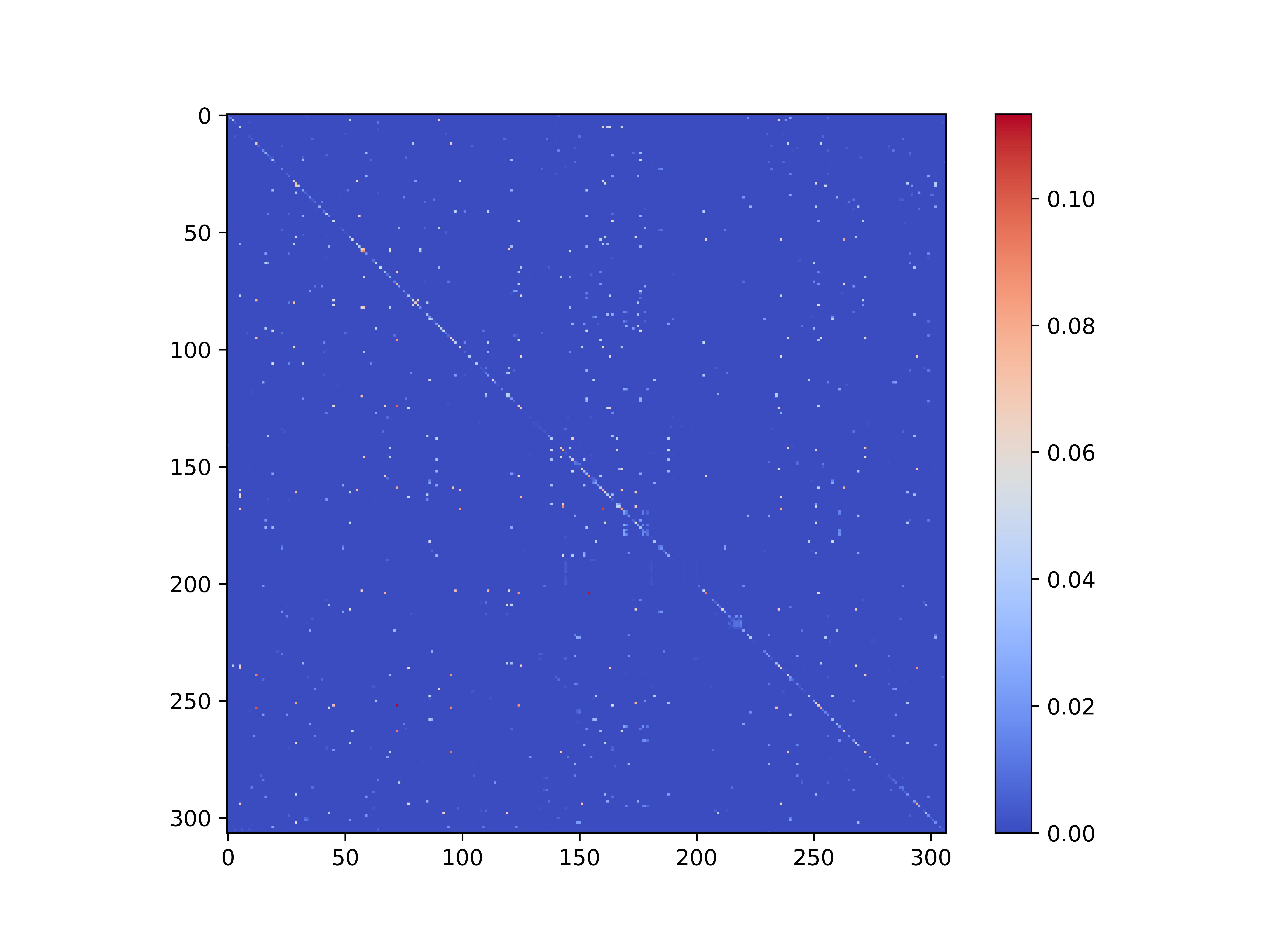}
    \label{fig:sub10}
  }
  \hfill
  \subfloat[Stage 1: H2, L3]{%
    \includegraphics[width=0.24\textwidth]{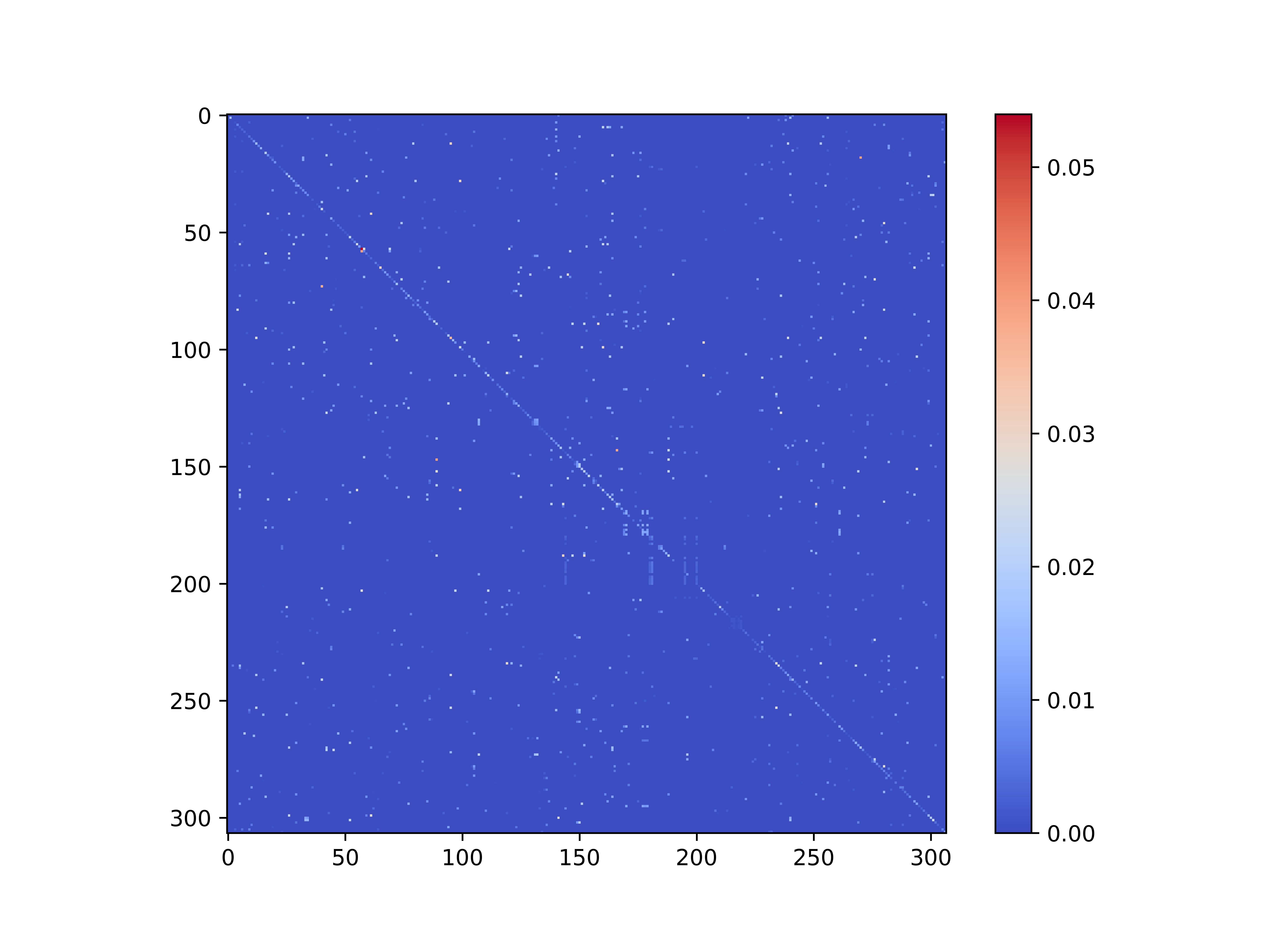}
    \label{fig:sub11}
  }
  \hfill
  \subfloat[Stage 2: H1, L3]{%
    \includegraphics[width=0.24\textwidth]{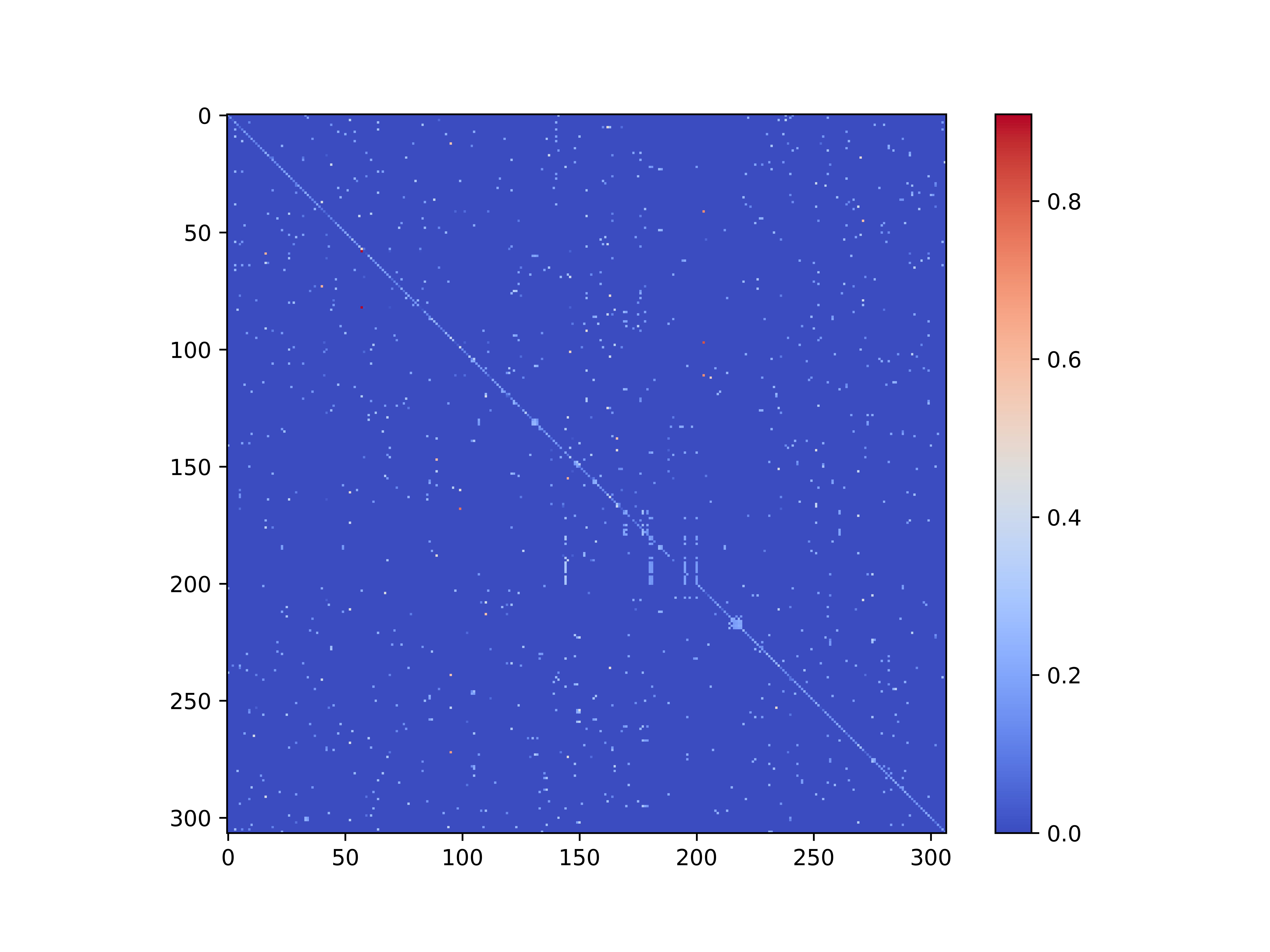}
    \label{fig:sub12}
  }
  \hfill
  \subfloat[Stage 2: H2, L3]{%
    \includegraphics[width=0.24\textwidth]{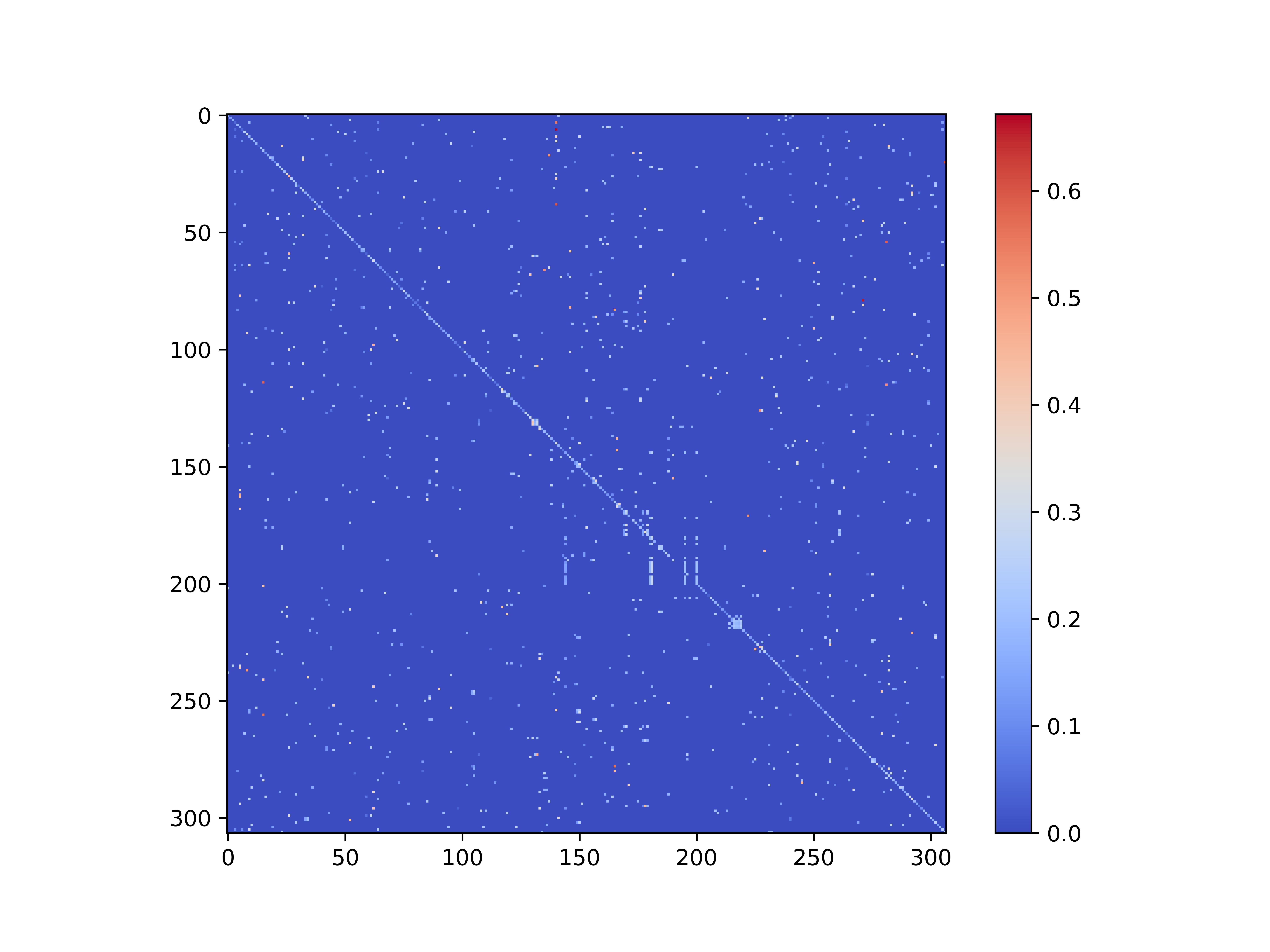}
    \label{fig:sub13}
  }\\[-6pt]
  \subfloat[Stage 1: H1, L4]{%
    \includegraphics[width=0.24\textwidth]{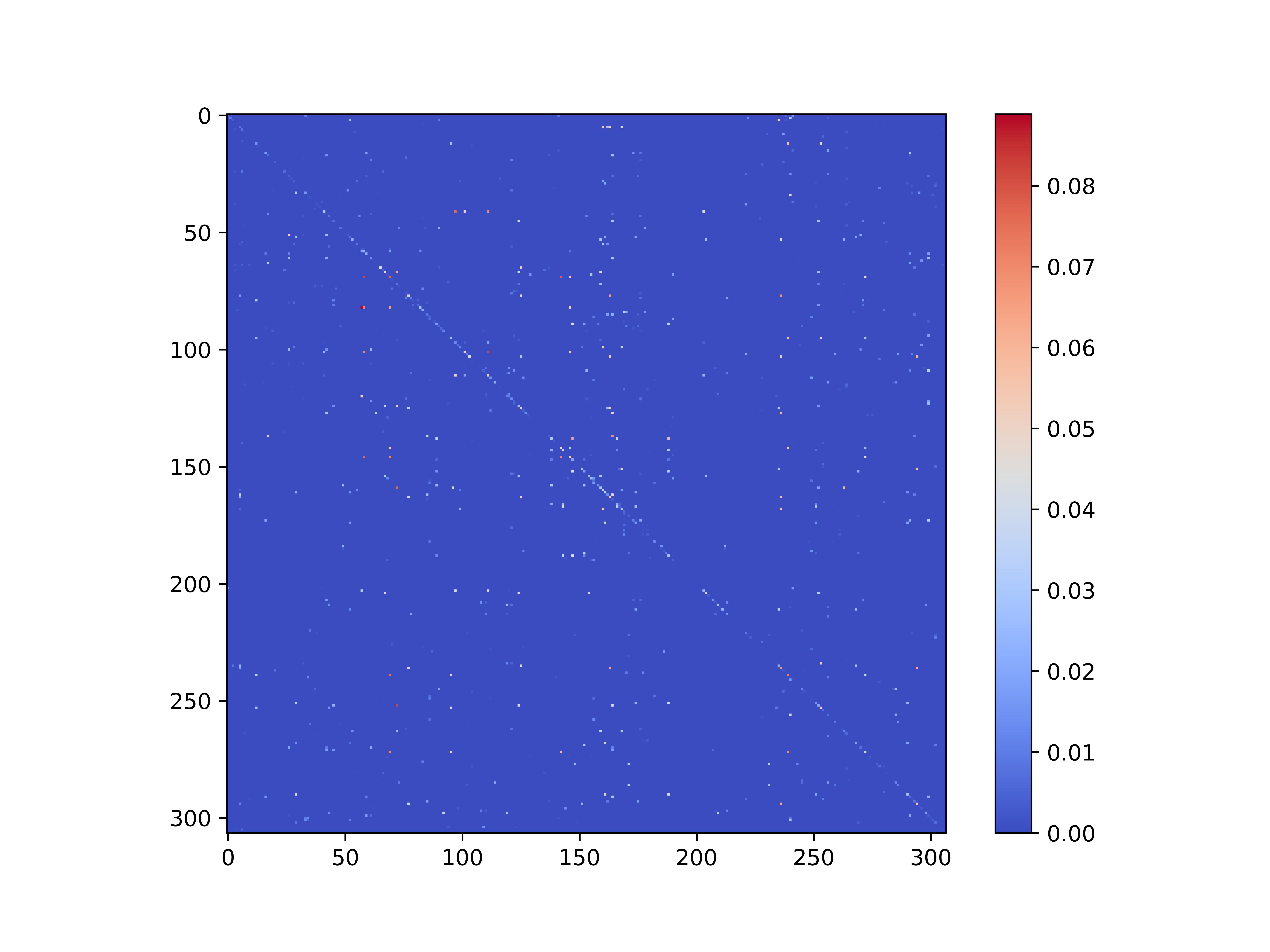}
    \label{fig:sub14}
  }
  \hfill
  \subfloat[Stage 1: H2, L4]{%
    \includegraphics[width=0.24\textwidth]{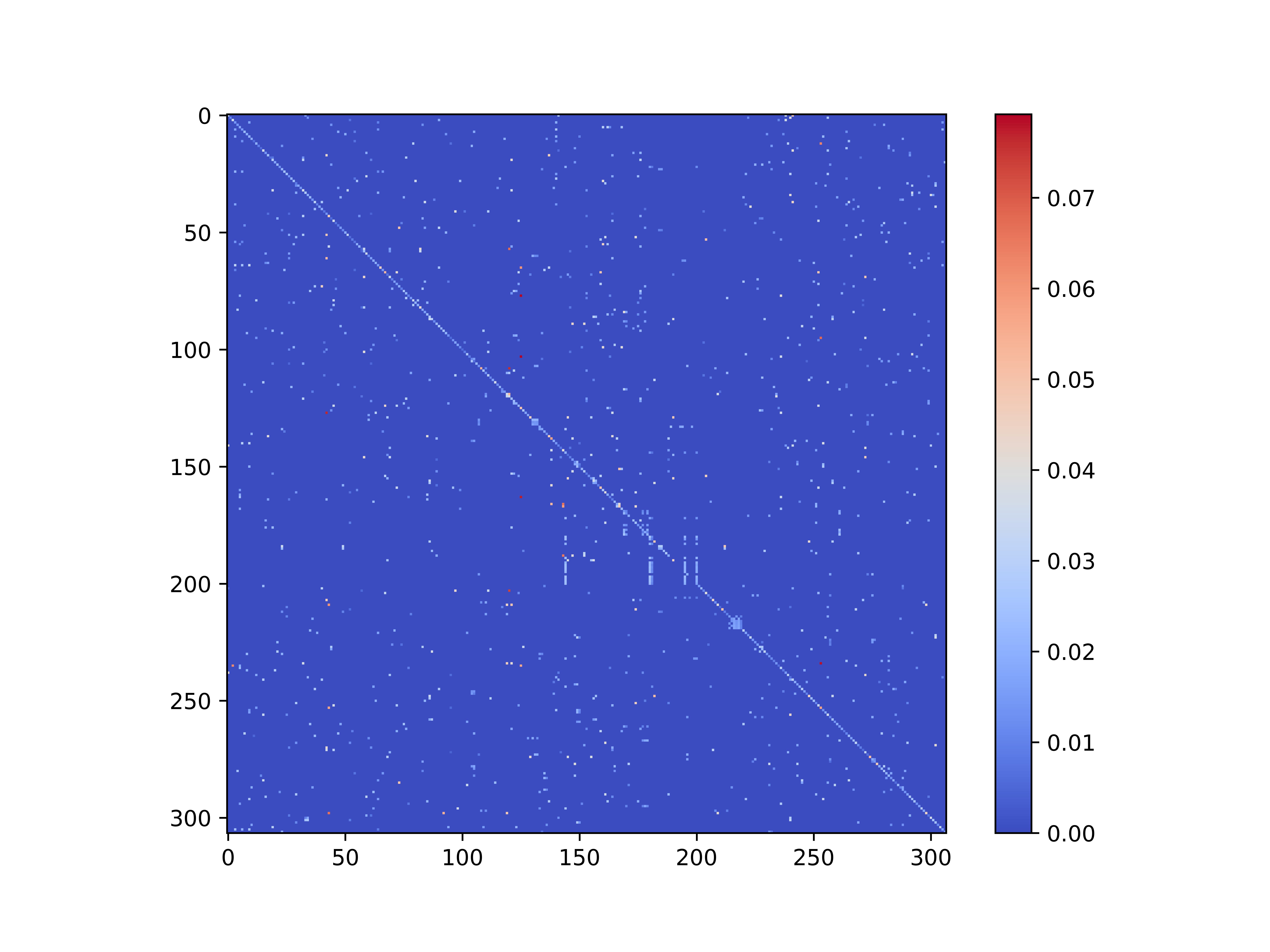}
    \label{fig:sub15}
  }
  \hfill
  \subfloat[Stage 2: H1, L4]{%
    \includegraphics[width=0.24\textwidth]{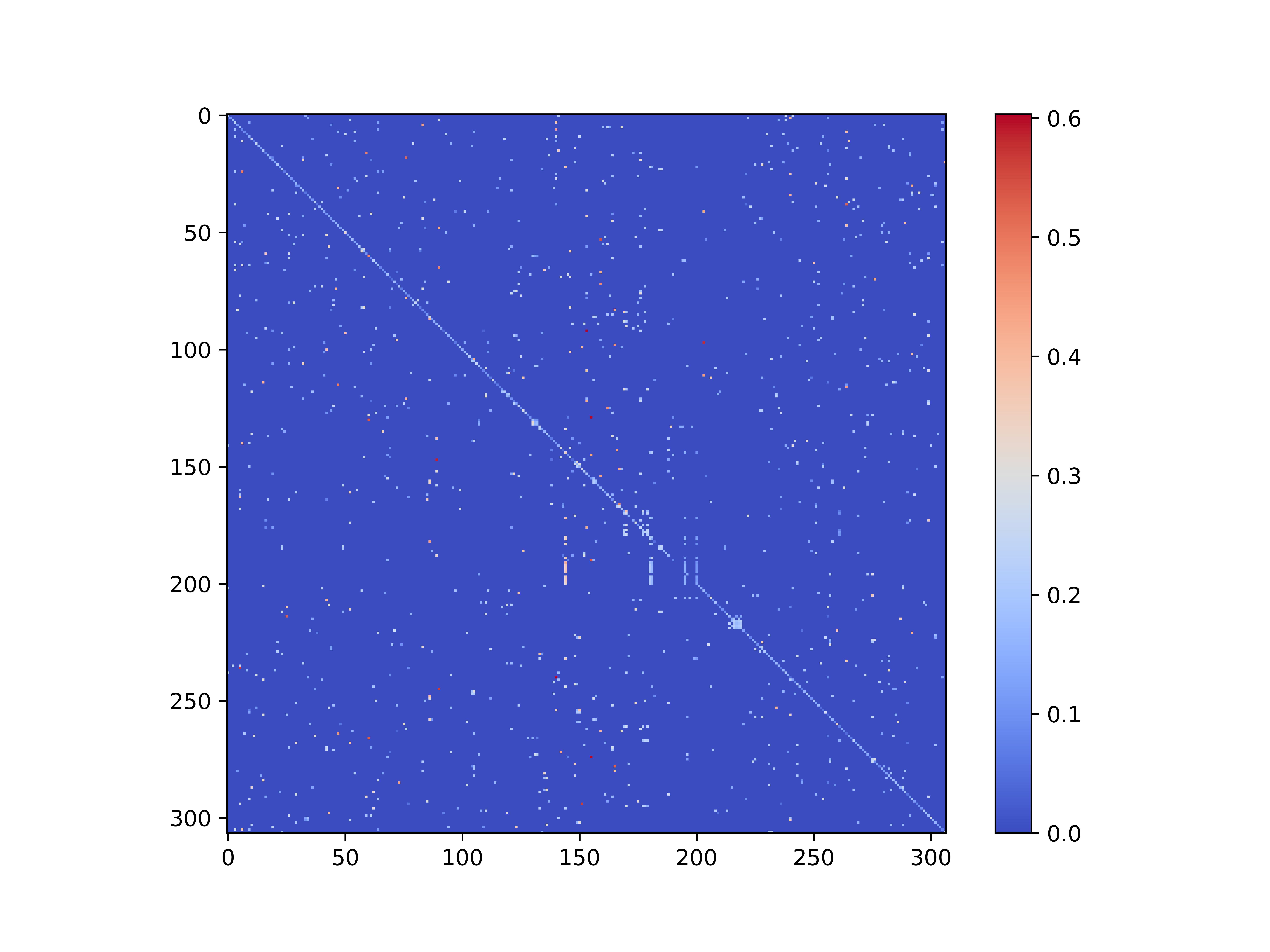}
    \label{fig:sub16}
  }
  \hfill
  \subfloat[Stage 2: H2, L4]{%
    \includegraphics[width=0.24\textwidth]{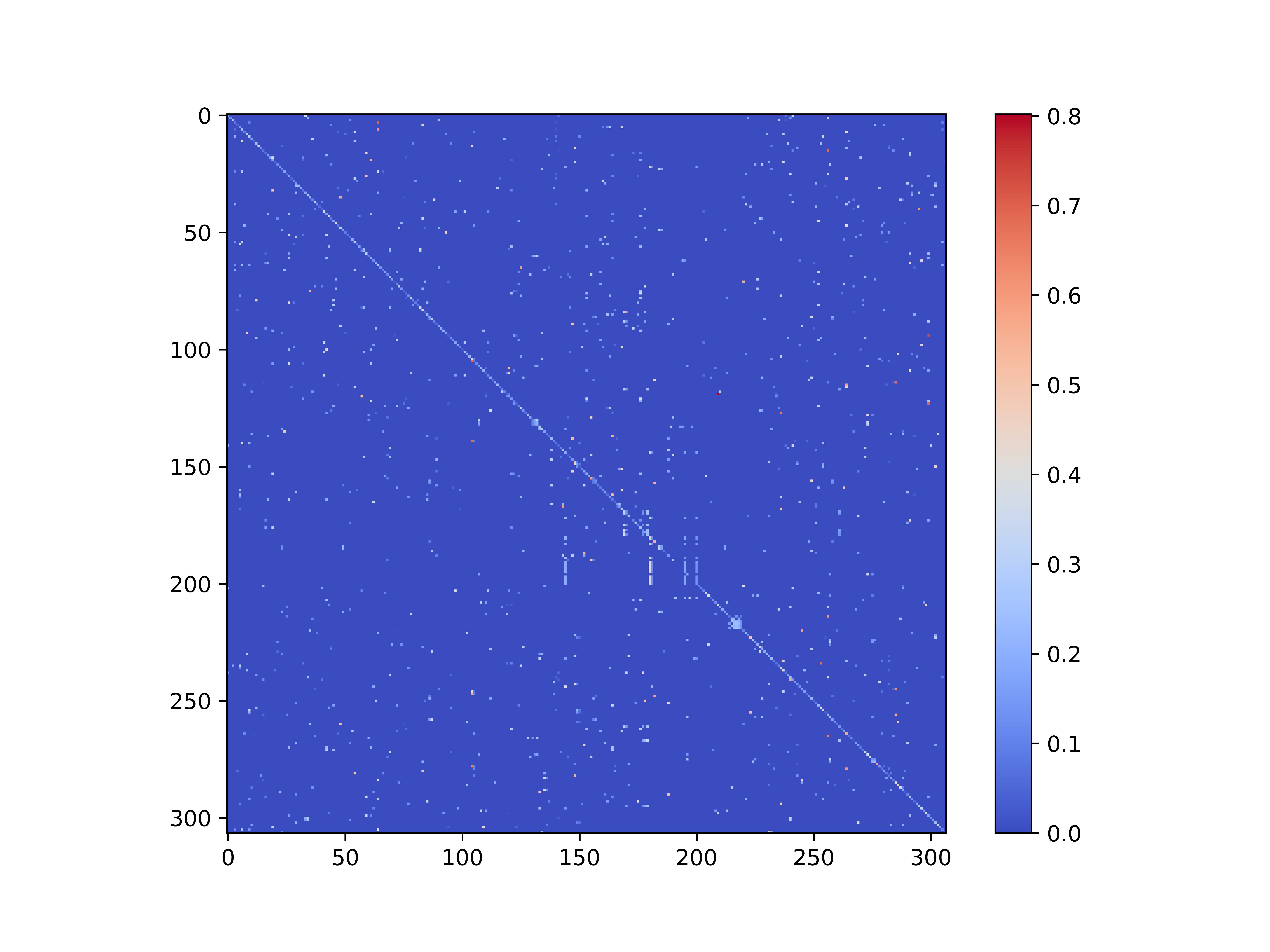}
    \label{fig:sub17}
  }
  \\[-6pt]
  \subfloat[Stage 1: H1, L5]{%
    \includegraphics[width=0.24\textwidth]{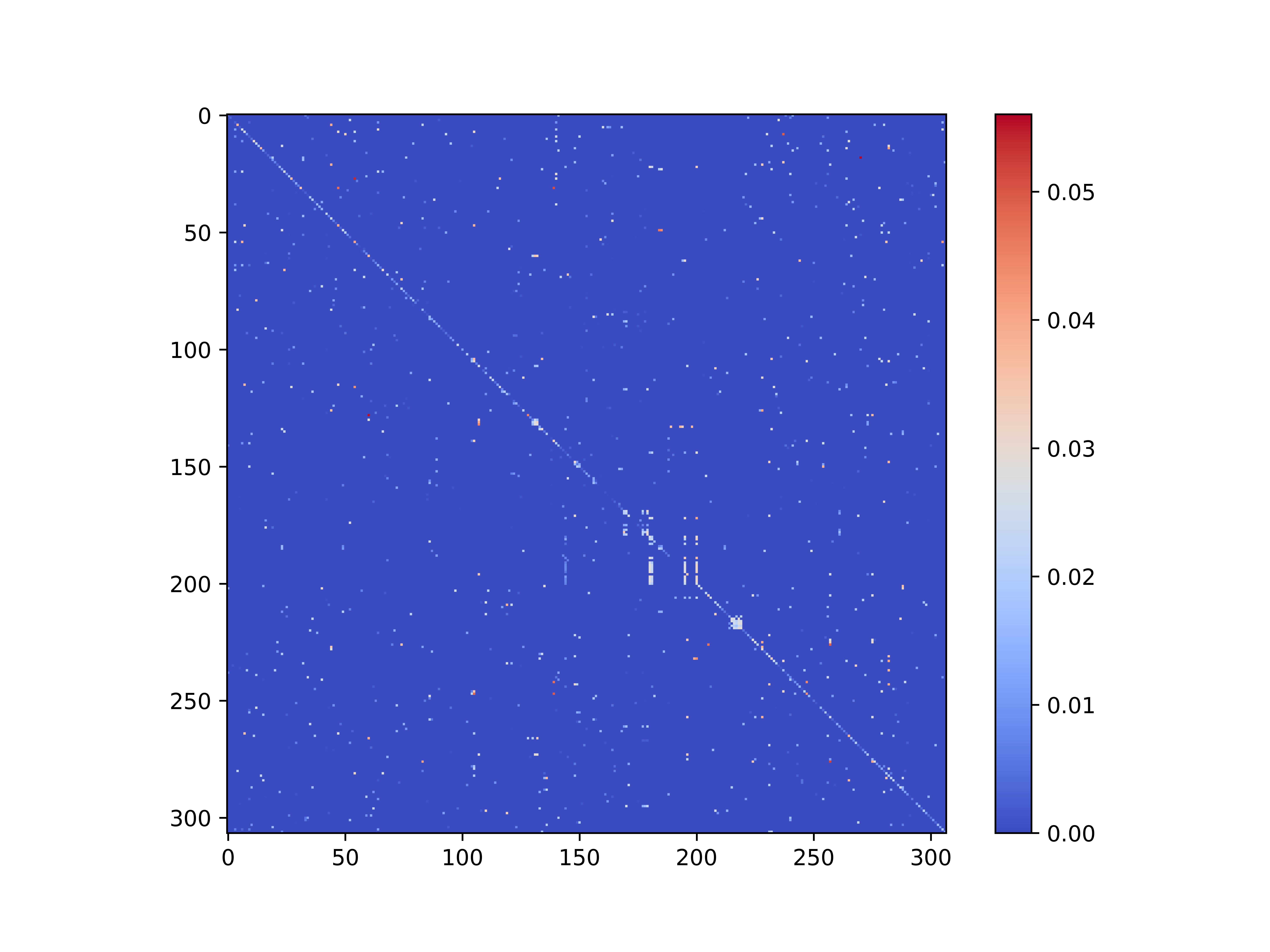}
    \label{fig:sub18}
  }
  \hfill
  \subfloat[Stage 1: H2, L5]{%
    \includegraphics[width=0.24\textwidth]{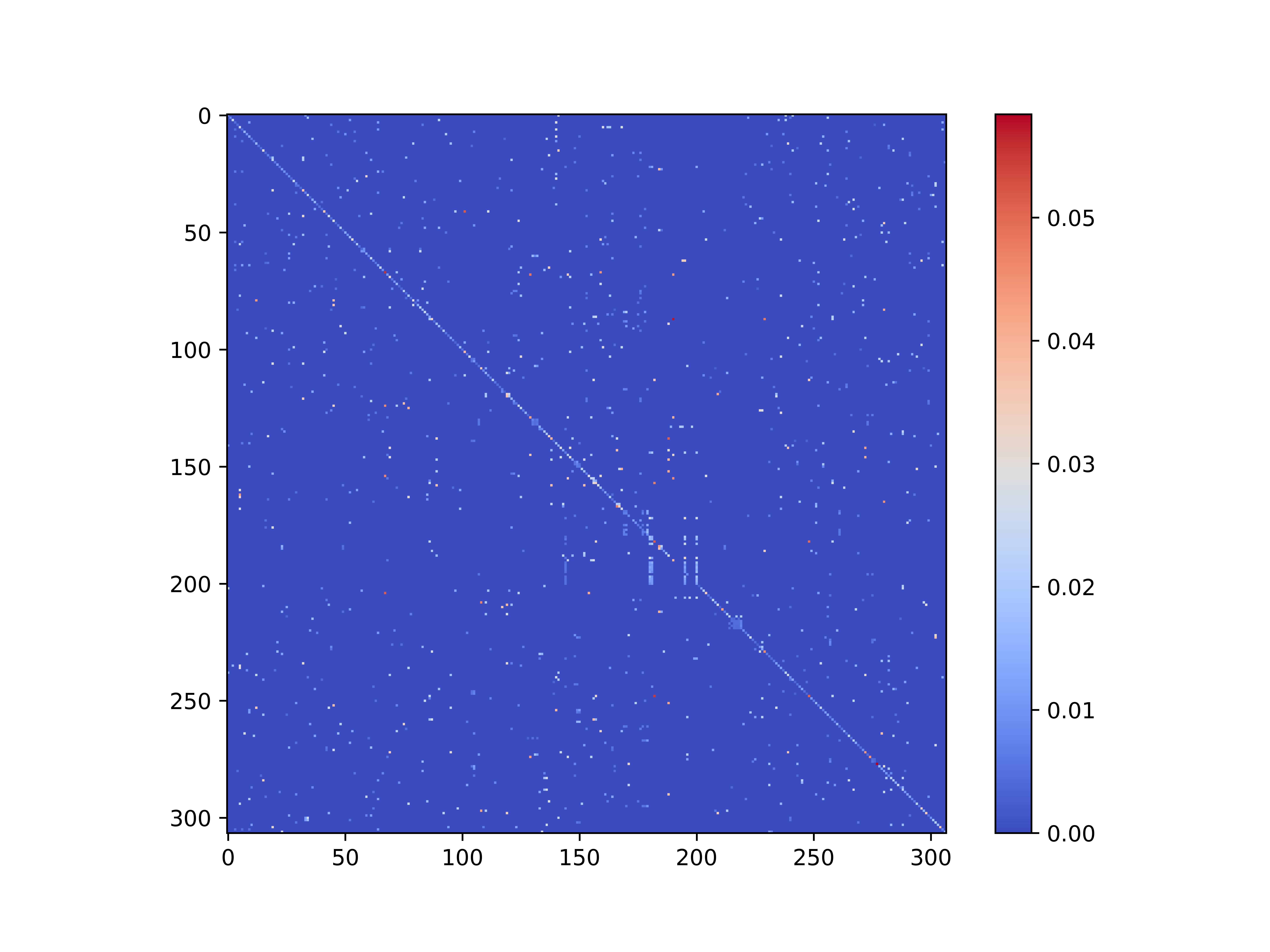}
    \label{fig:sub19}
  }
  \hfill
  \subfloat[Stage 2: H1, L5]{%
  \includegraphics[width=0.24\textwidth] {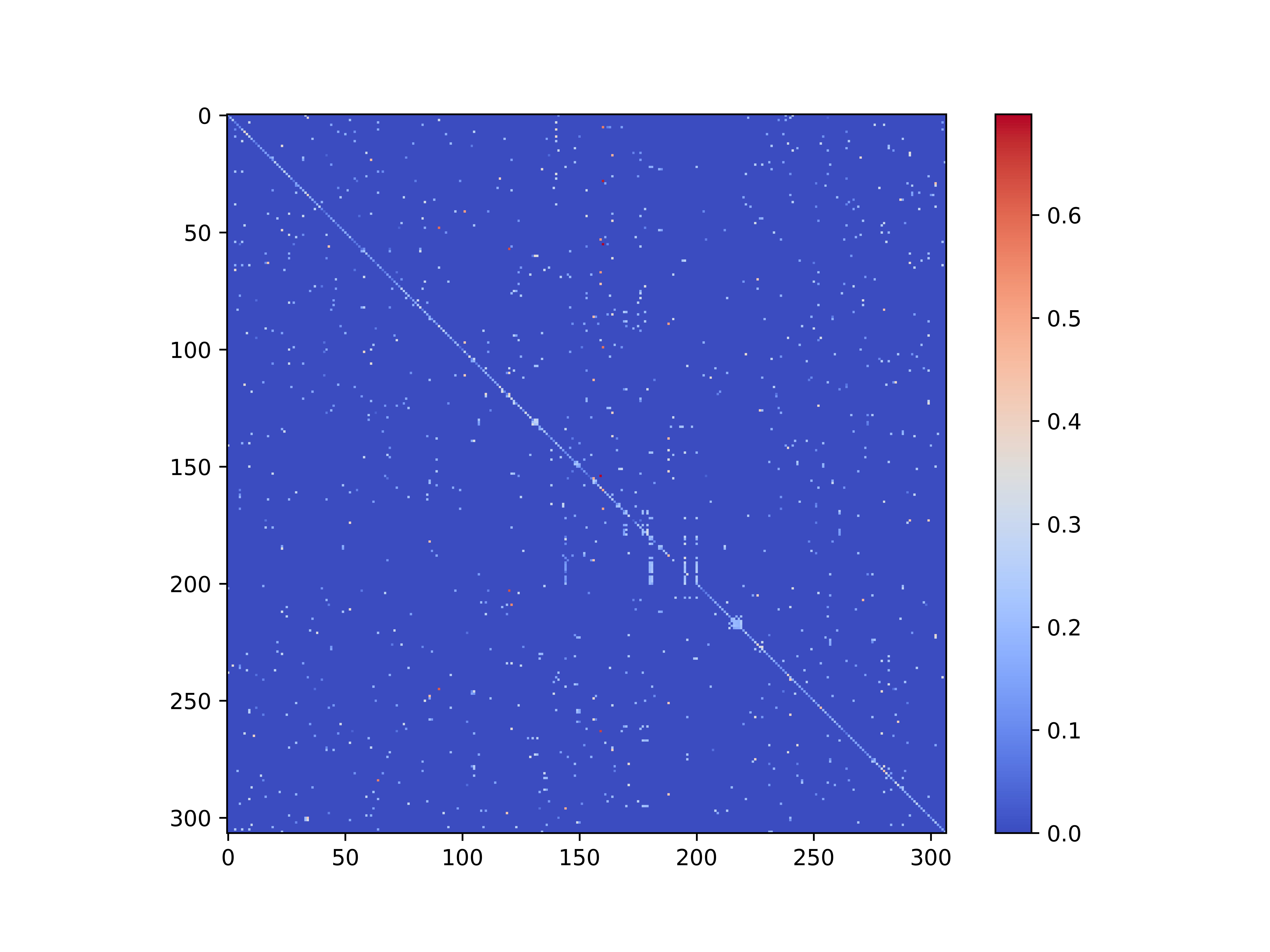}
    \label{fig:sub20}
    }
  \hfill
  \subfloat[Stage 2: H2, L5]{%
    \includegraphics[width=0.24\textwidth]{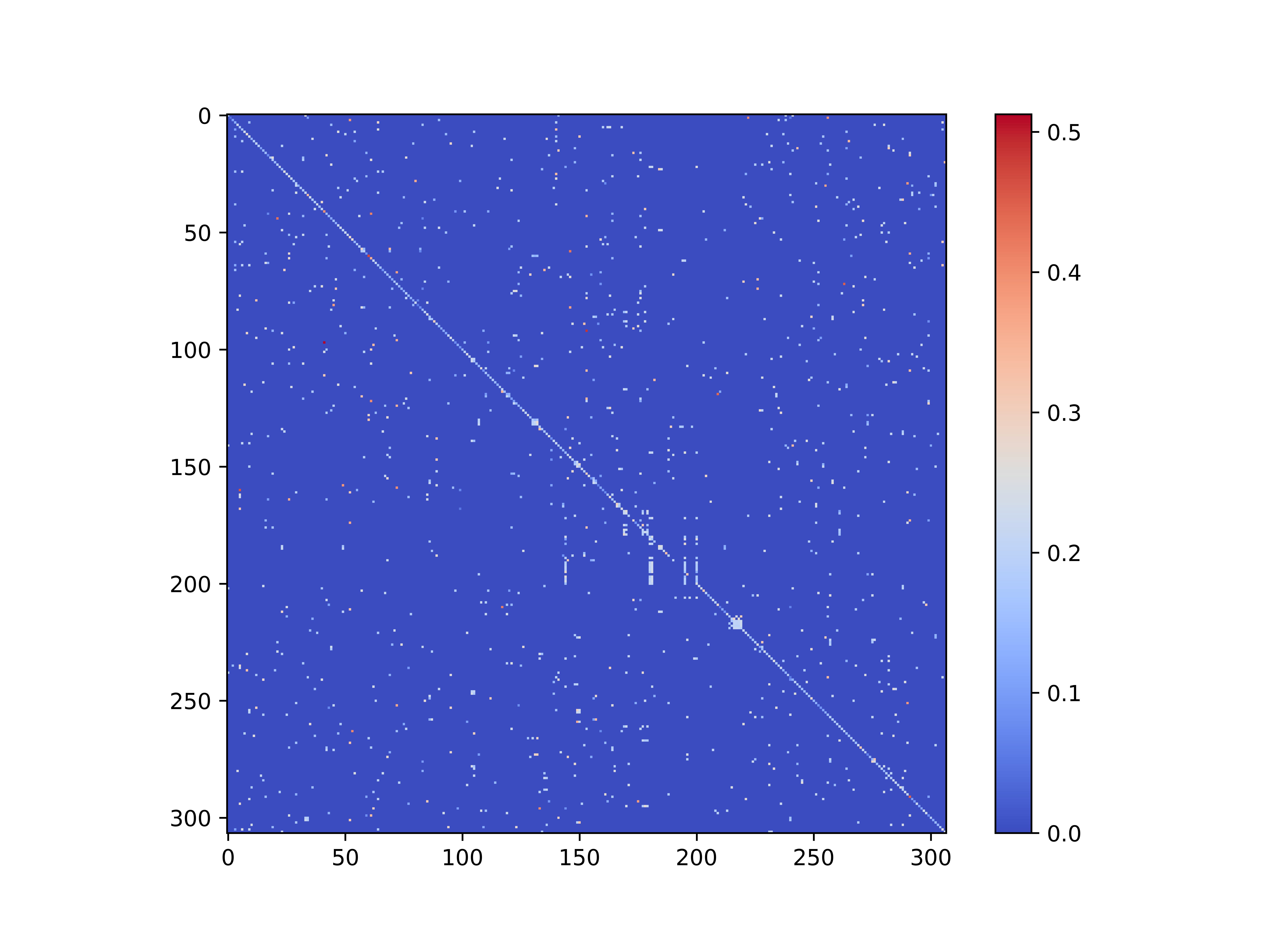}
    \label{fig:sub21}
  }
  \\[-6pt]
  \subfloat[Stage 1: H1, L6]{%
    \includegraphics[width=0.24\textwidth]{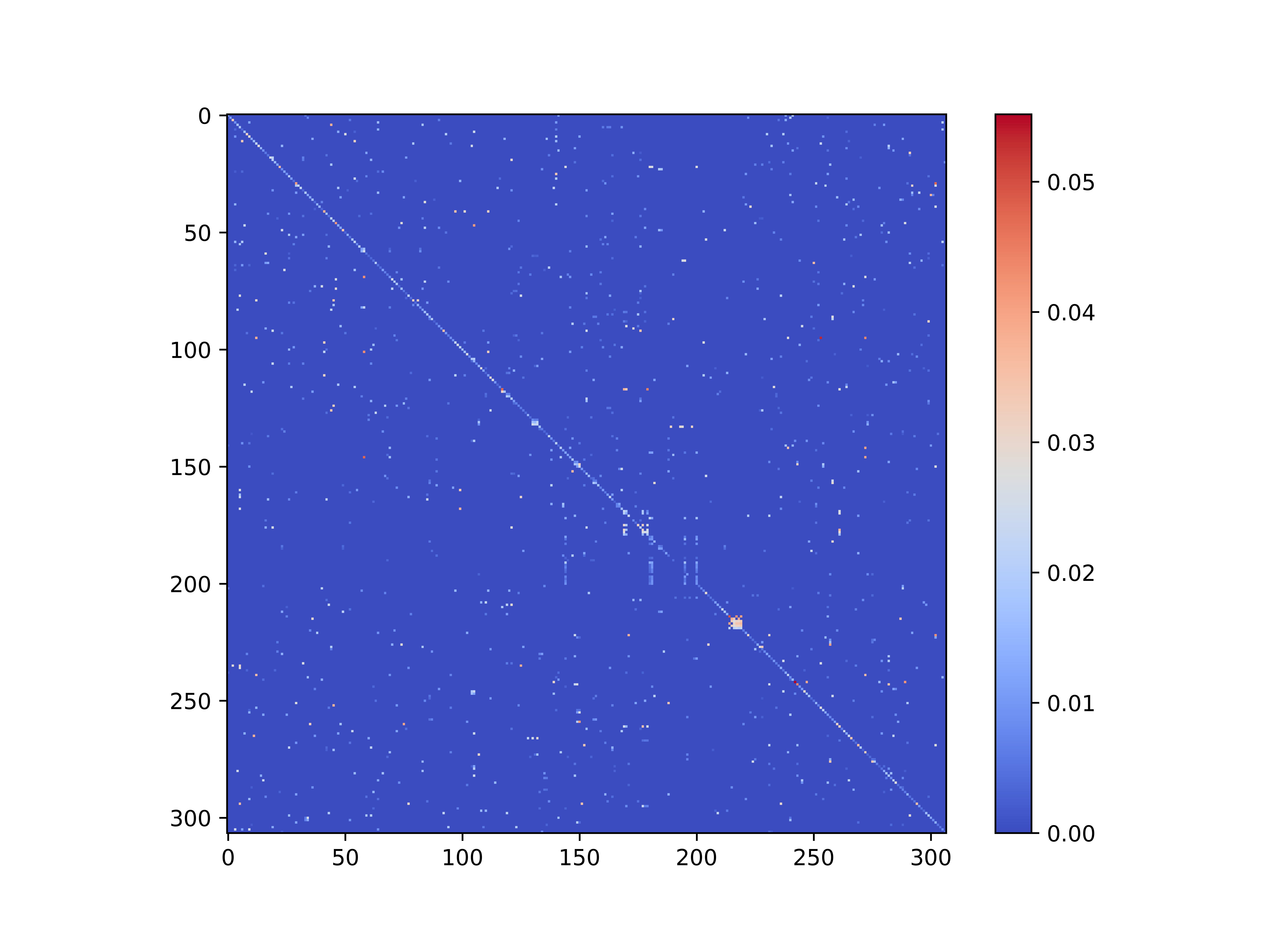}
    \label{fig:sub22}
  }
  \hfill
  \subfloat[Stage 1: H2, L6]{%
    \includegraphics[width=0.24\textwidth]{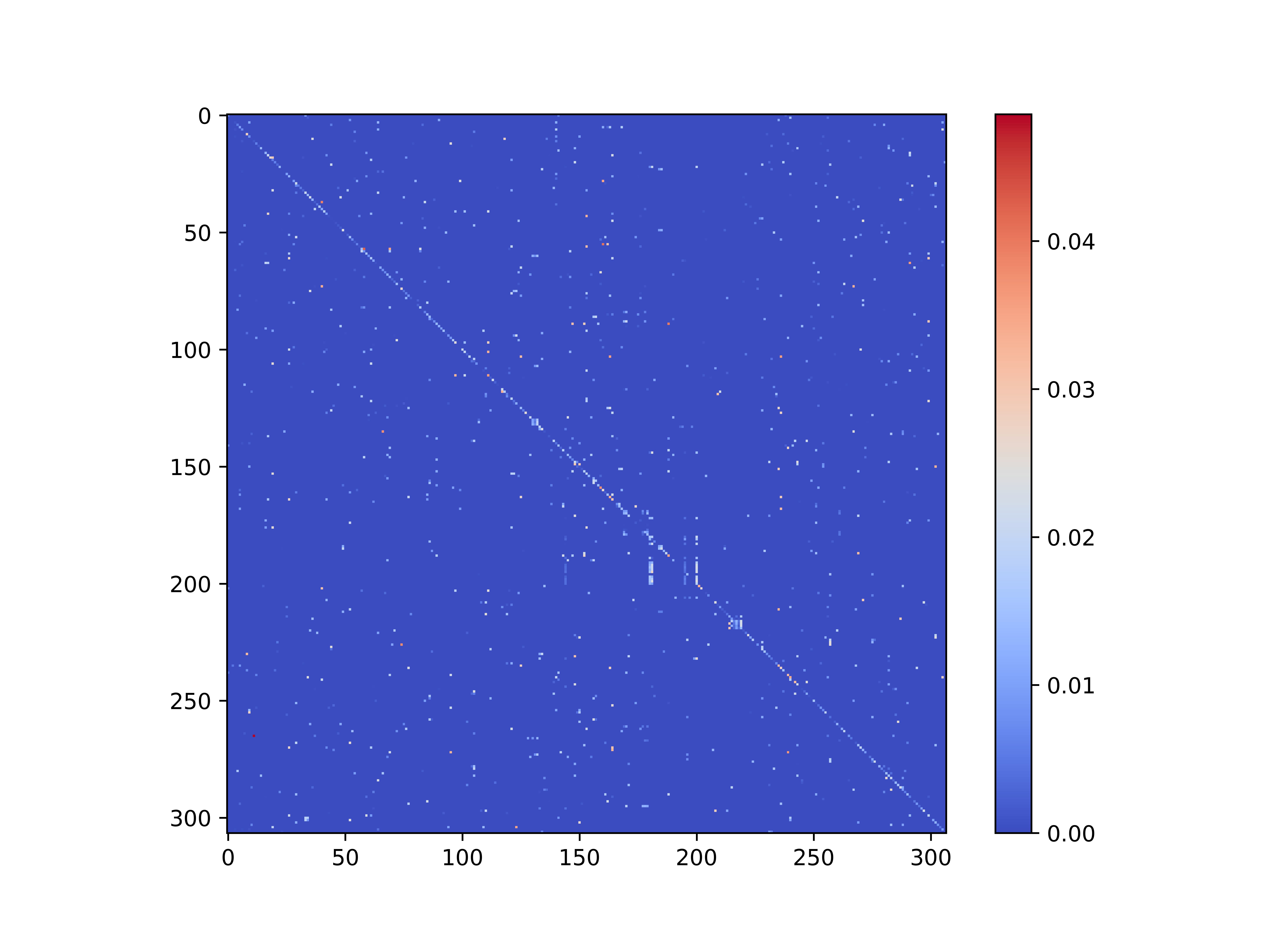}
    \label{fig:sub23}
  }
  \hfill
  \subfloat[Stage 2: H1, L6]{%
    \includegraphics[width=0.24\textwidth]{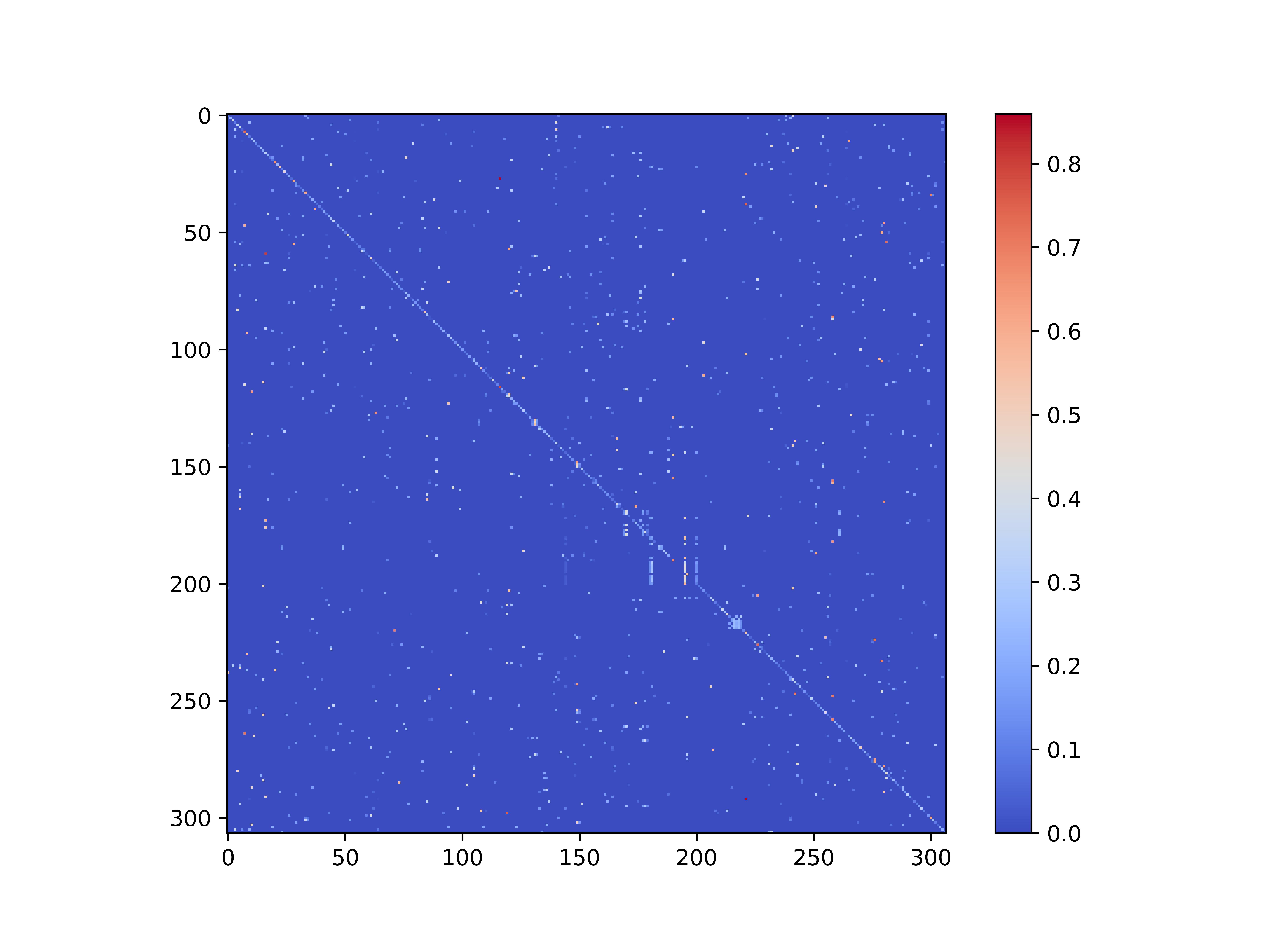}
    \label{fig:sub24}
  }
  \hfill
  \subfloat[Stage 2: H2, L6]{%
    \includegraphics[width=0.24\textwidth]{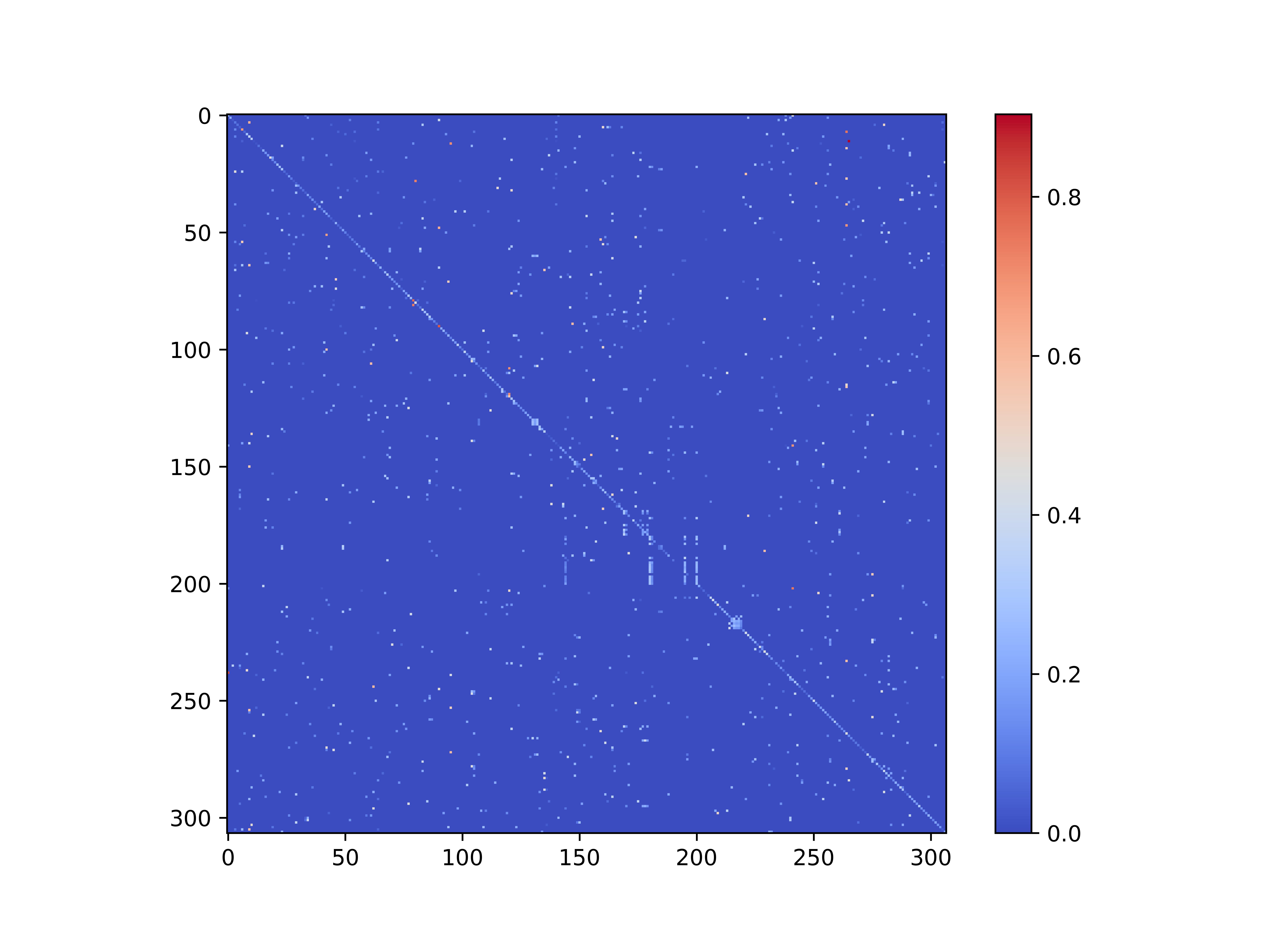}
    \label{fig:sub26}
  }
\end{figure*}

\begin{figure*}[htbp]
  \centering
  \caption{Attention scores in \textbf{ReTSA} by layers: horizontal axis shows stage progression and more heads; vertical axis reflects deeper layers. Note: H$n$ = Head $n$, L$n$ = Layer $n$}
\label{fig:attentionmap_retsa}
  \subfloat[Stage 1: H1, L1]{%
\includegraphics[width=0.24\textwidth]{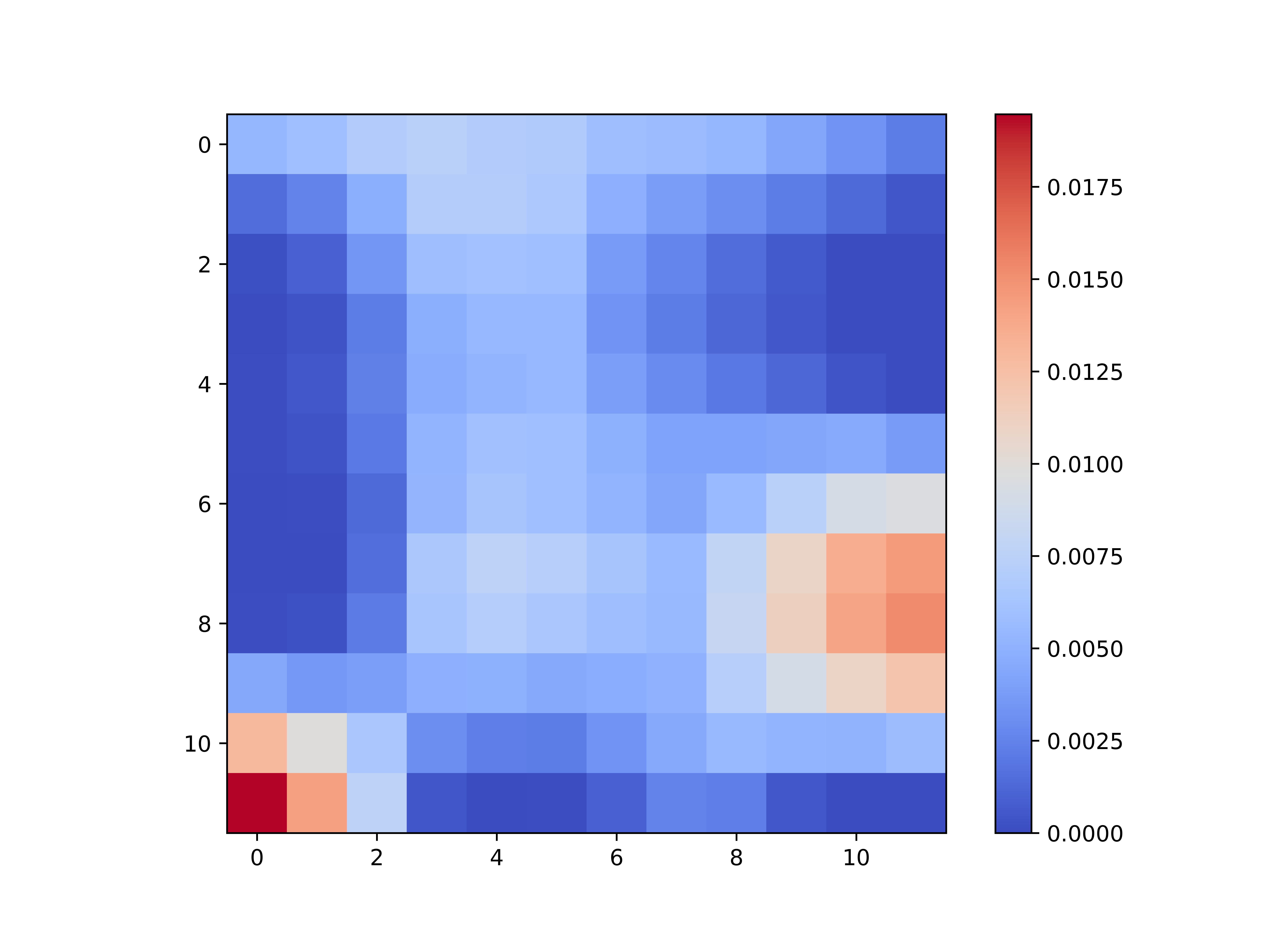}
    \label{fig:sub27}
  }
  \hfill
  \subfloat[Stage 1: H2, L1]{%
\includegraphics[width=0.24\textwidth]{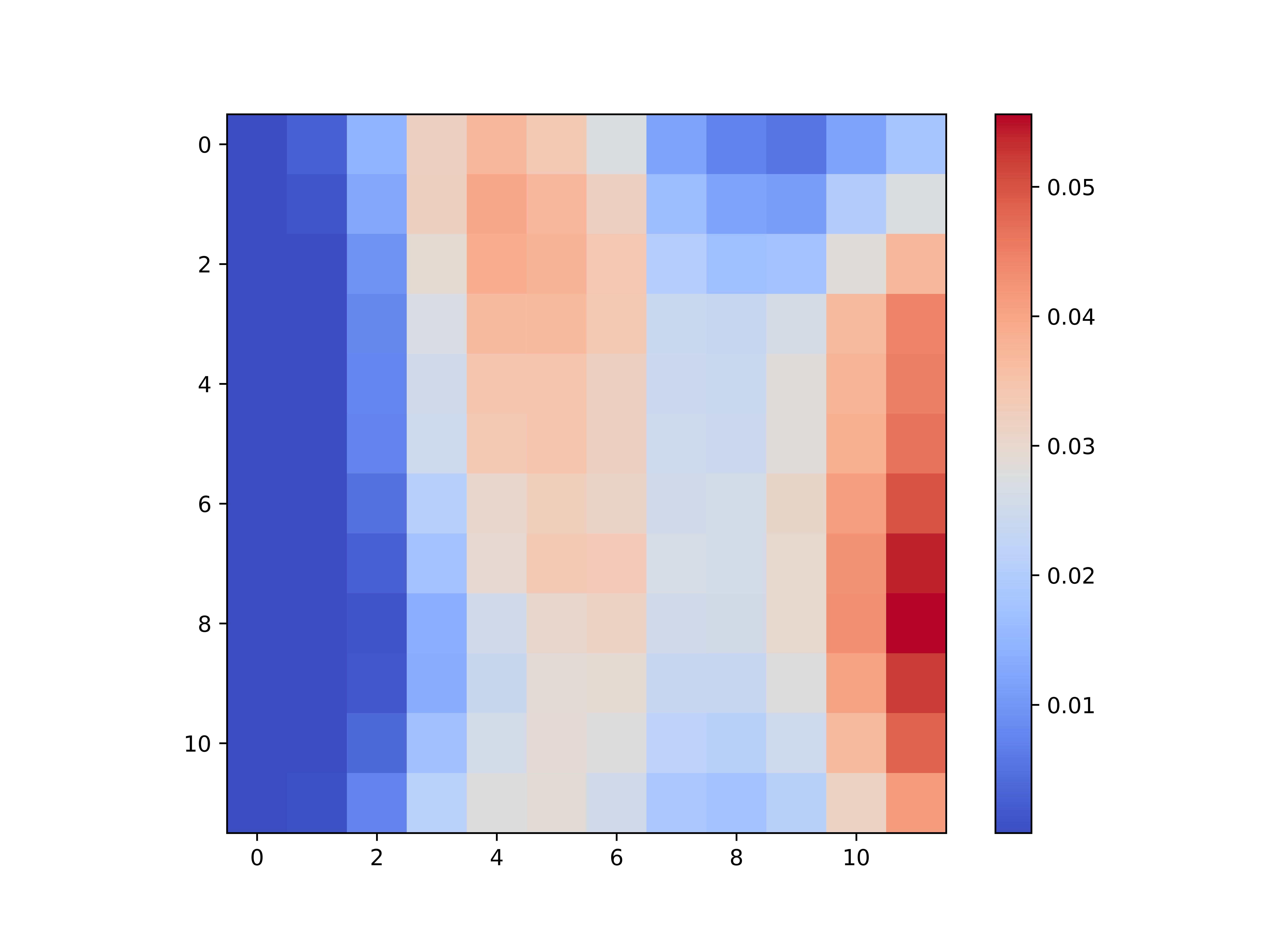}
    \label{fig:sub28}
  }
  \hfill
  \subfloat[Stage 2: H1, L1]{%
\includegraphics[width=0.24\textwidth]{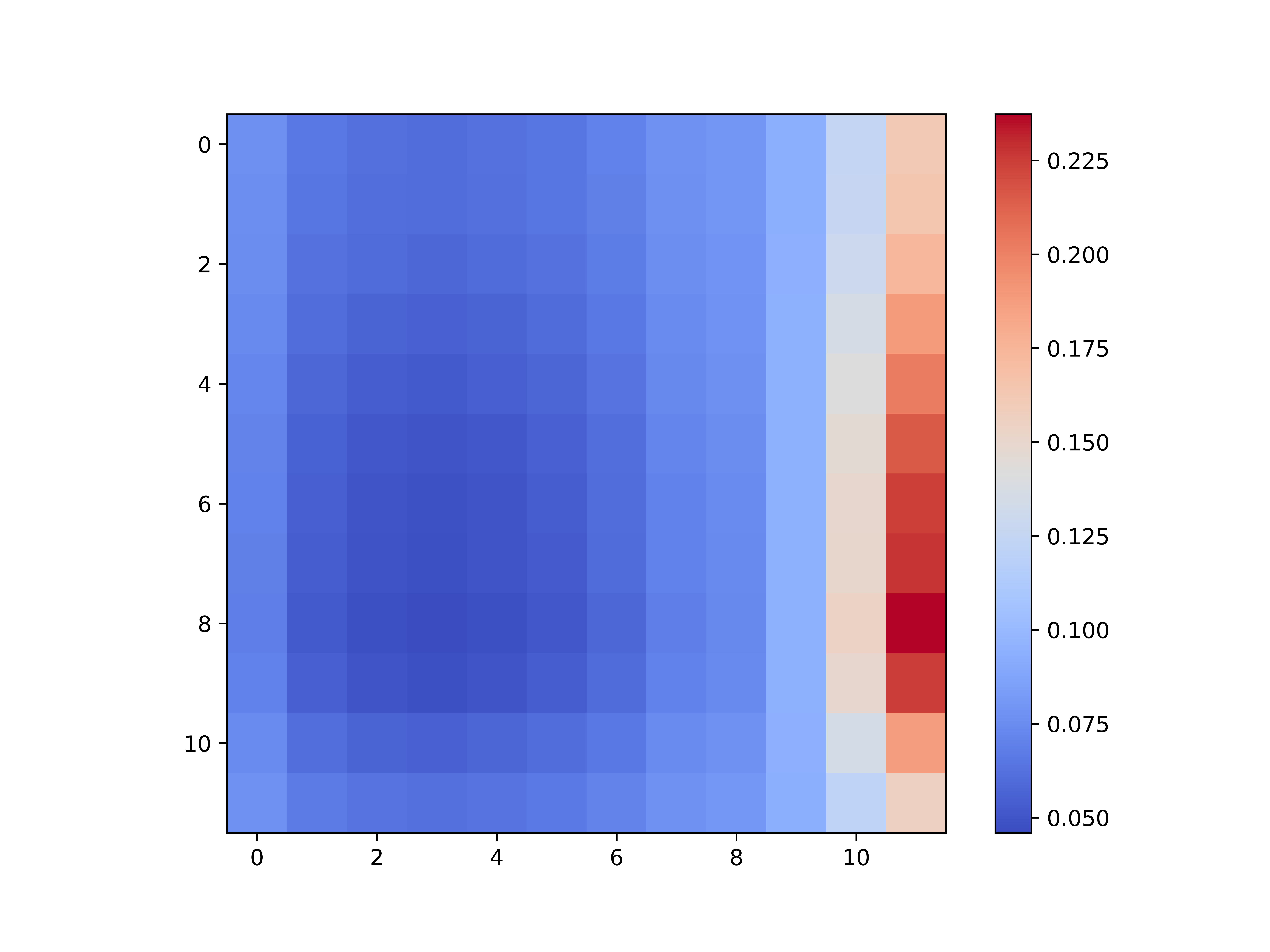}
    \label{fig:sub29}
  }
  \hfill
  \subfloat[Stage 2: H2, L1]{%
\includegraphics[width=0.24\textwidth]{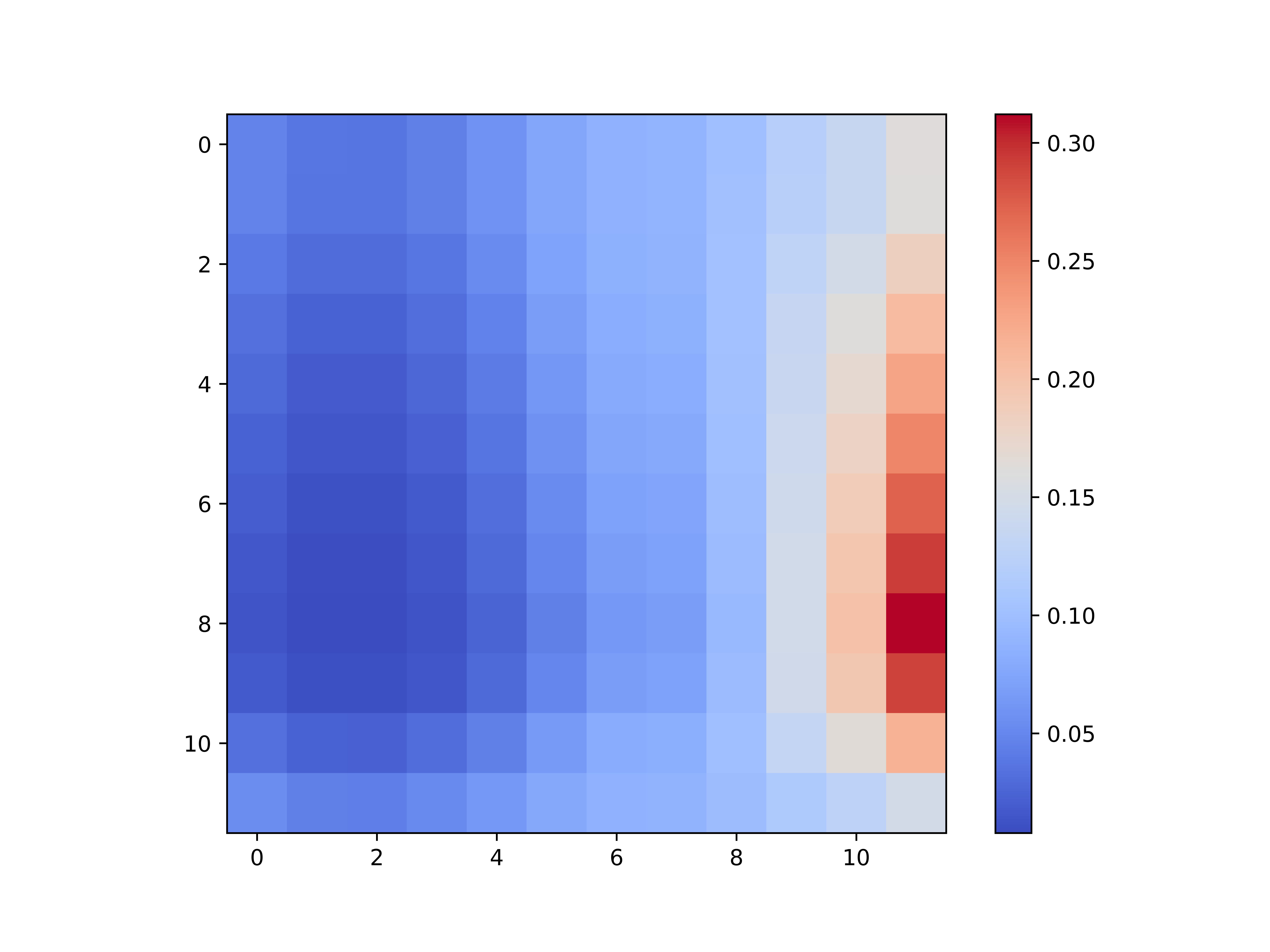}
    \label{fig:sub30}
  } 
  \\[-6pt]
  \subfloat[Stage 1: H1, L2]{%
    \includegraphics[width=0.24\textwidth]{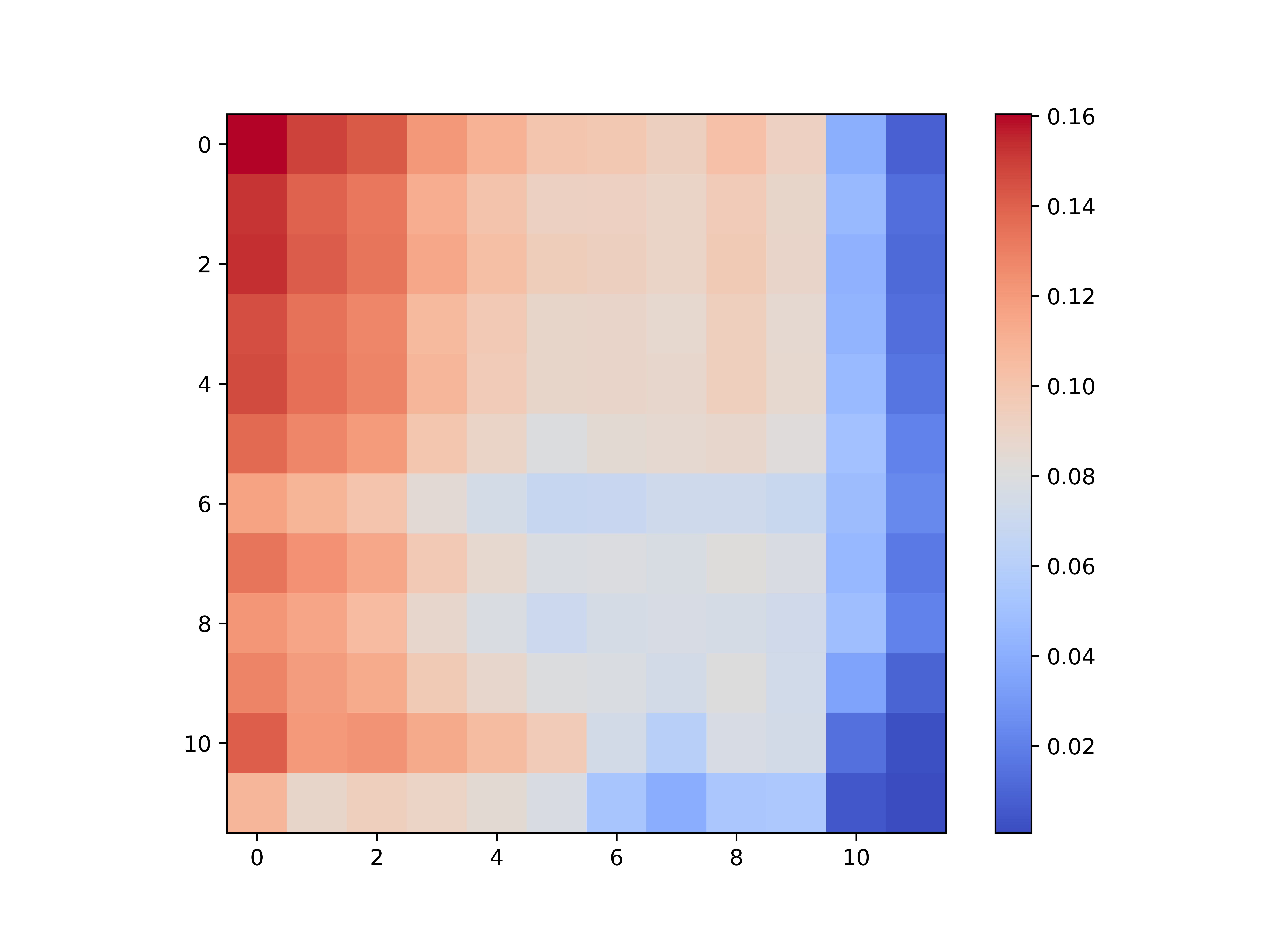}
    \label{fig:sub31}
  }
  \hfill
  \subfloat[Stage 1: H2, L2]{%
    \includegraphics[width=0.24\textwidth]{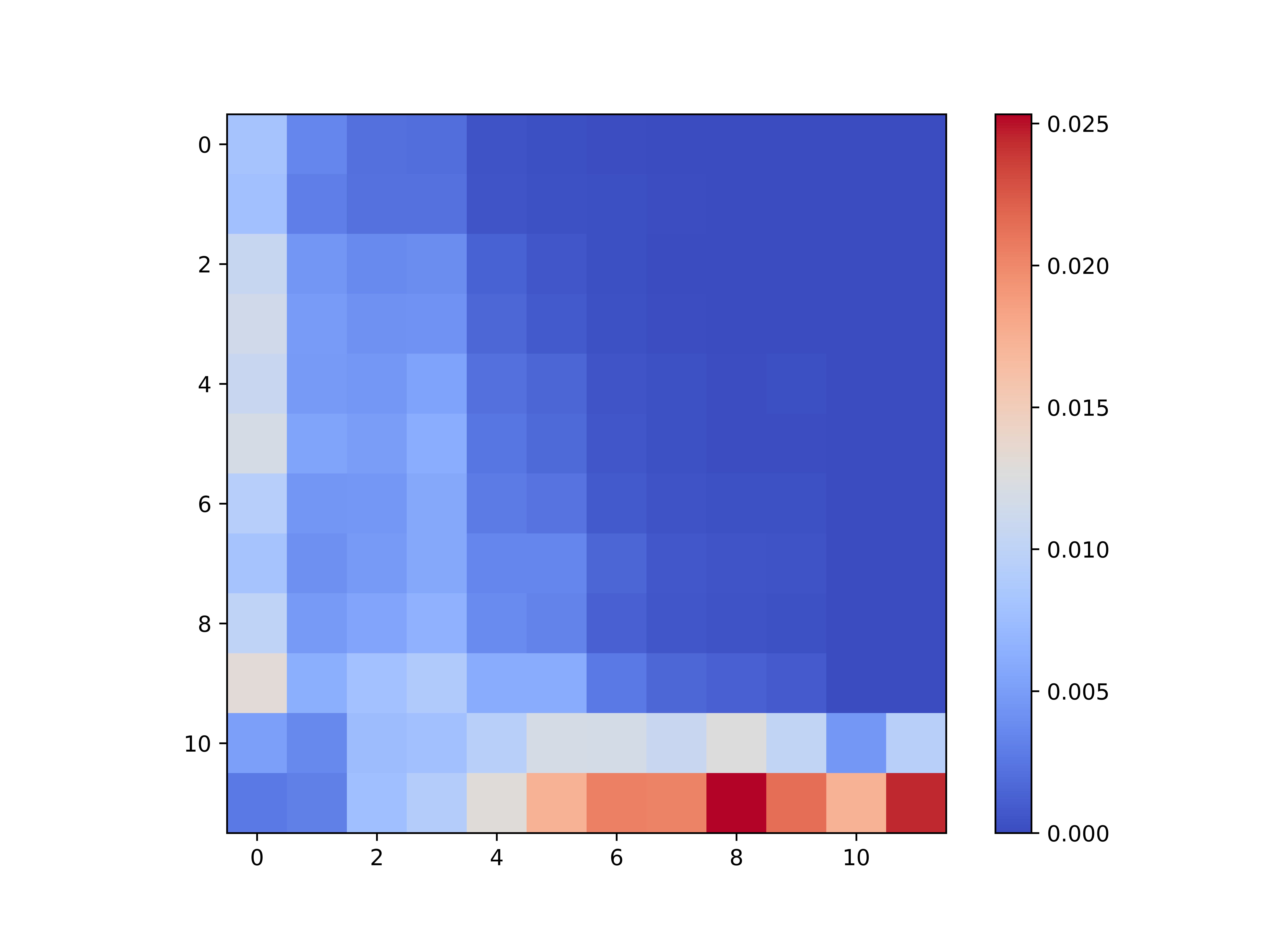}
    \label{fig:sub32}
  }
  \hfill
  \subfloat[Stage 2: H1, L2]{%
    \includegraphics[width=0.24\textwidth]{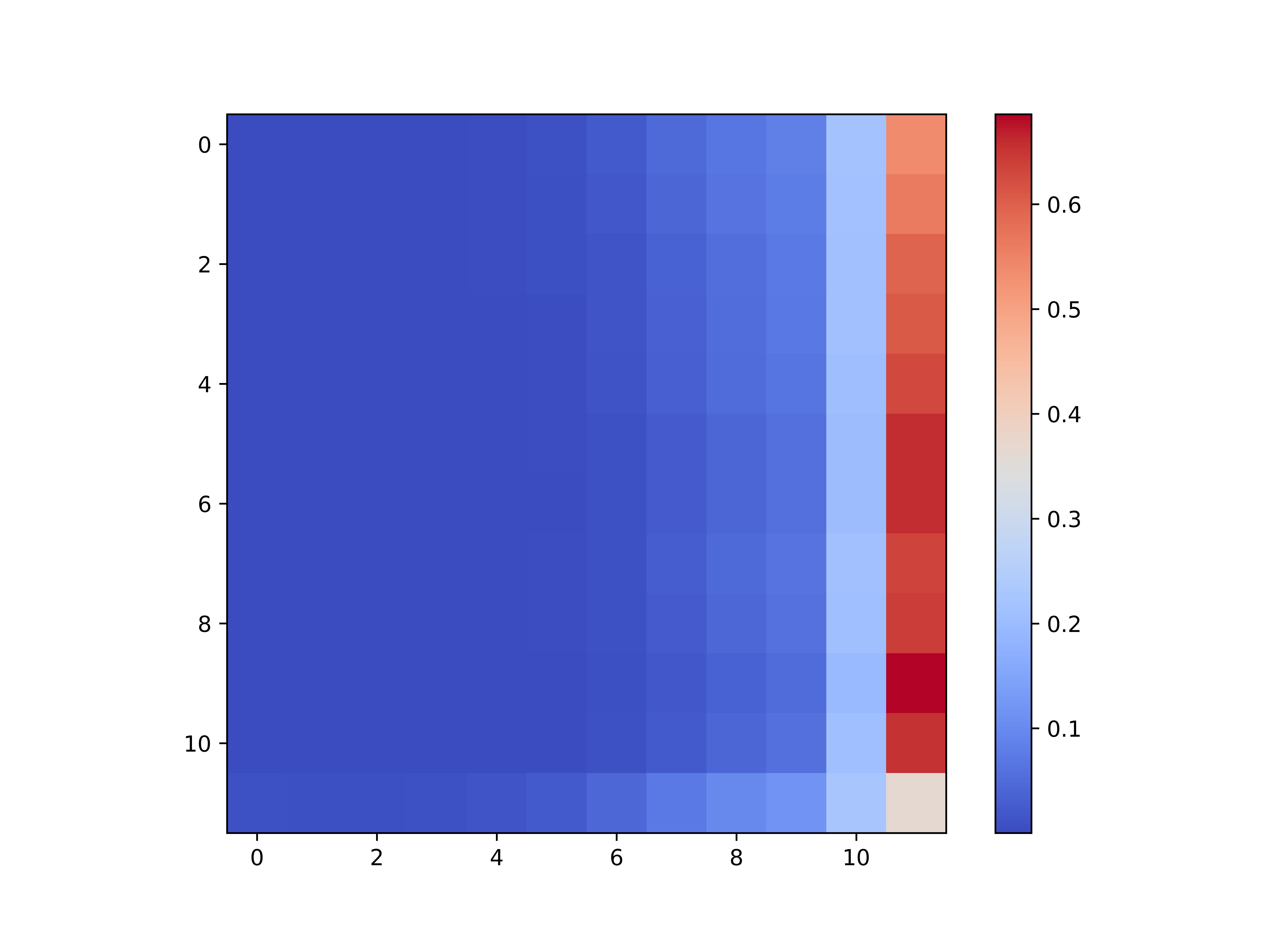}
    \label{fig:sub33}
  }
  \hfill
  \subfloat[Stage 2: H2, L2]{%
    \includegraphics[width=0.24\textwidth]{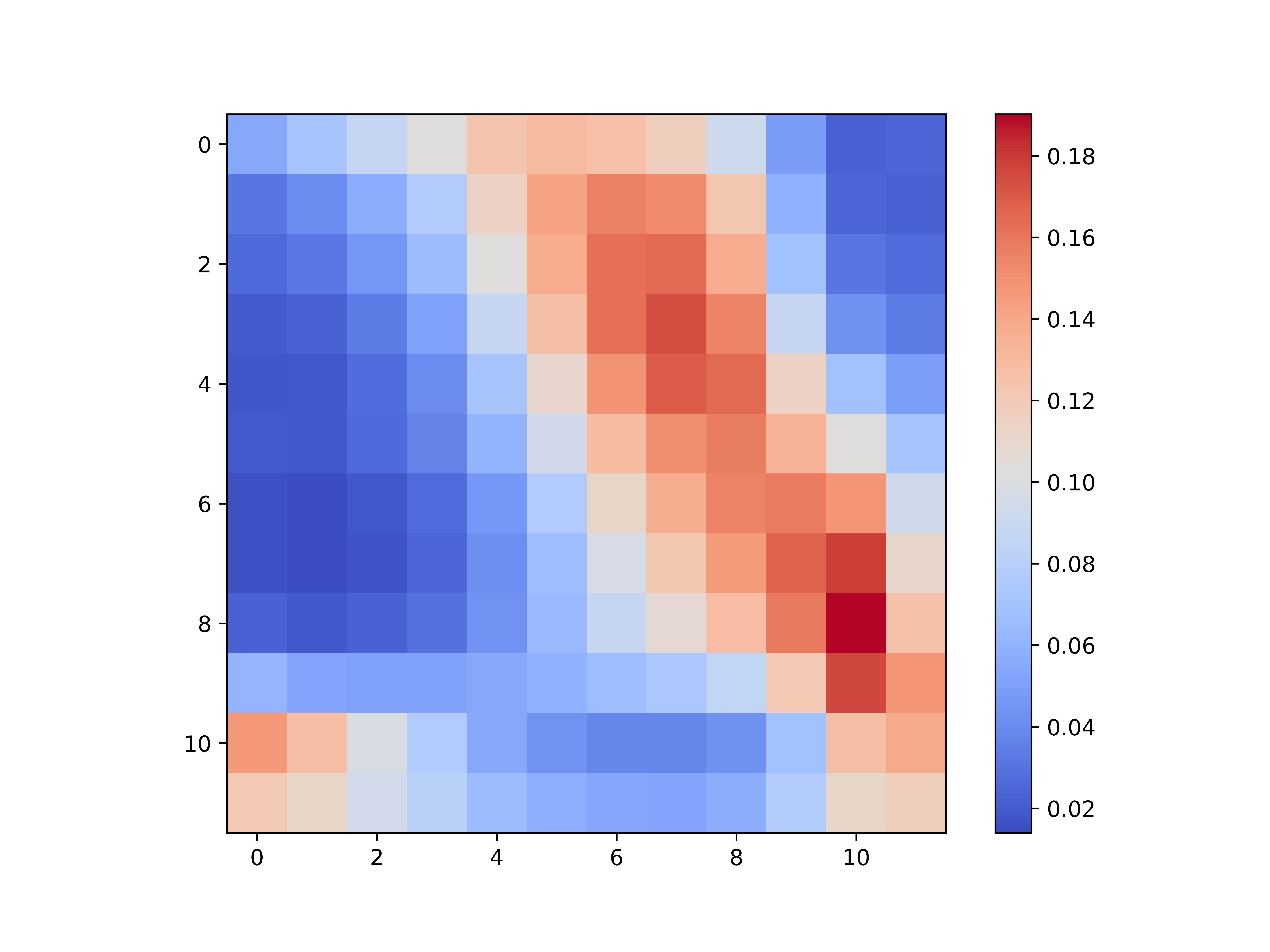}
    \label{fig:sub34}
  }
\\[-6pt]
  \subfloat[Stage 1: H1, L3]{%
    \includegraphics[width=0.24\textwidth]{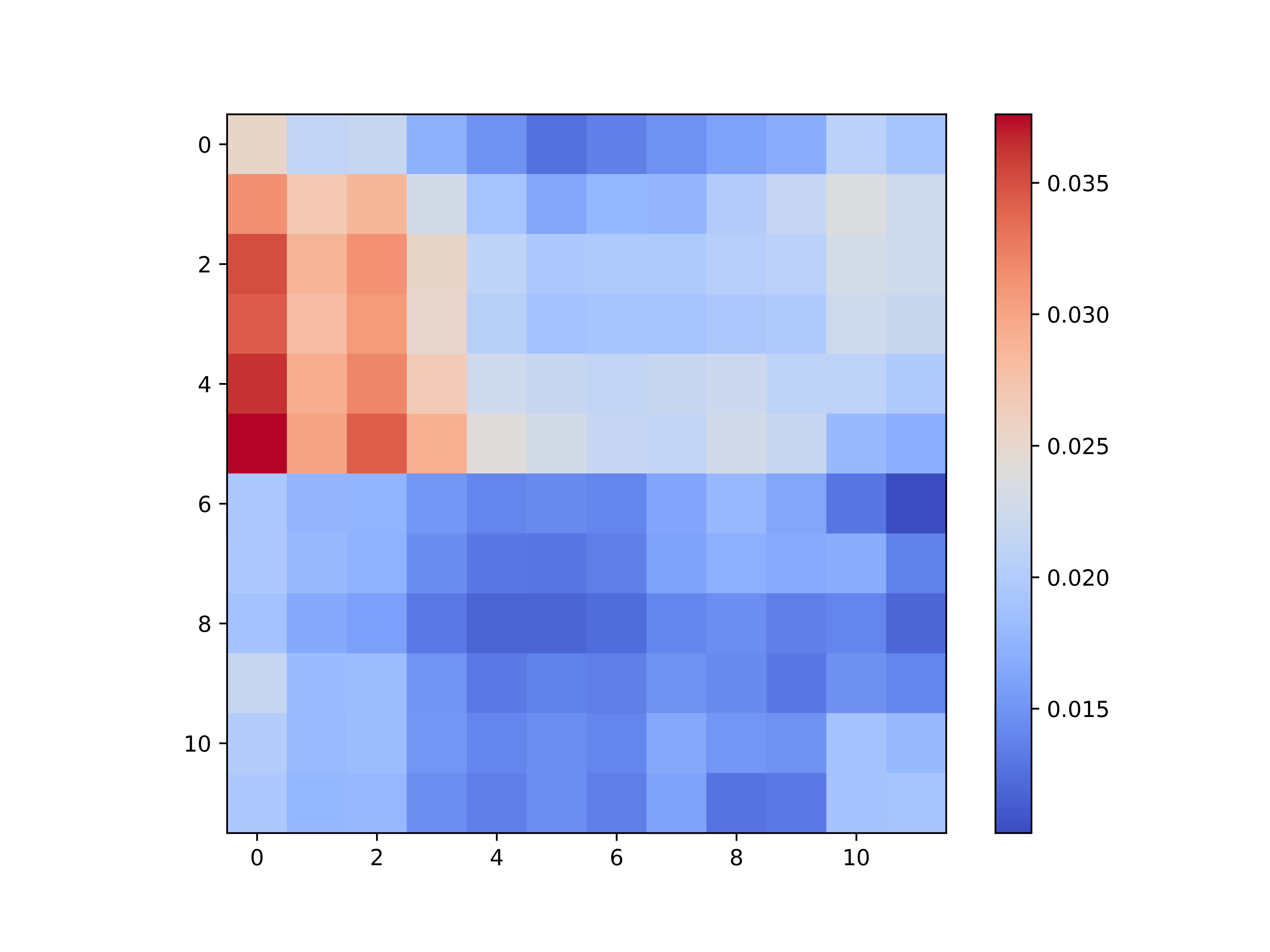}
    \label{fig:sub35}
  }
  \hfill
  \subfloat[Stage 1: H2, L3]{%
    \includegraphics[width=0.24\textwidth]{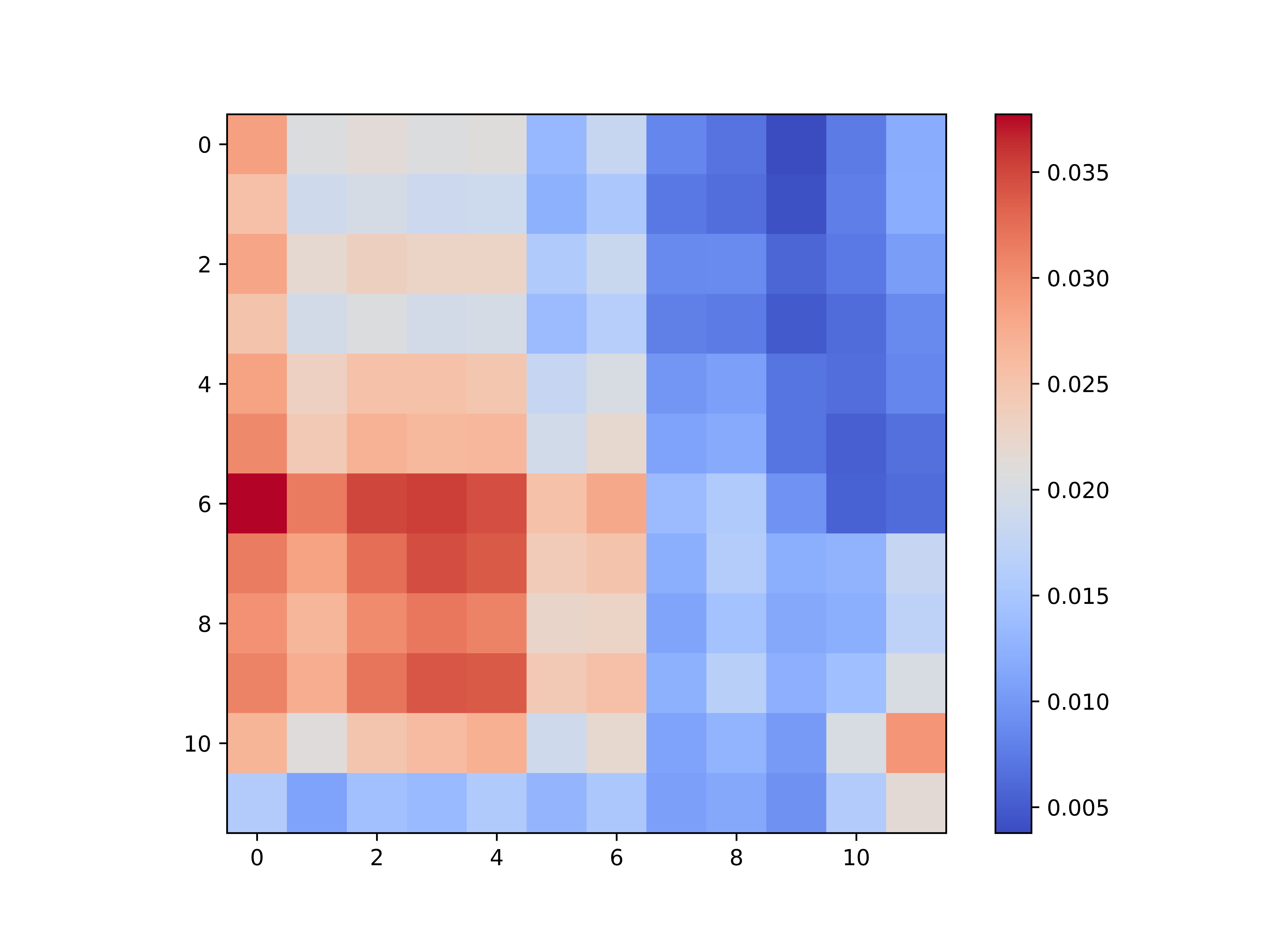}
    \label{fig:sub36}
  }
  \hfill
  \subfloat[Stage 2: H1, L3]{%
    \includegraphics[width=0.24\textwidth]{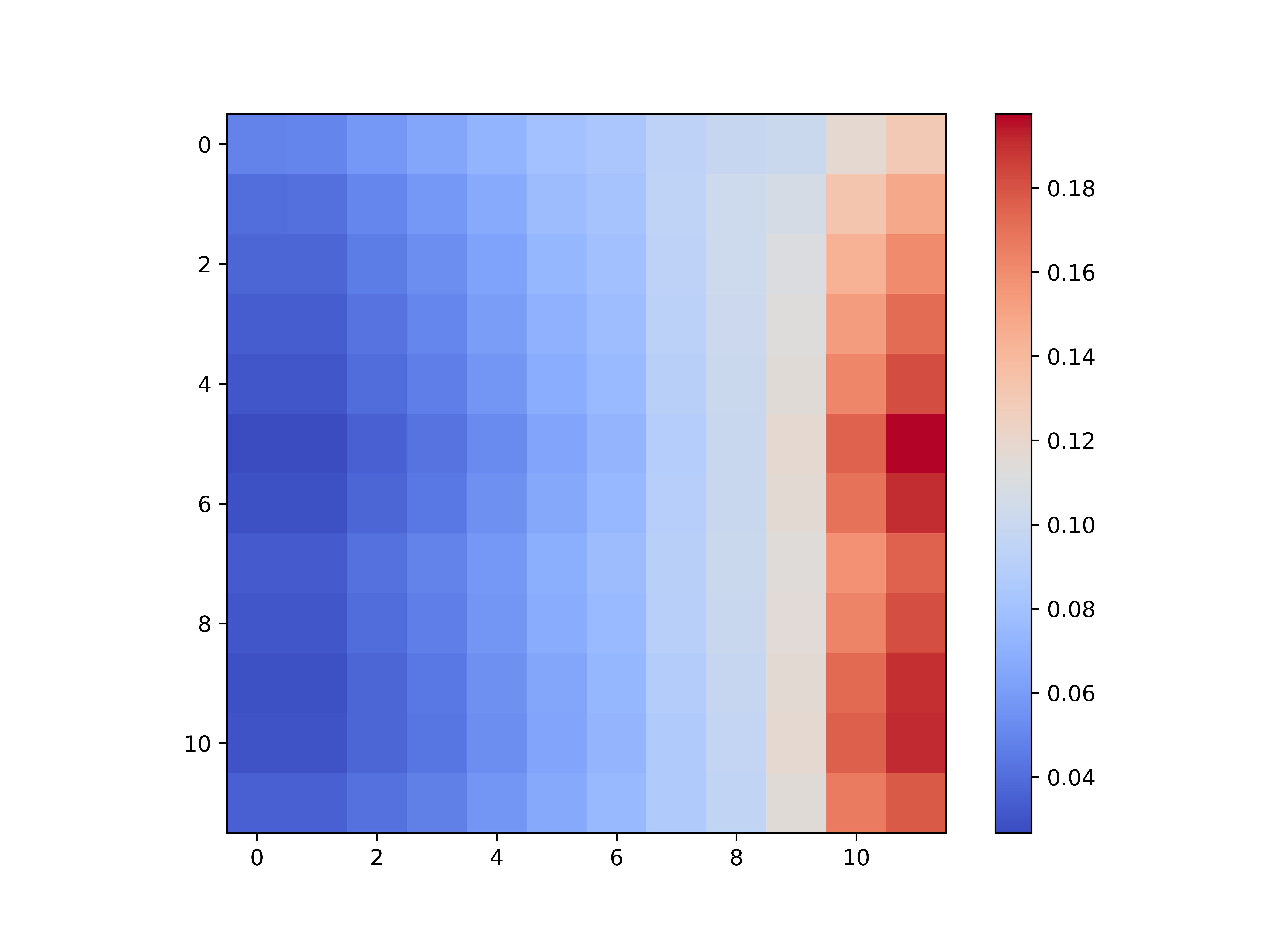}
    \label{fig:sub37}
  }
  \hfill
  \subfloat[Stage 2: H2, L3]{%
    \includegraphics[width=0.24\textwidth]{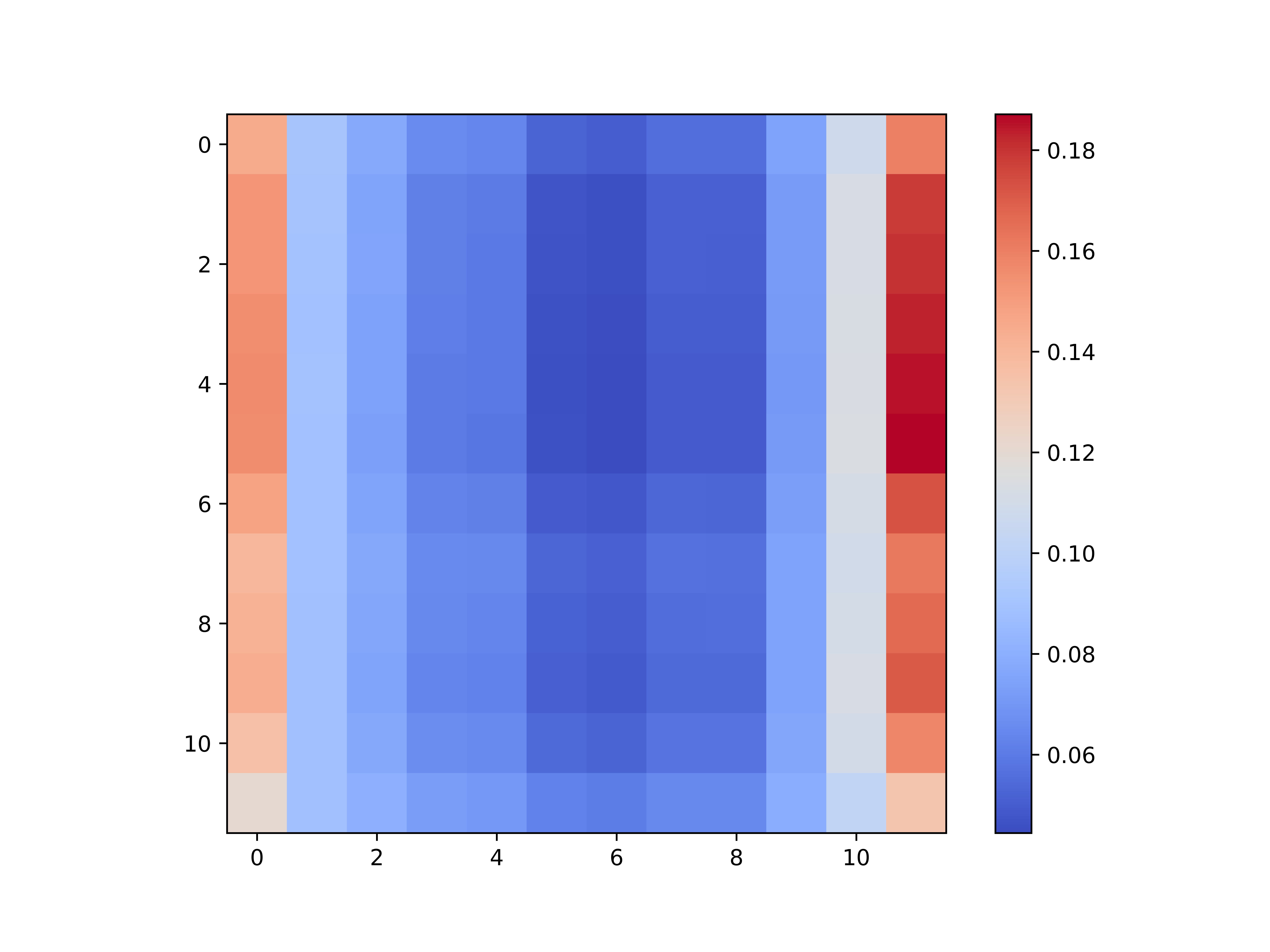}
    \label{fig:sub38}
  }
  \\[-6pt] 
  \subfloat[Stage 1: H1, L4]{%
    \includegraphics[width=0.24\textwidth]{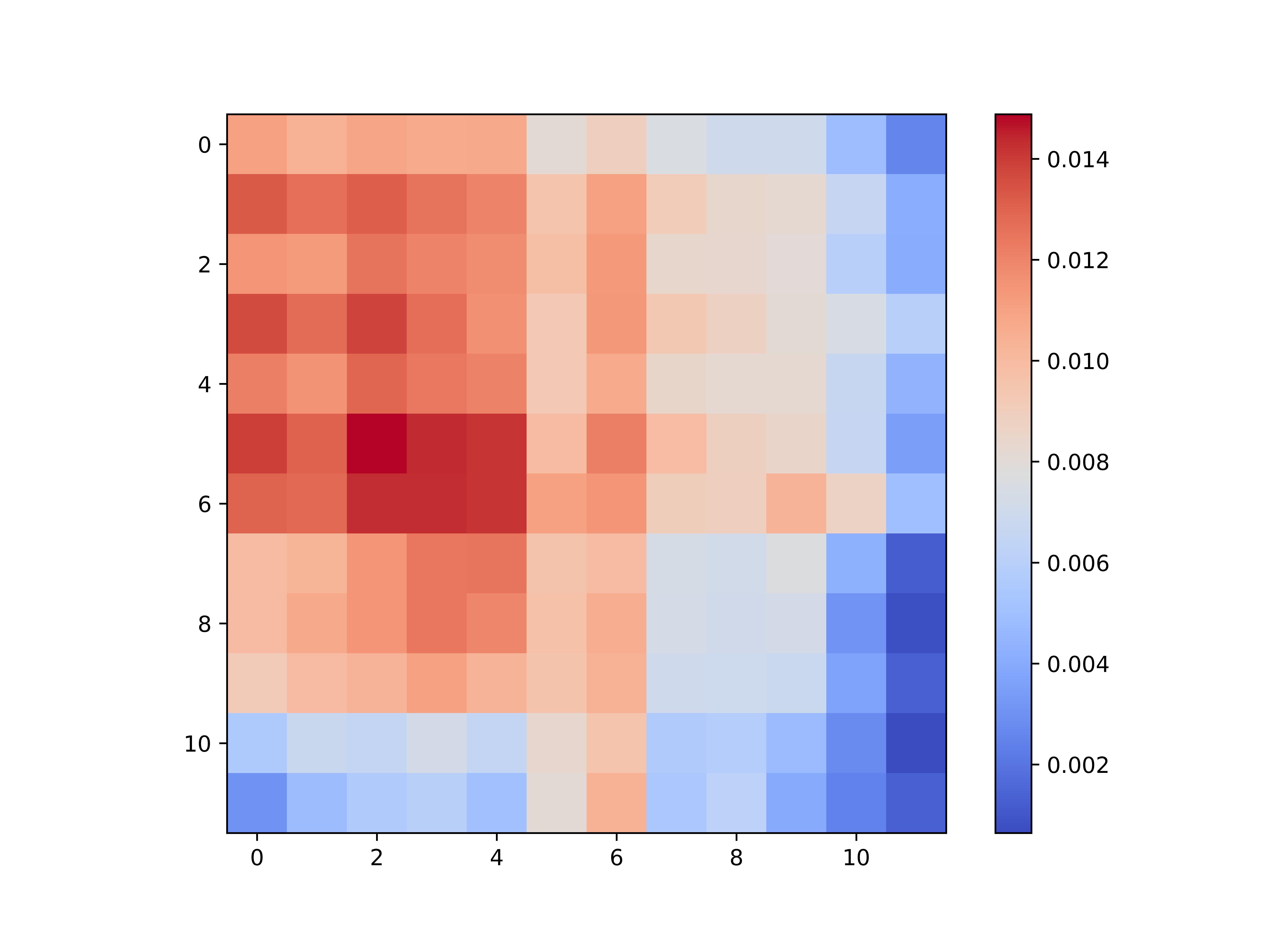}
    \label{fig:sub39}
  }
  \hfill
  \subfloat[Stage 1: H2, L4]{%
    \includegraphics[width=0.24\textwidth]{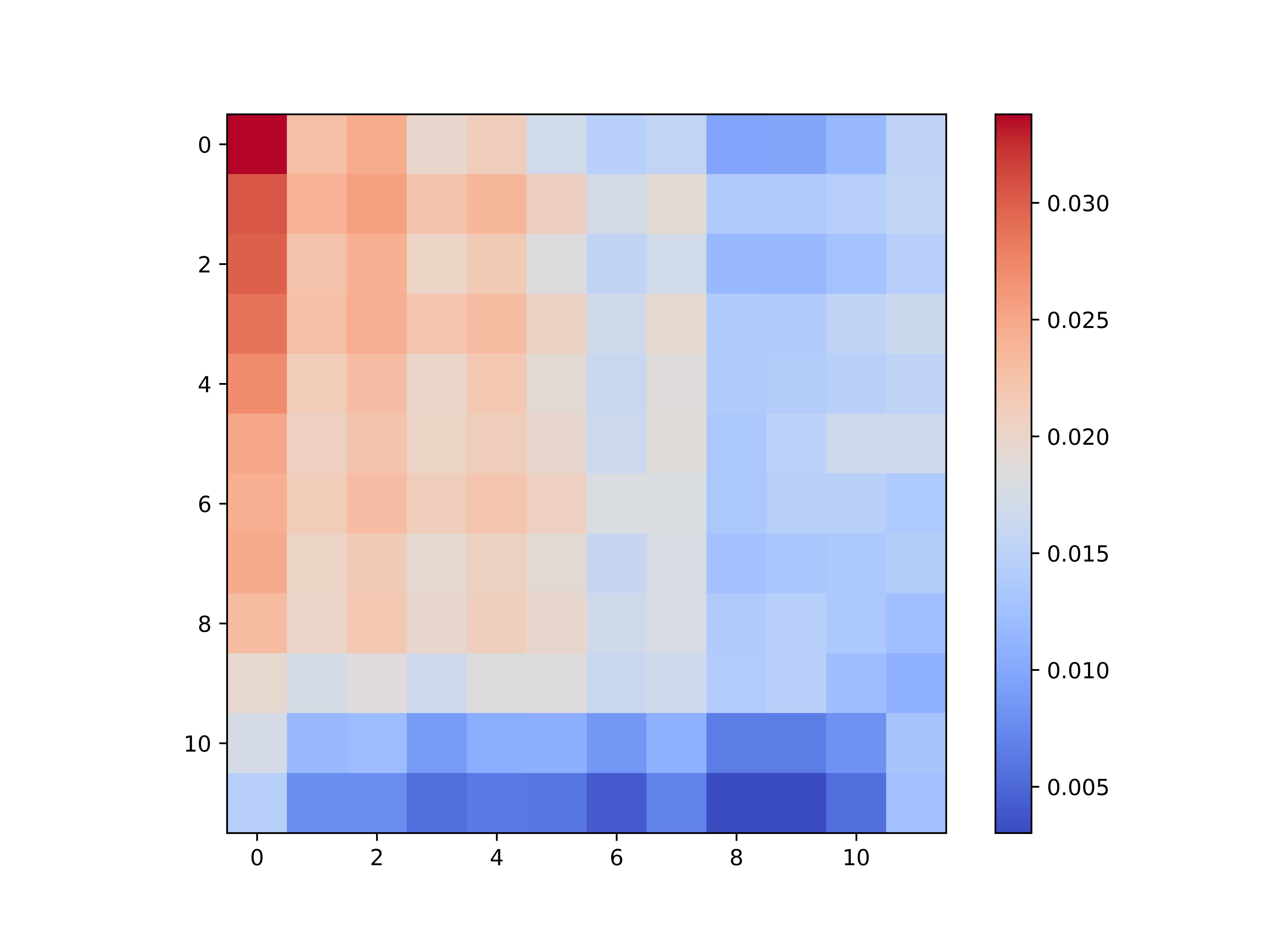}
    \label{fig:sub40}
  }
  \hfill
  \subfloat[Stage 2: H1, L4]{%
    \includegraphics[width=0.24\textwidth]{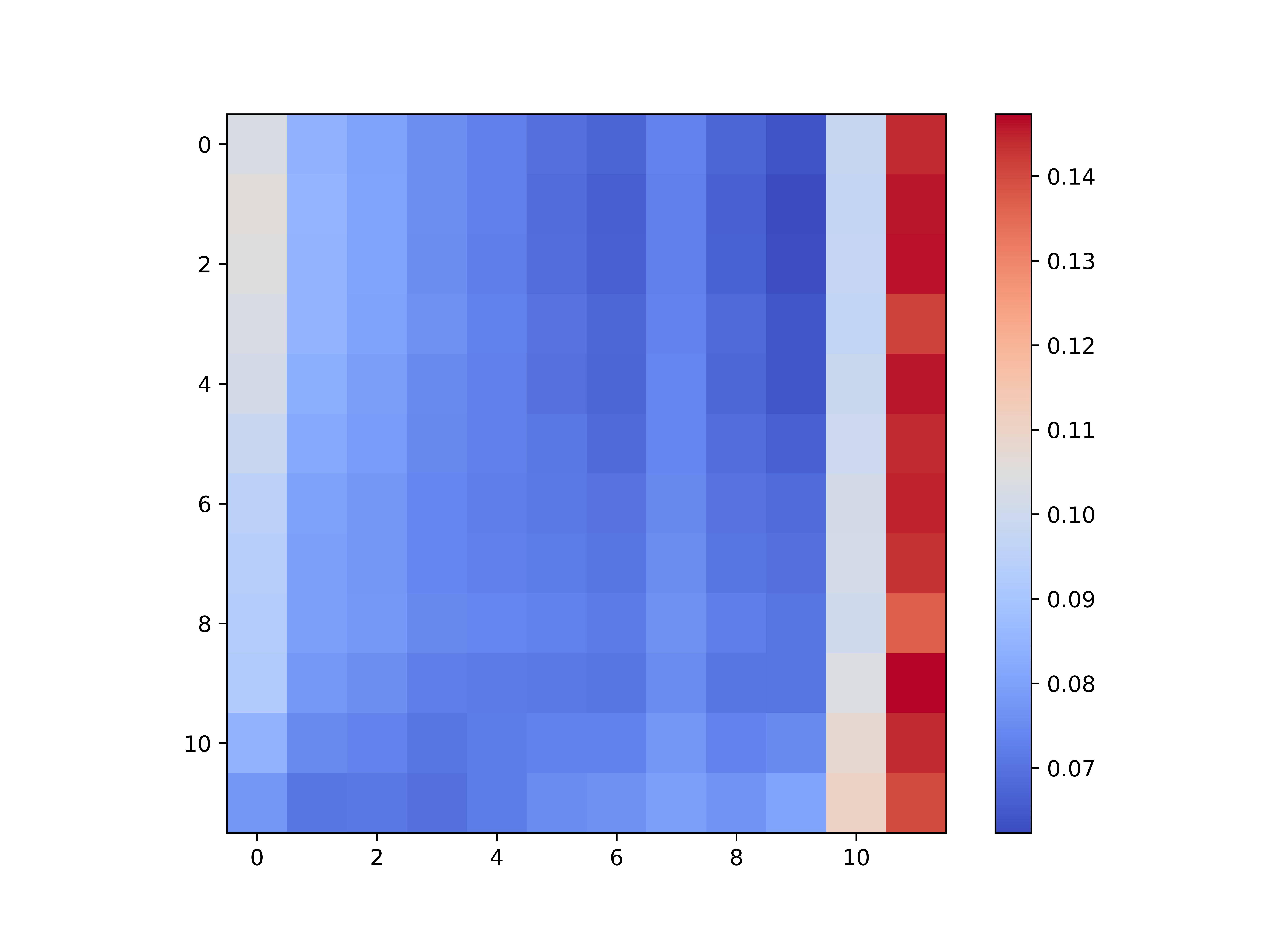}
    \label{fig:sub41}
  }
  \hfill
  \subfloat[Stage 2: H2, L4]{%
    \includegraphics[width=0.24\textwidth]{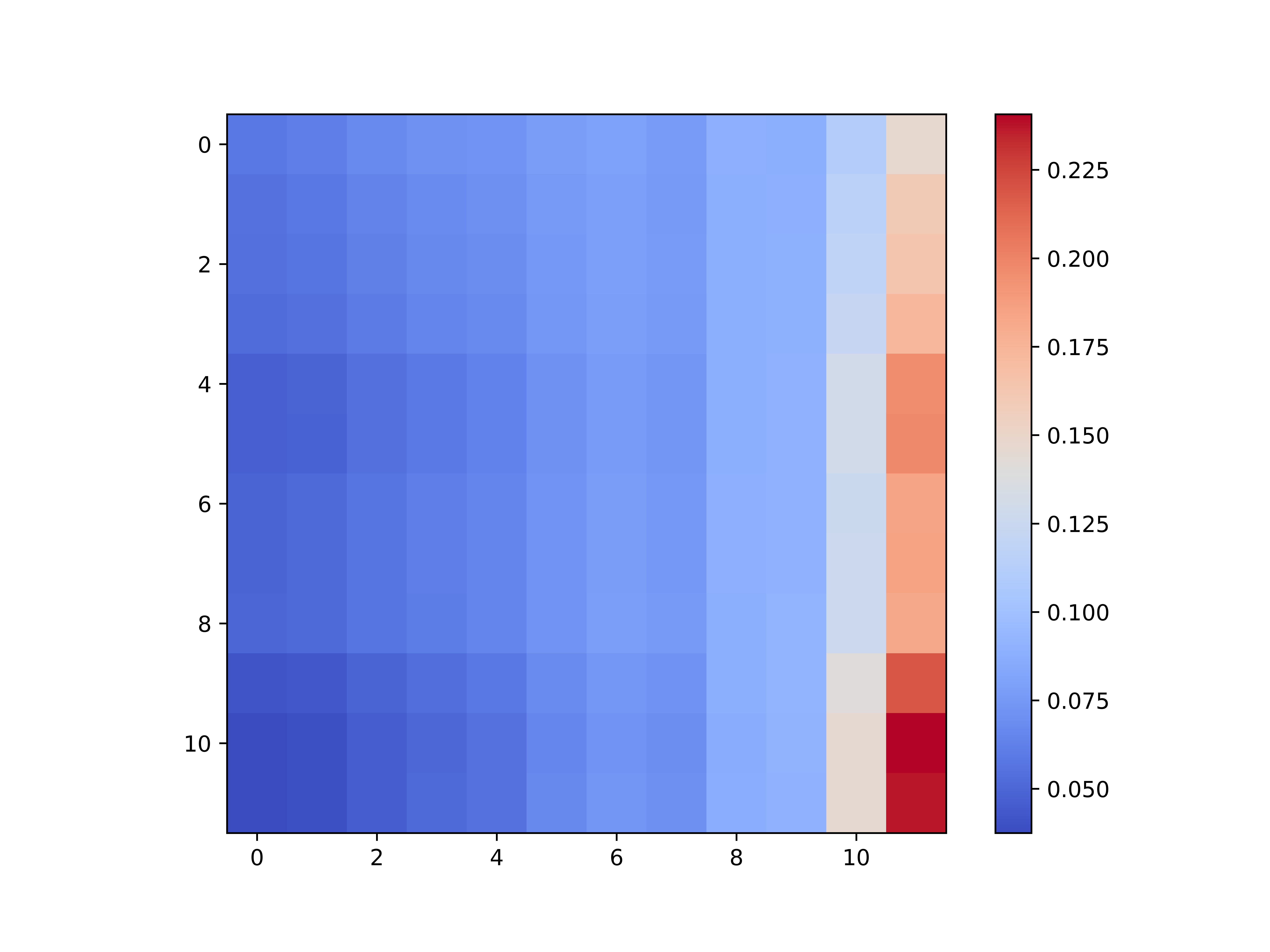}
    \label{fig:sub42}
  }
  \\[-6pt] 
\subfloat[Stage 1: H1, L5]{%
\includegraphics[width=0.24\textwidth]{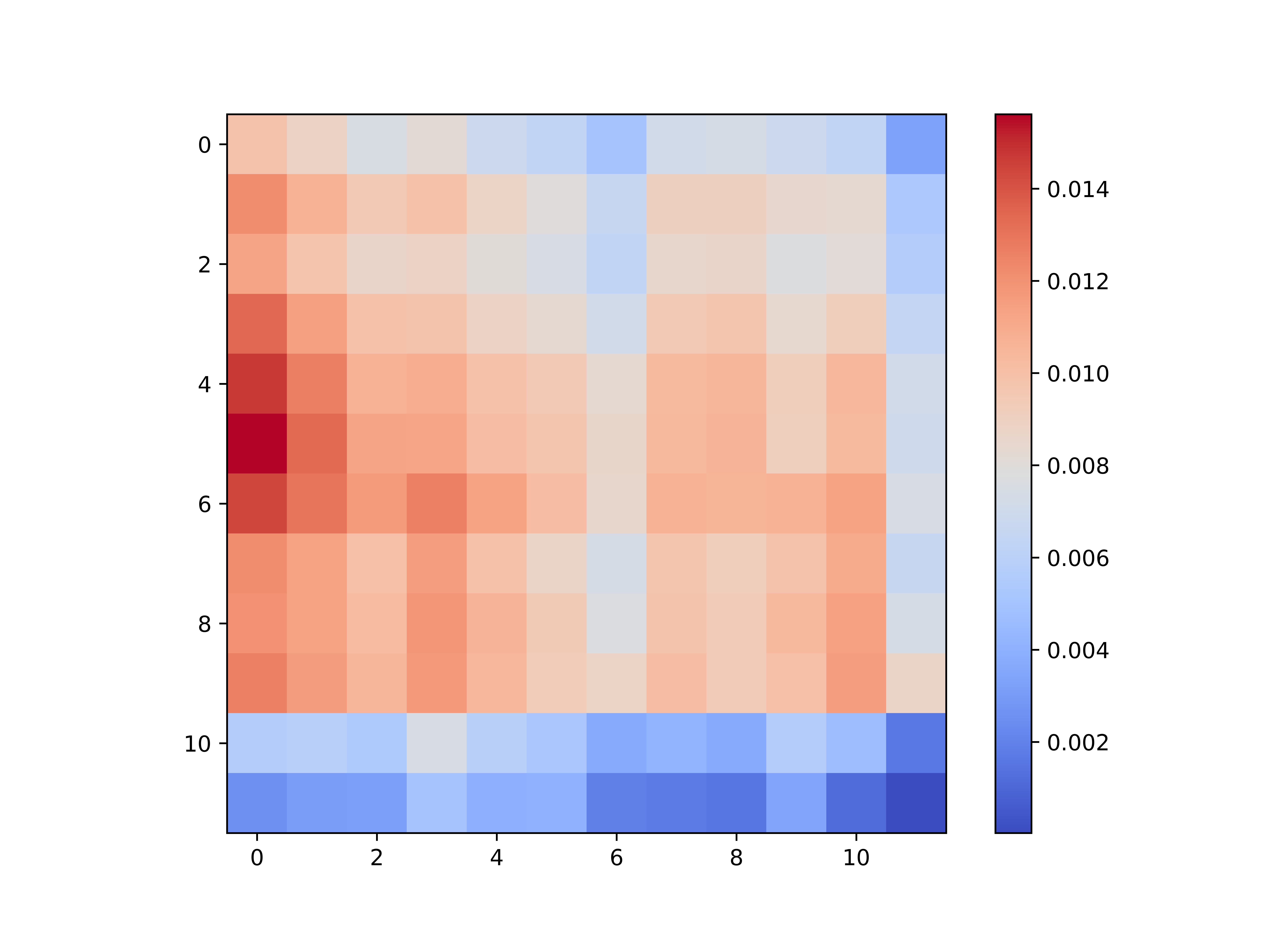}
\label{fig:sub43}
}
\hfill
\subfloat[Stage 1: H2, L5]{%
\includegraphics[width=0.24\textwidth]{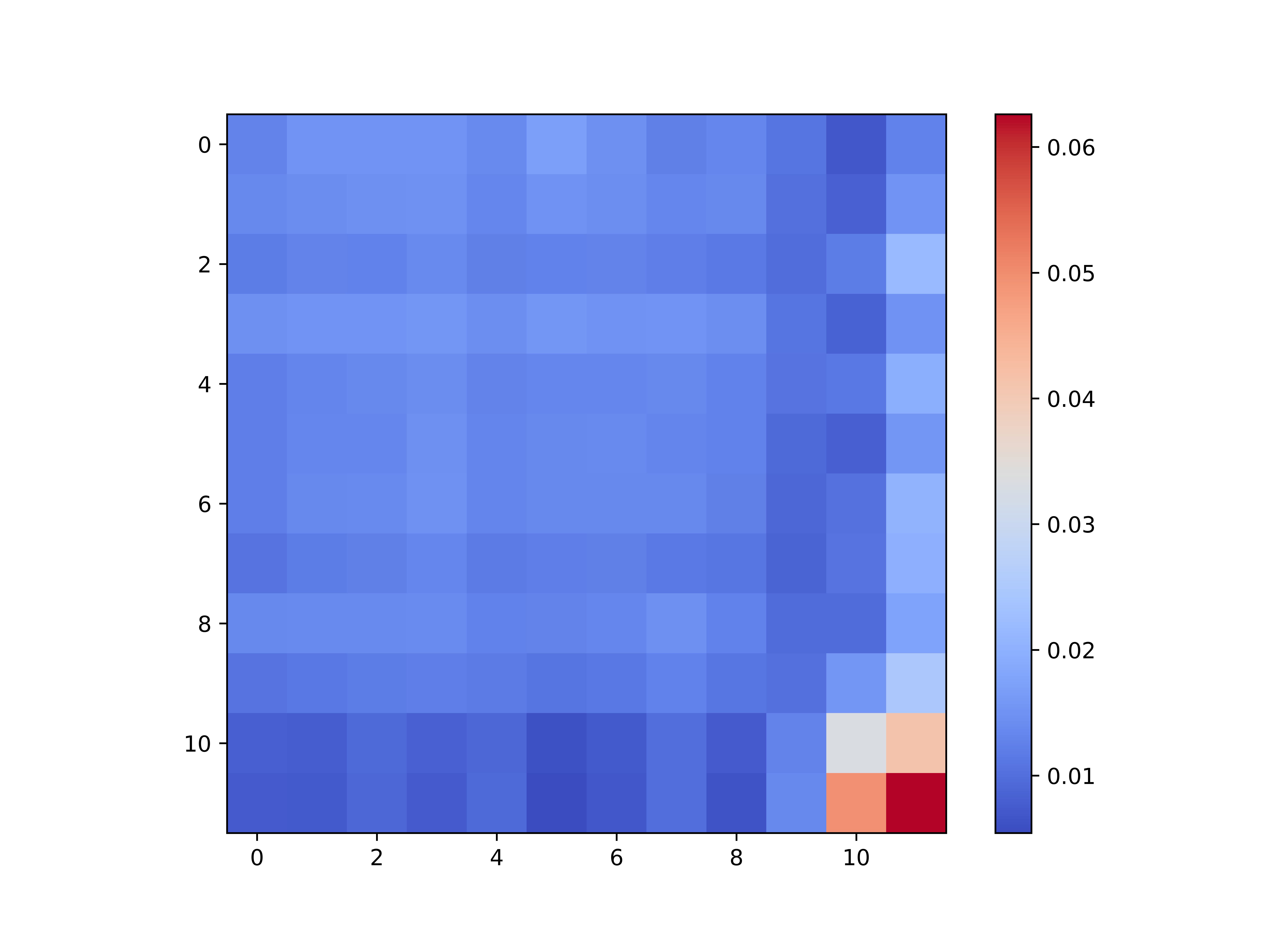}
\label{fig:sub44}
}
\hfill
\subfloat[Stage 2: H1, L5]{%
\includegraphics[width=0.24\textwidth]{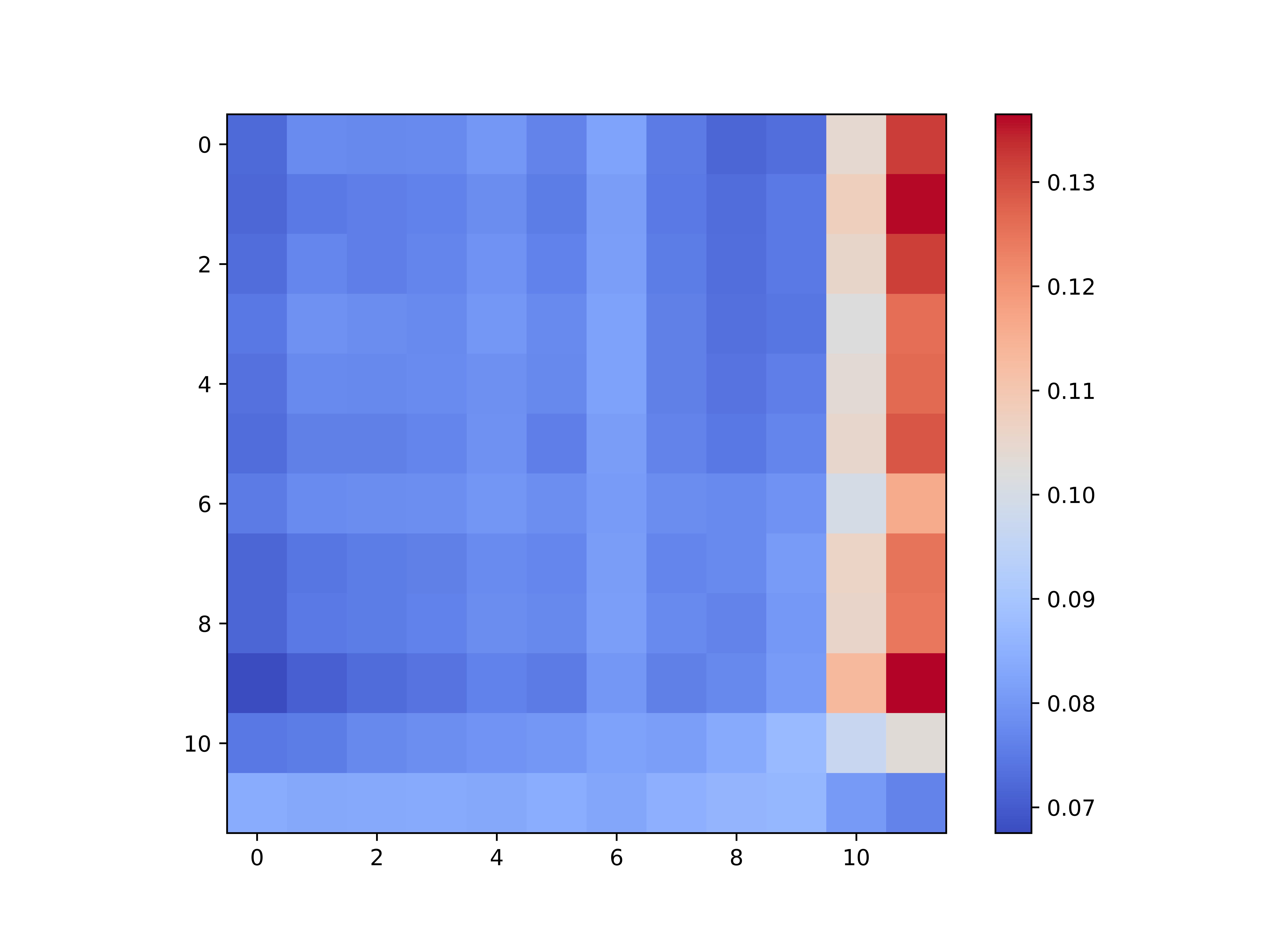}
\label{fig:sub45}
}
\hfill
\subfloat[Stage 2: H2, L5]{%
\includegraphics[width=0.24\textwidth]{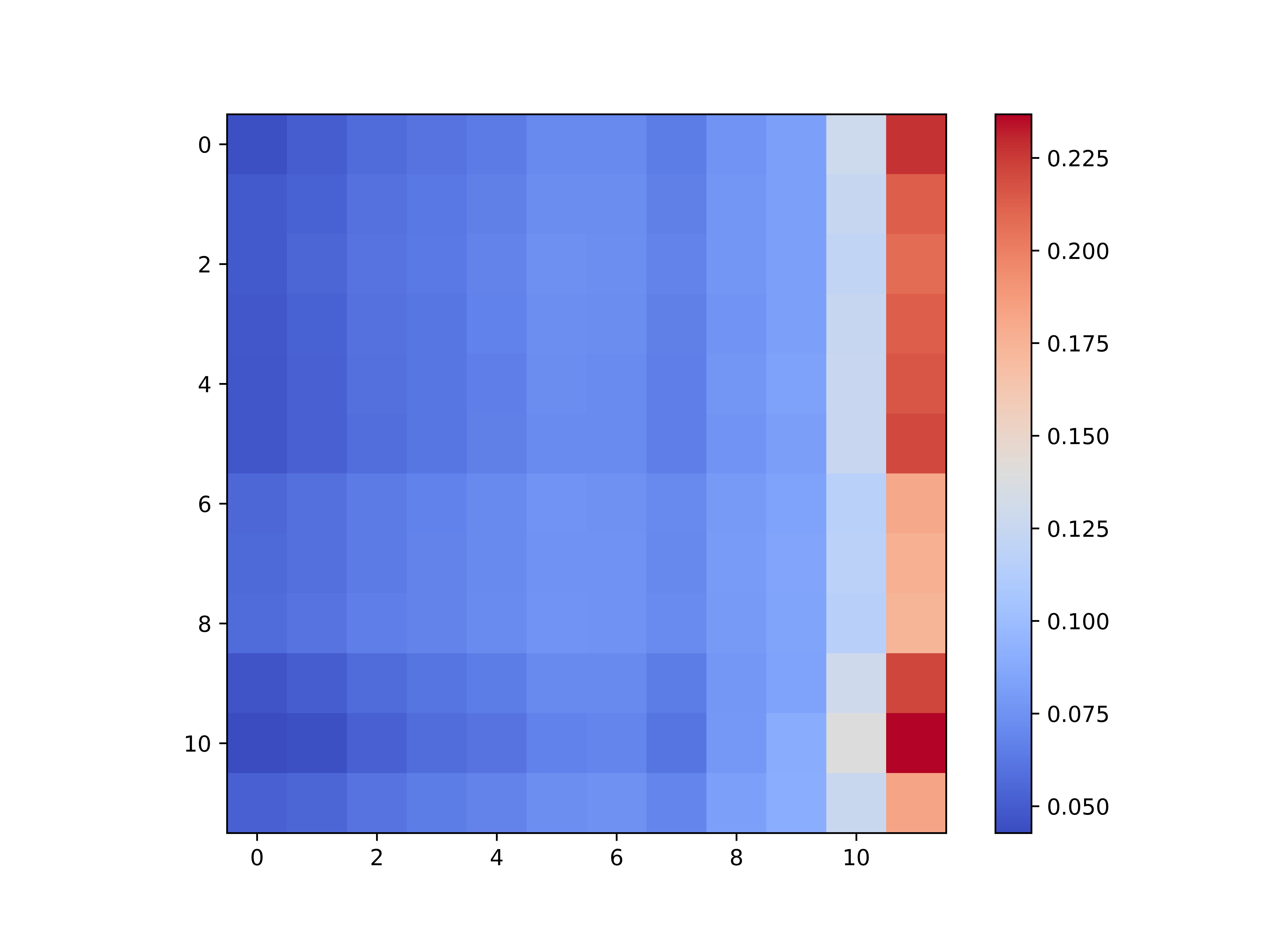}
\label{fig:sub46}
}
\\[-6pt]
\subfloat[Stage 1: H1, L6]{%
\includegraphics[width=0.24\textwidth]{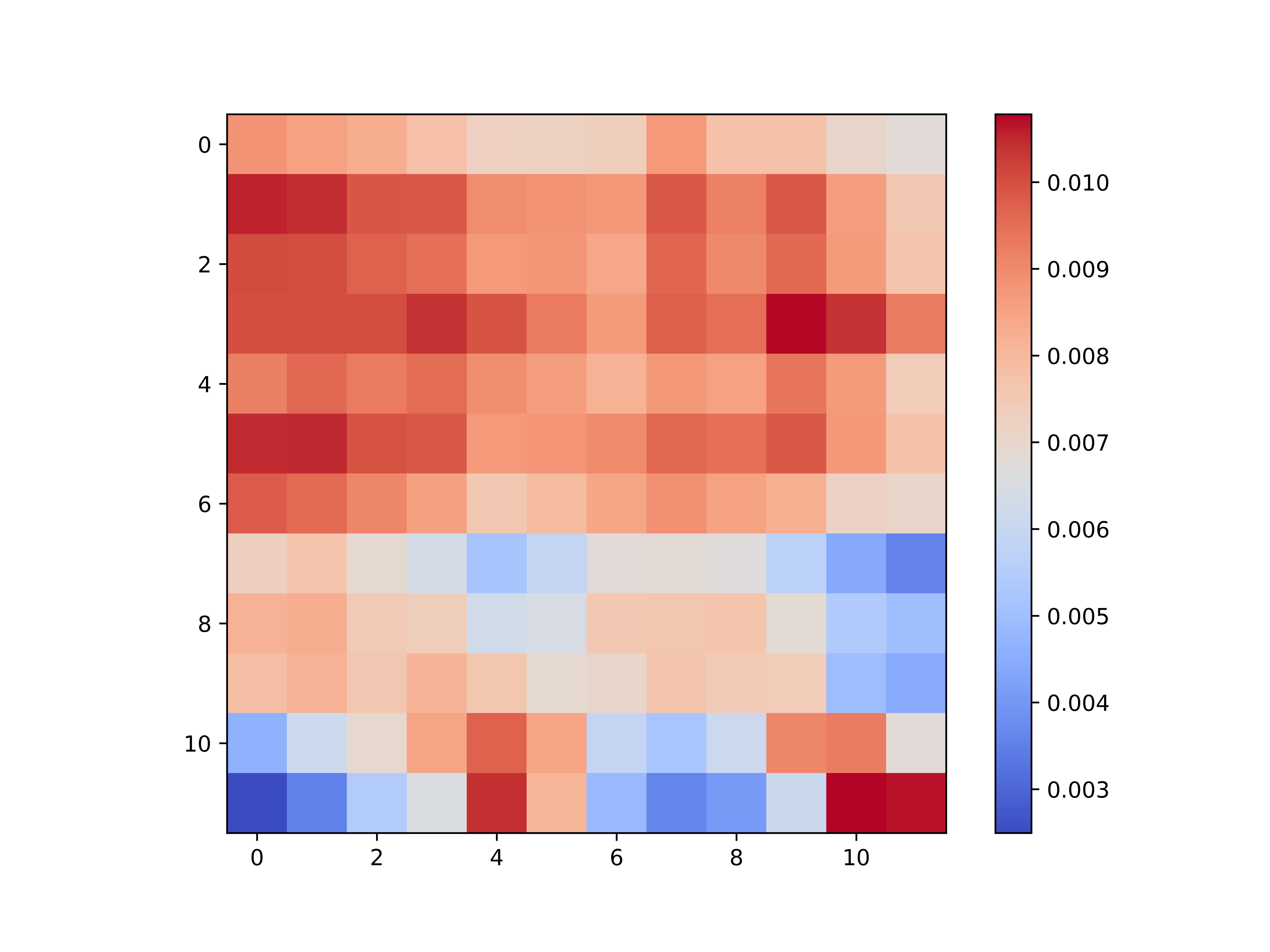}
\label{fig:sub47}
}
\hfill
\subfloat[Stage 1: H2, L6]{%
\includegraphics[width=0.24\textwidth]{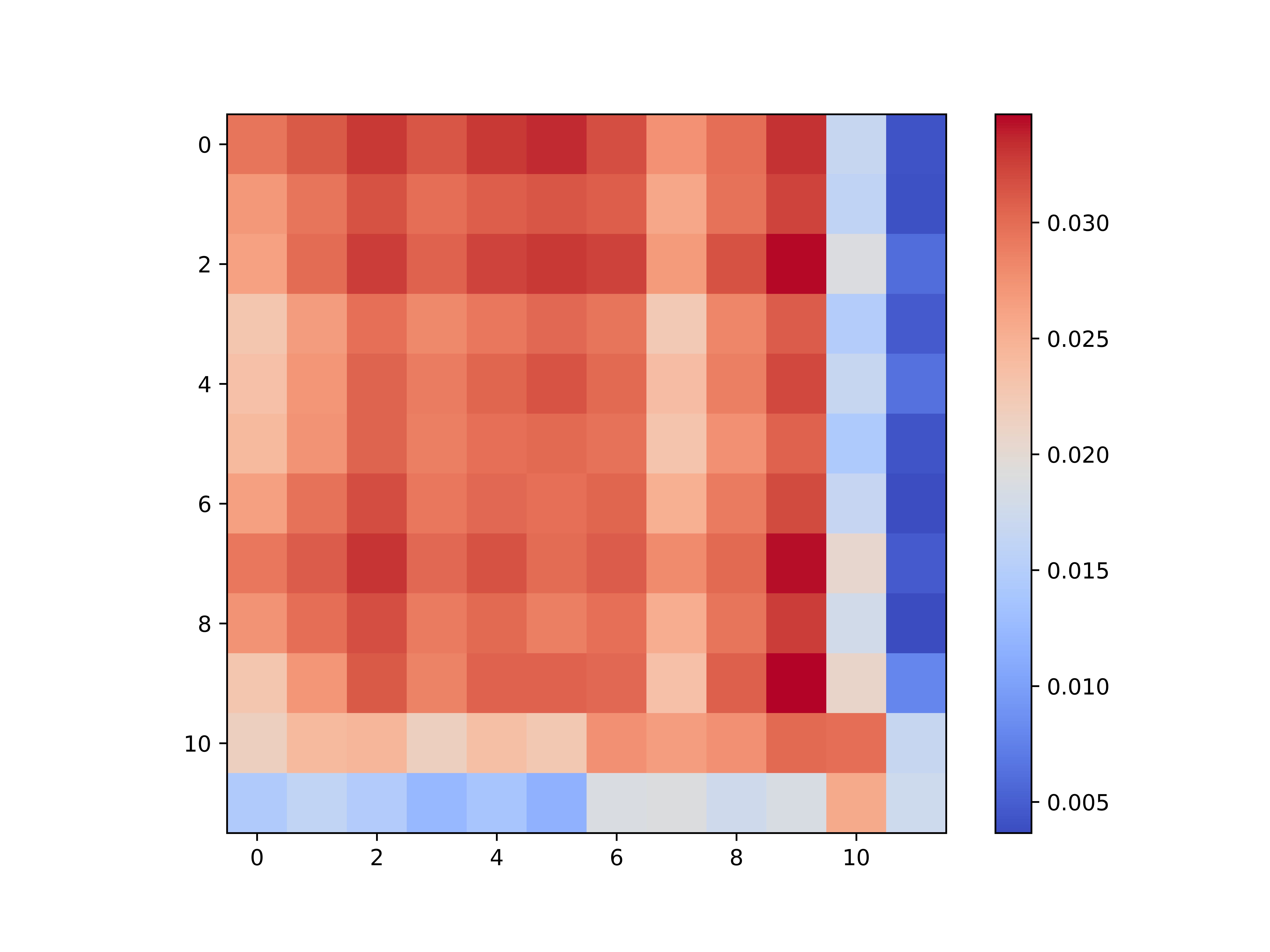}
\label{fig:sub48}
}
\hfill
\subfloat[Stage 2: H1, L6]{%
\includegraphics[width=0.24\textwidth]{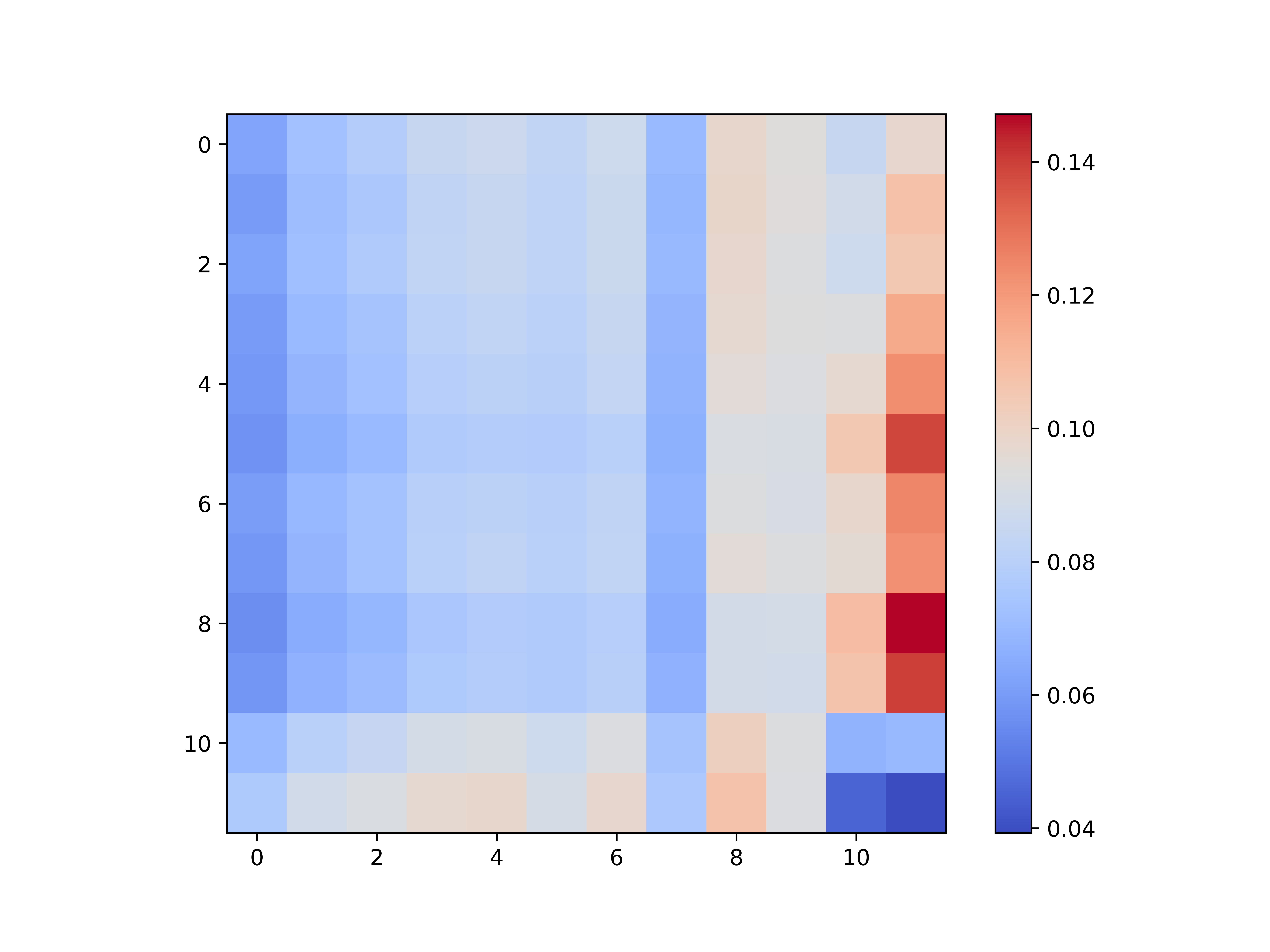}
\label{fig:sub49}
}
\hfill
\subfloat[Stage 2: H2, L6]{%
\includegraphics[width=0.24\textwidth]{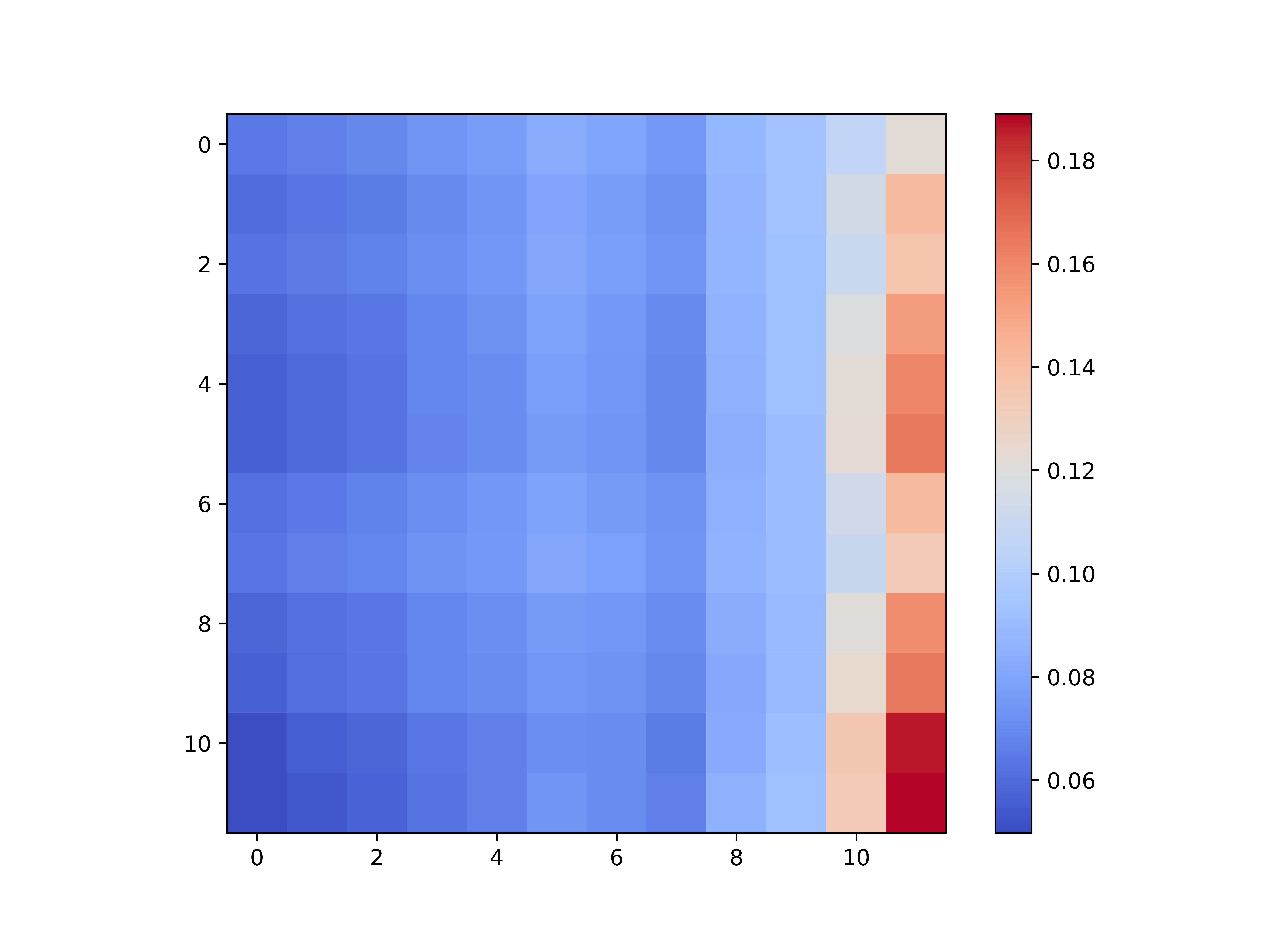}
\label{fig:sub50}
} 
\end{figure*}

\begin{figure*}[htbp]
  \centering
    \caption{Attention scores in \textbf{ReDASA} across layers: horizontal axis shows more heads, vertical indicates deeper layers. Note: H$n$ = Head $n$, L$n$ = Layer $n$}\label{fig:attentionmap_redasa}
  \subfloat[Stage 1: H1, L1]{%
    \includegraphics[width=0.24\textwidth]{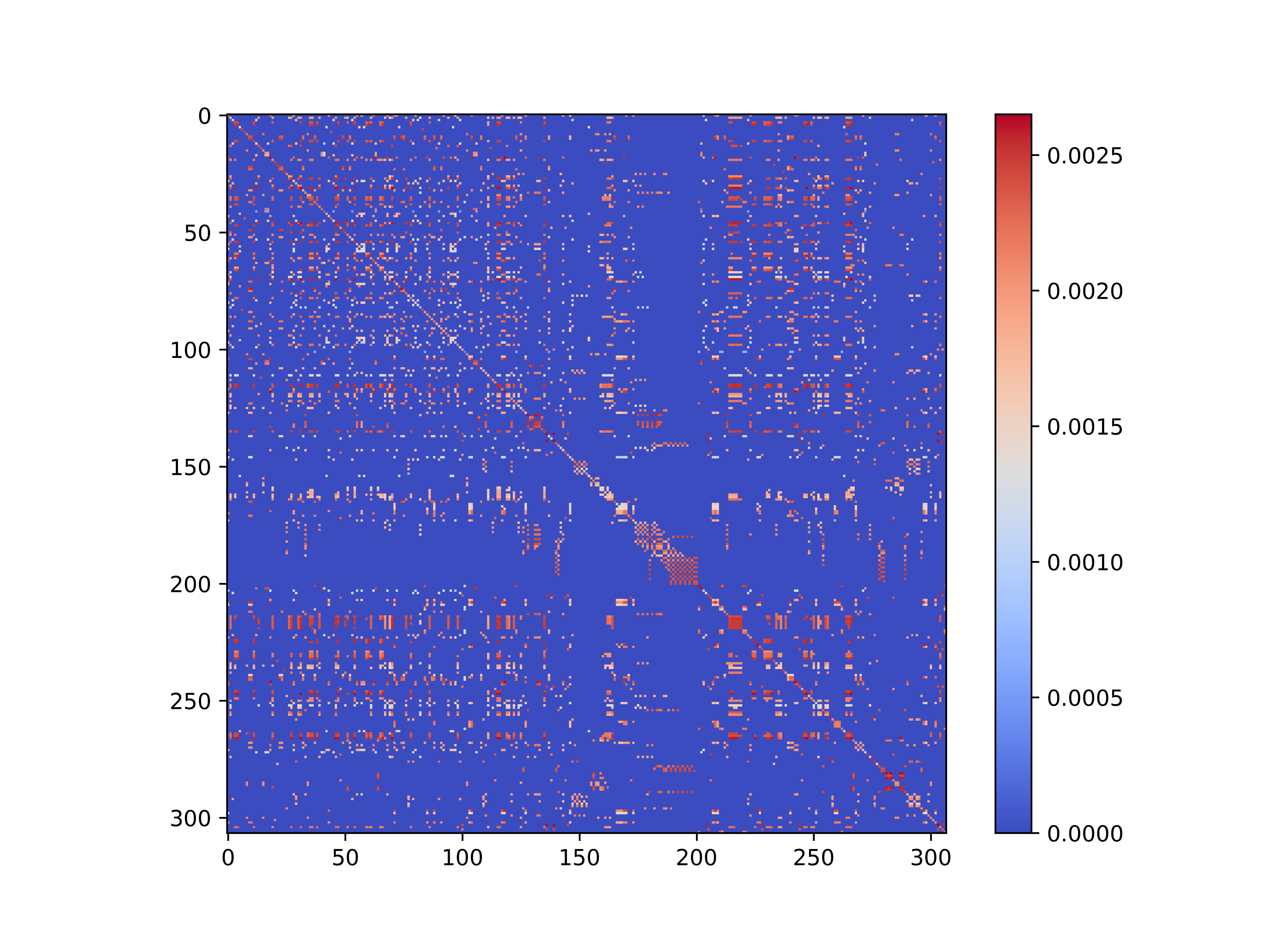}
    \label{fig:sub51}
  }
  \hfill
  \subfloat[Stage 1: H2, L1]{%
    \includegraphics[width=0.24\textwidth]{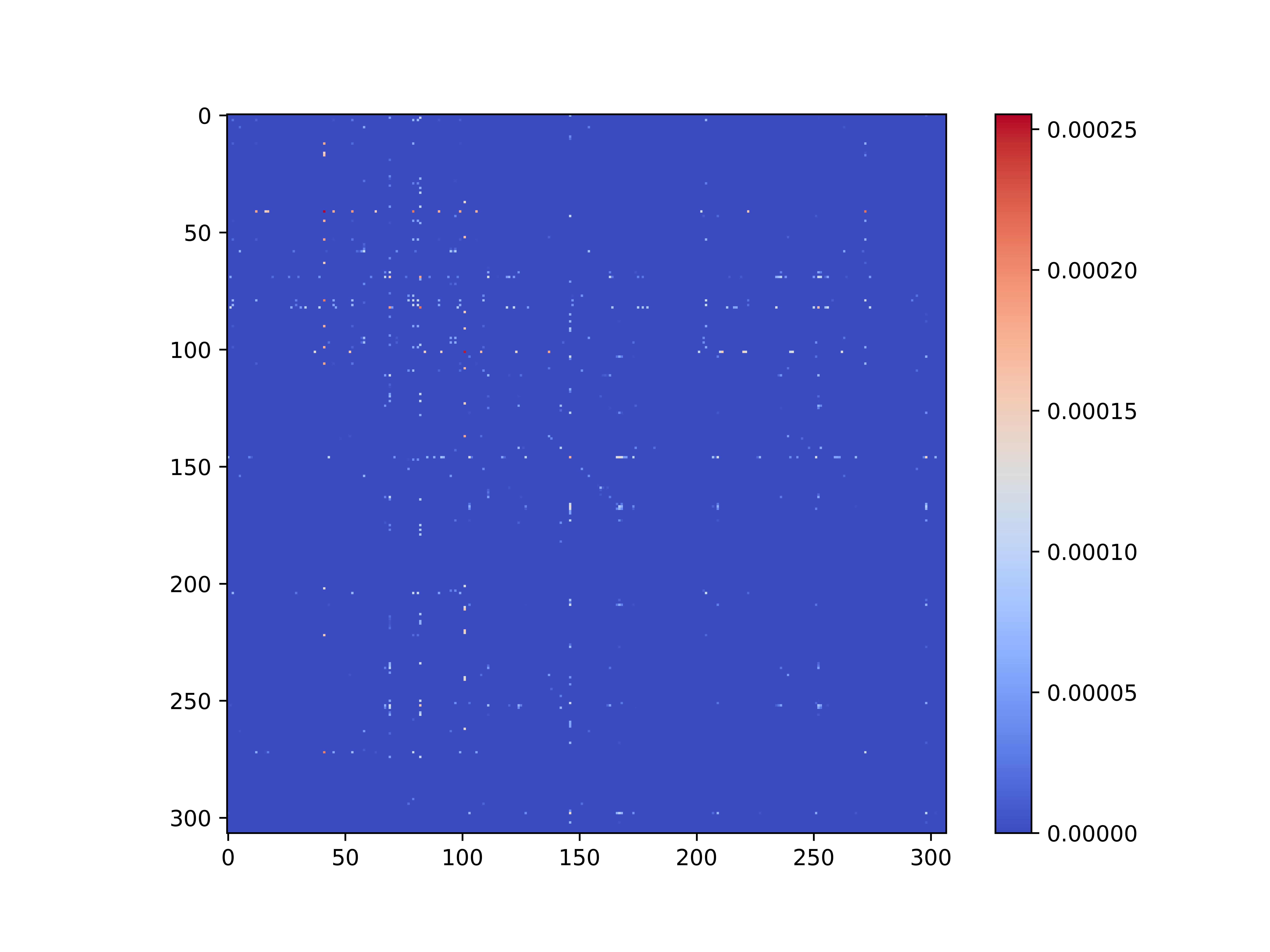}
    \label{fig:sub52}
  }
  \hfill
  \subfloat[Stage 2: H3, L1]{%
    \includegraphics[width=0.24\textwidth]{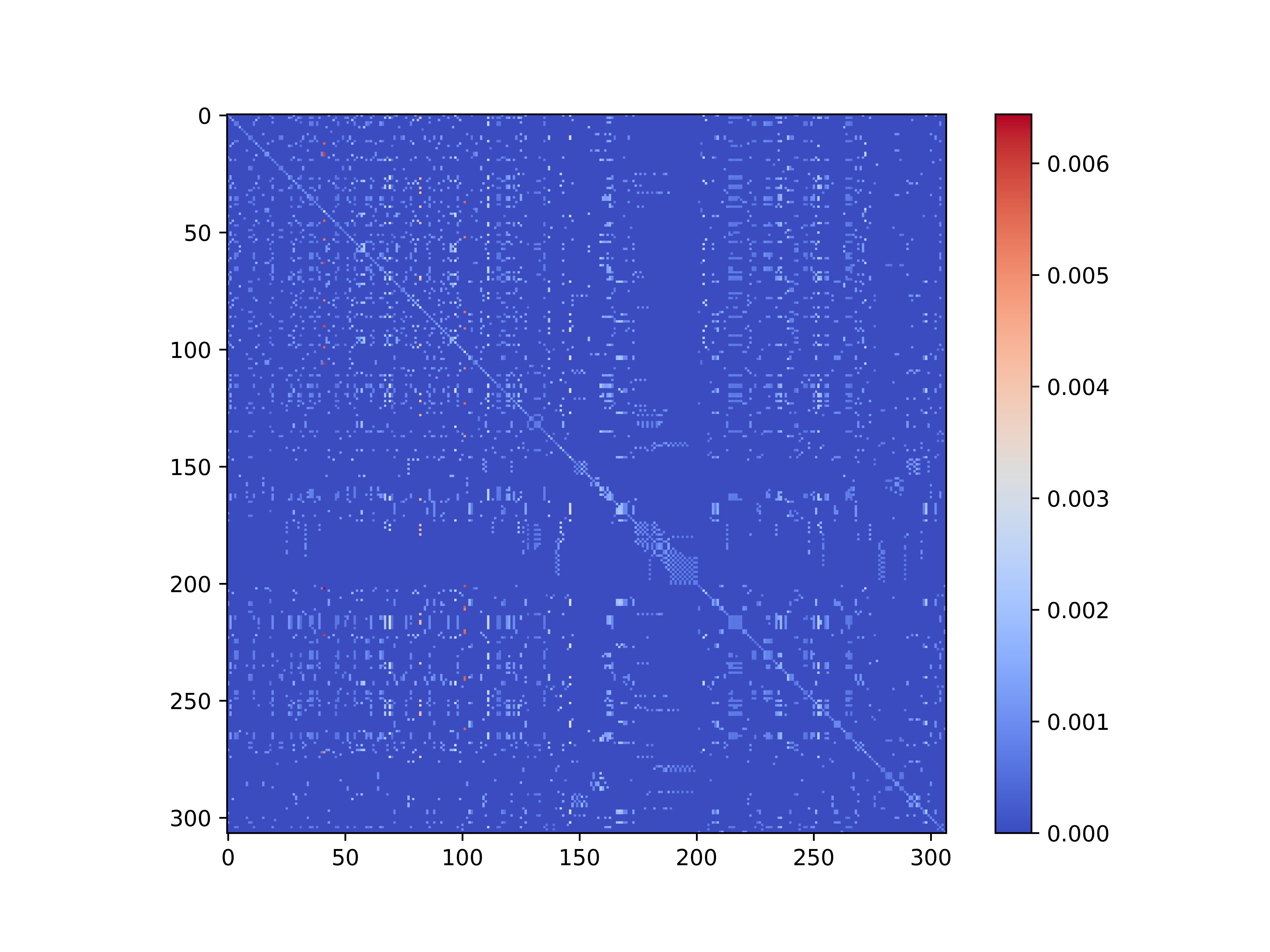}
    \label{fig:sub53}
  }
  \hfill
  \subfloat[Stage 2: H4, L1]{%
    \includegraphics[width=0.24\textwidth]{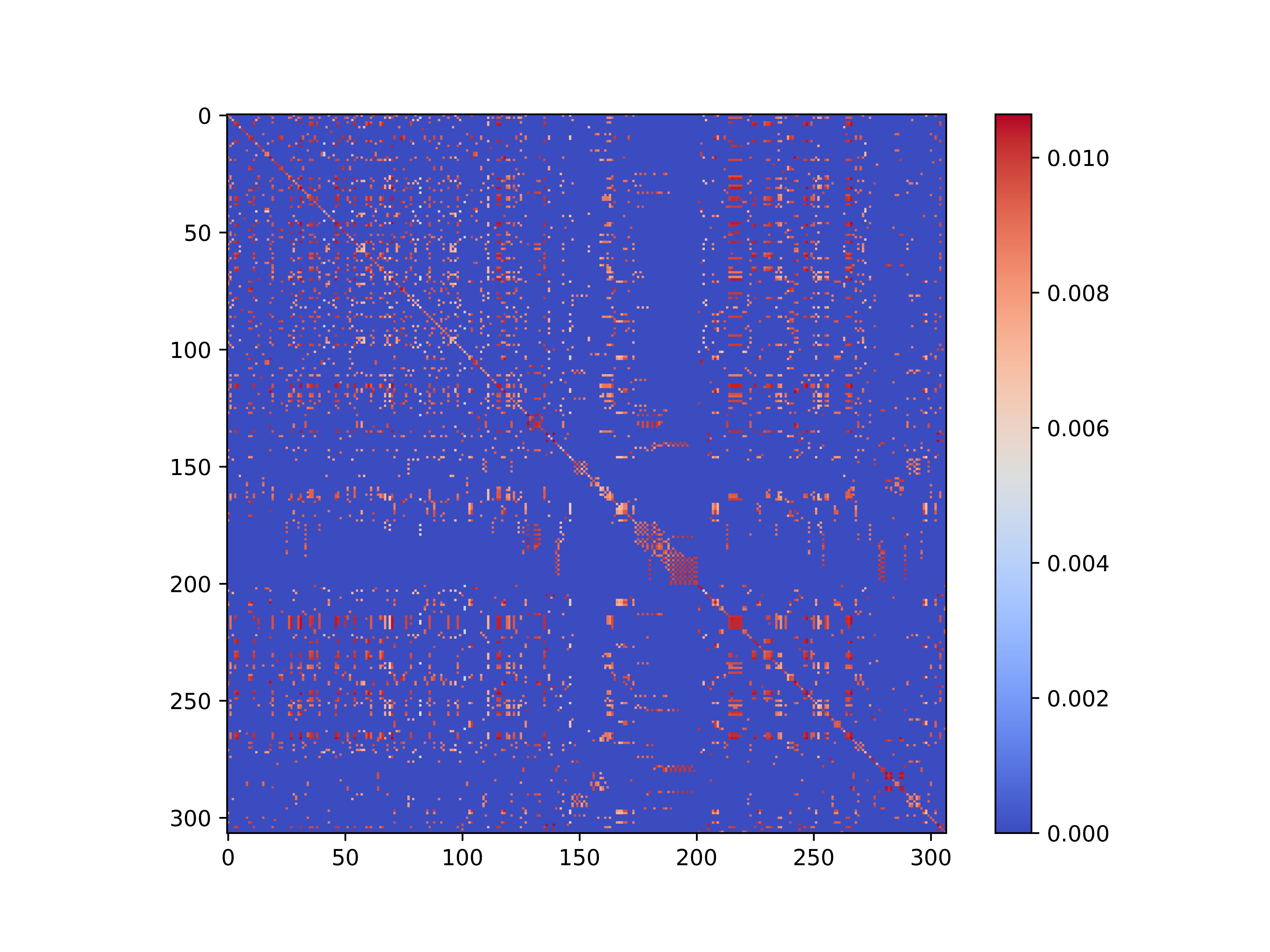}
    \label{fig:sub54}
  }
  \\[-6pt] 
  \subfloat[Stage 1: H1, L2]{%
    \includegraphics[width=0.24\textwidth]{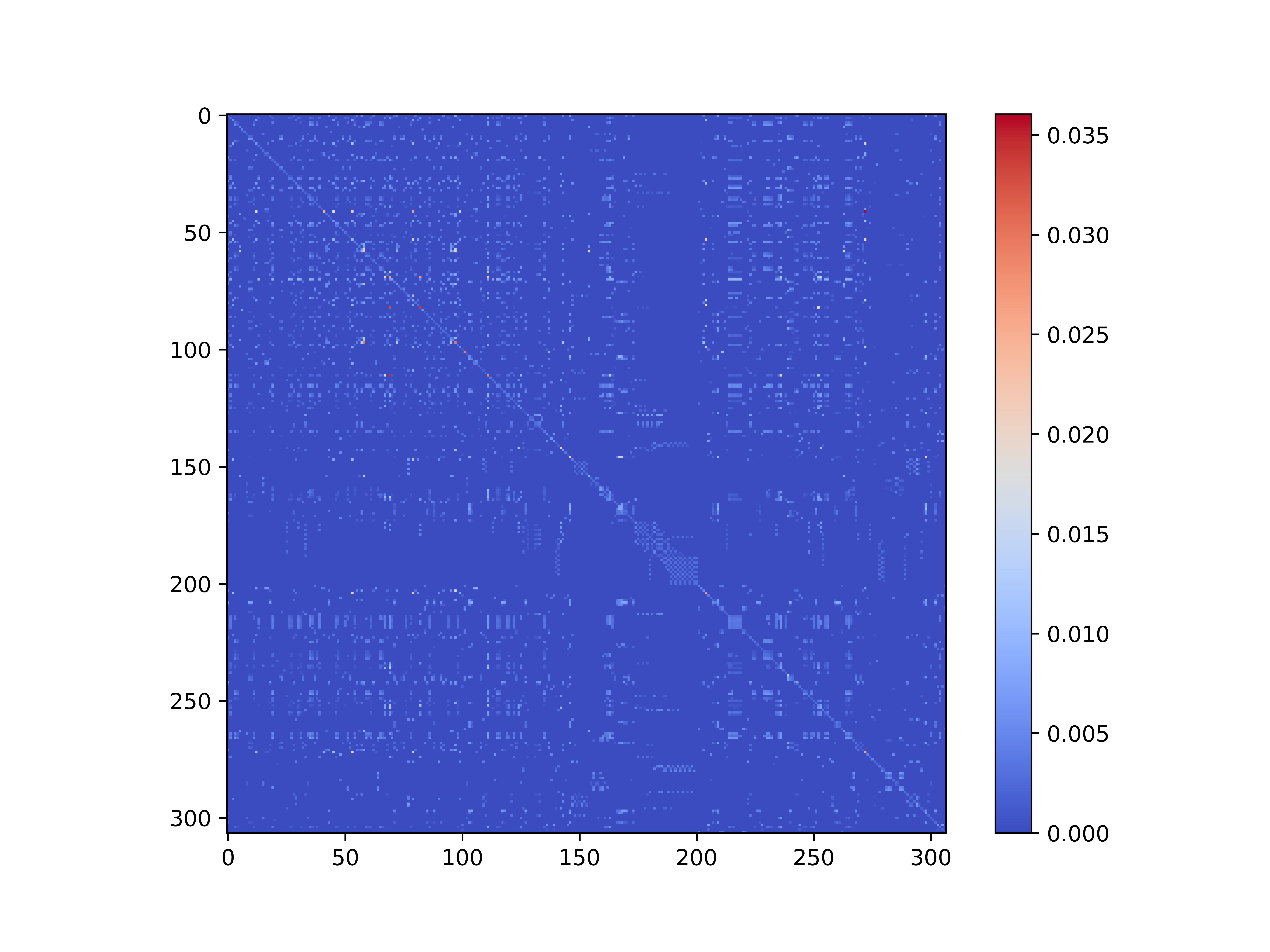}
    \label{fig:sub55}
  }
  \hfill
  \subfloat[Stage 1: H2, L2]{%
    \includegraphics[width=0.24\textwidth]{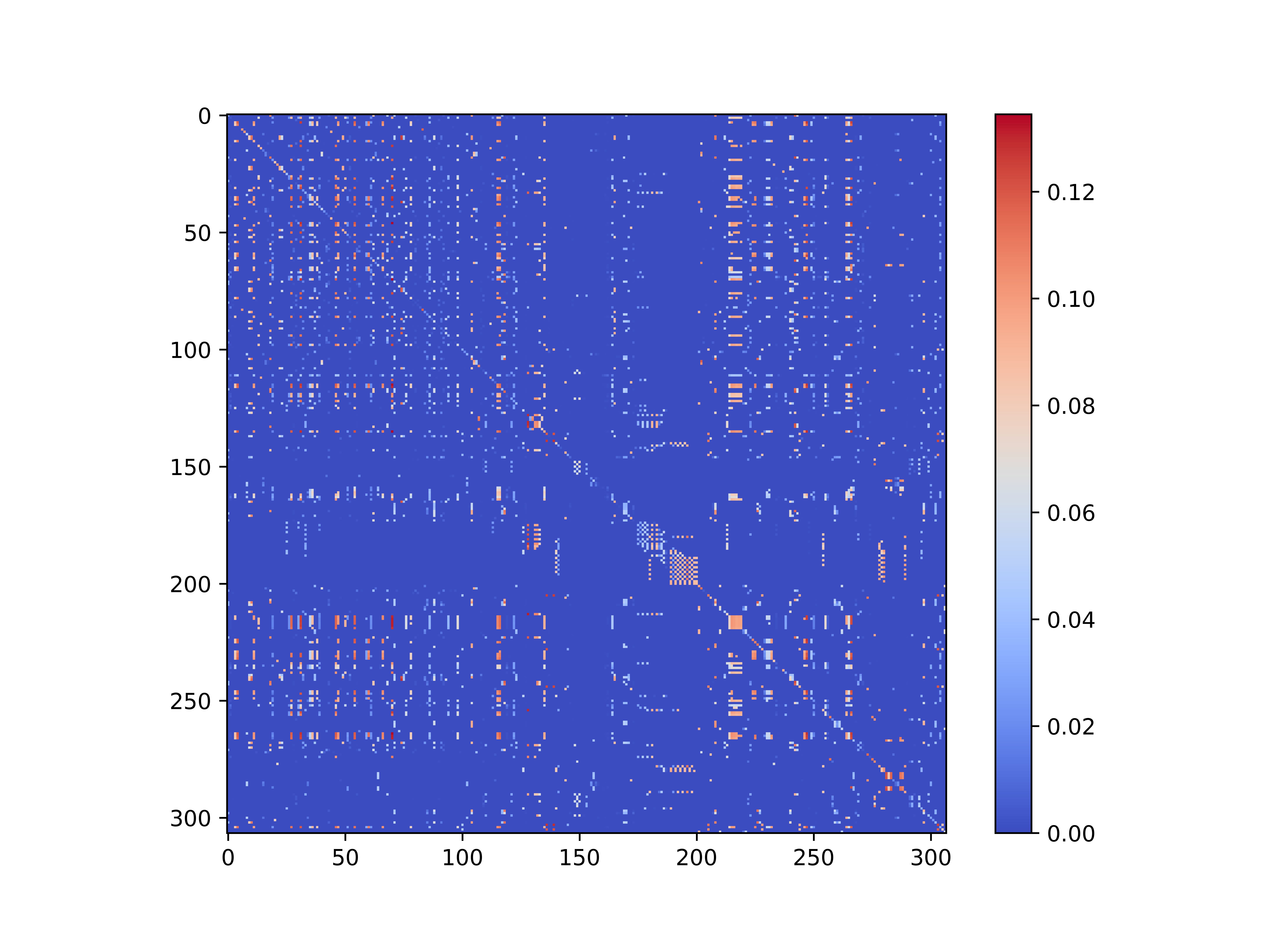}
    \label{fig:sub56}
  }
  \hfill
  \subfloat[Stage 2: H3, L2]{%
    \includegraphics[width=0.24\textwidth]{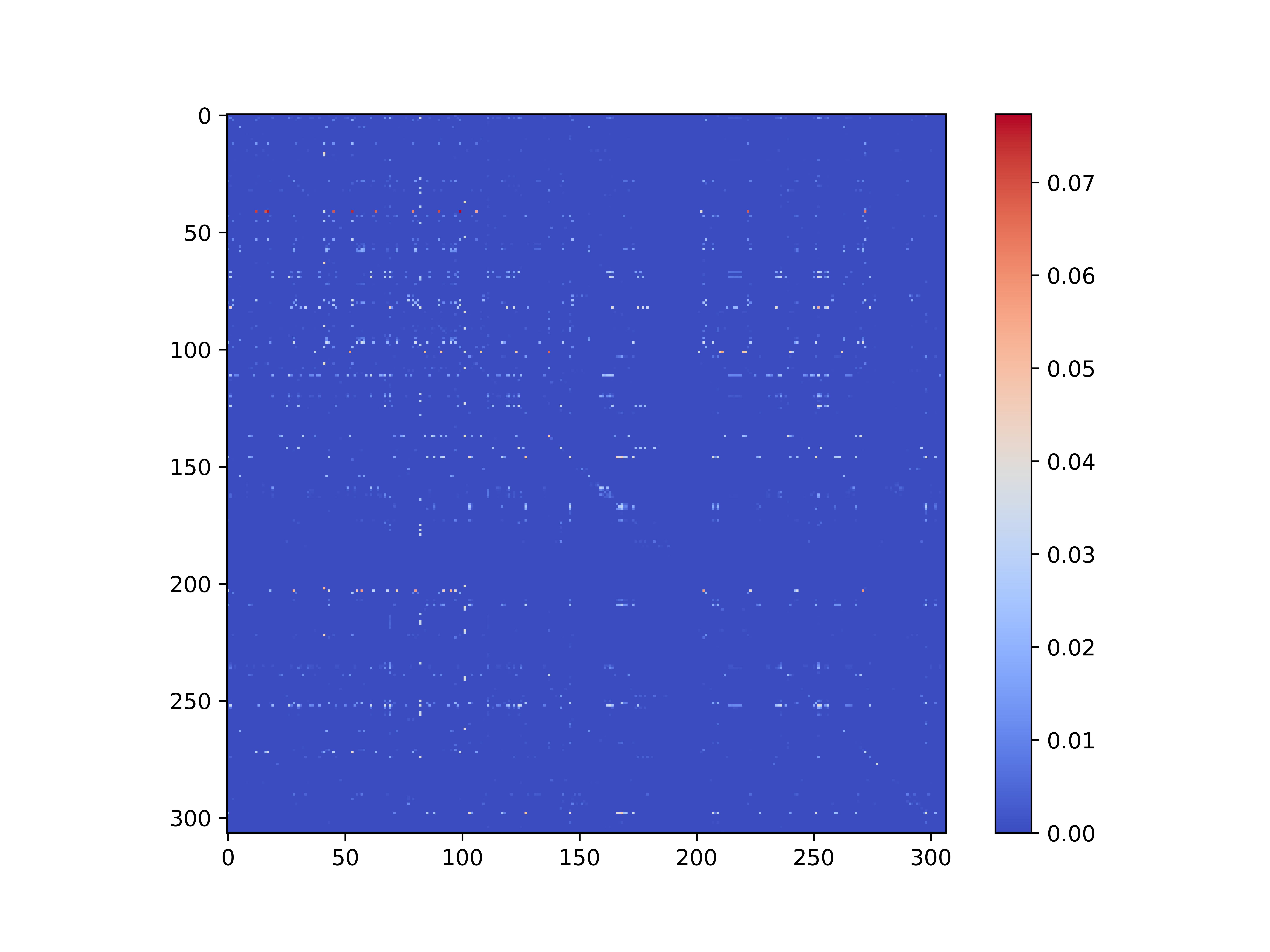}
    \label{fig:sub57}
  }
  \hfill
  \subfloat[Stage 2: H4, L2]{%
    \includegraphics[width=0.24\textwidth]{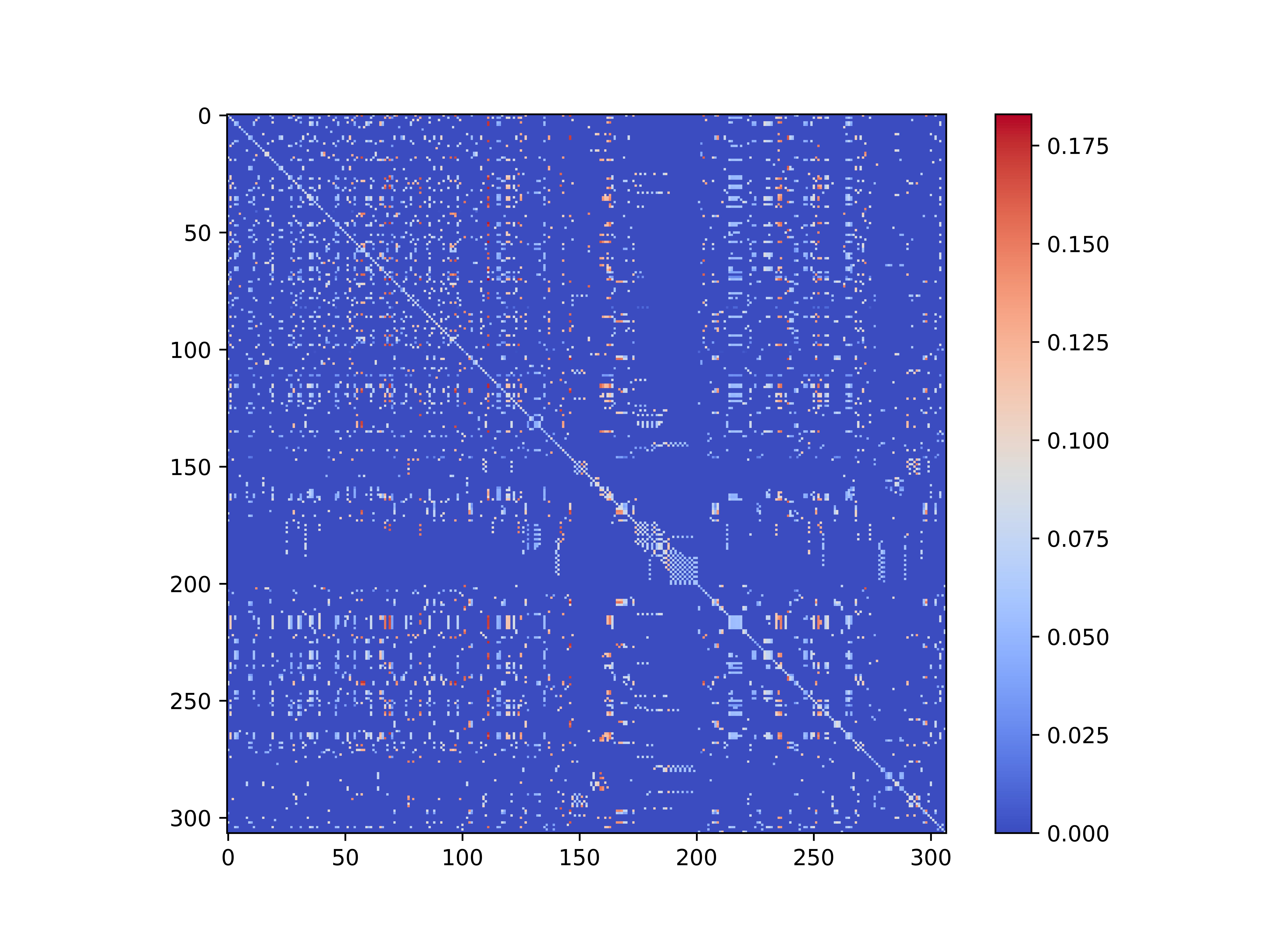}
    \label{fig:sub58}
  }
  \\[-6pt] 
  \subfloat[Stage 1: H1, L3]{%
    \includegraphics[width=0.24\textwidth]{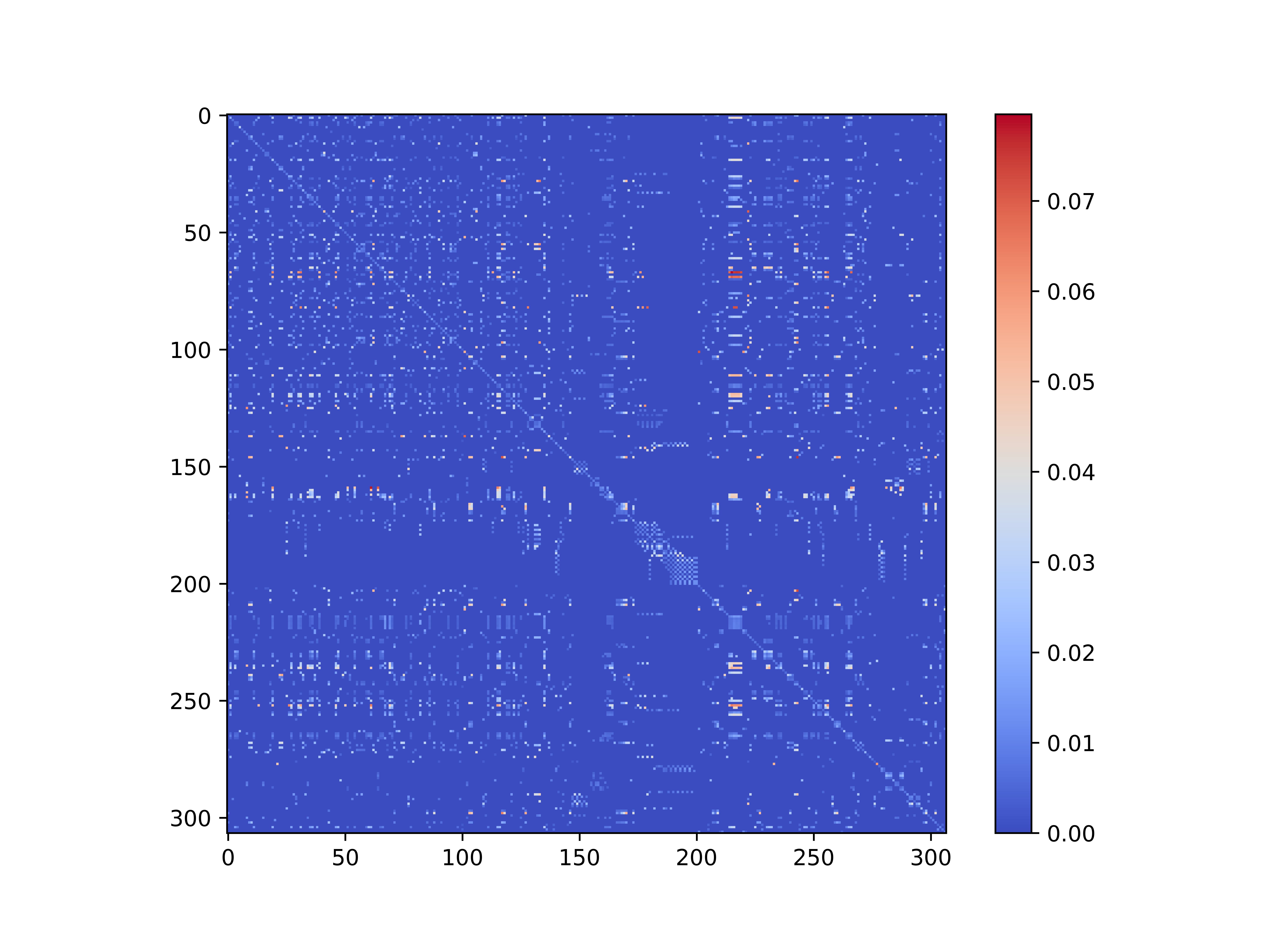}
    \label{fig:sub59}
  }
  \hfill
  \subfloat[Stage 1: H2, L3]{%
    \includegraphics[width=0.24\textwidth]{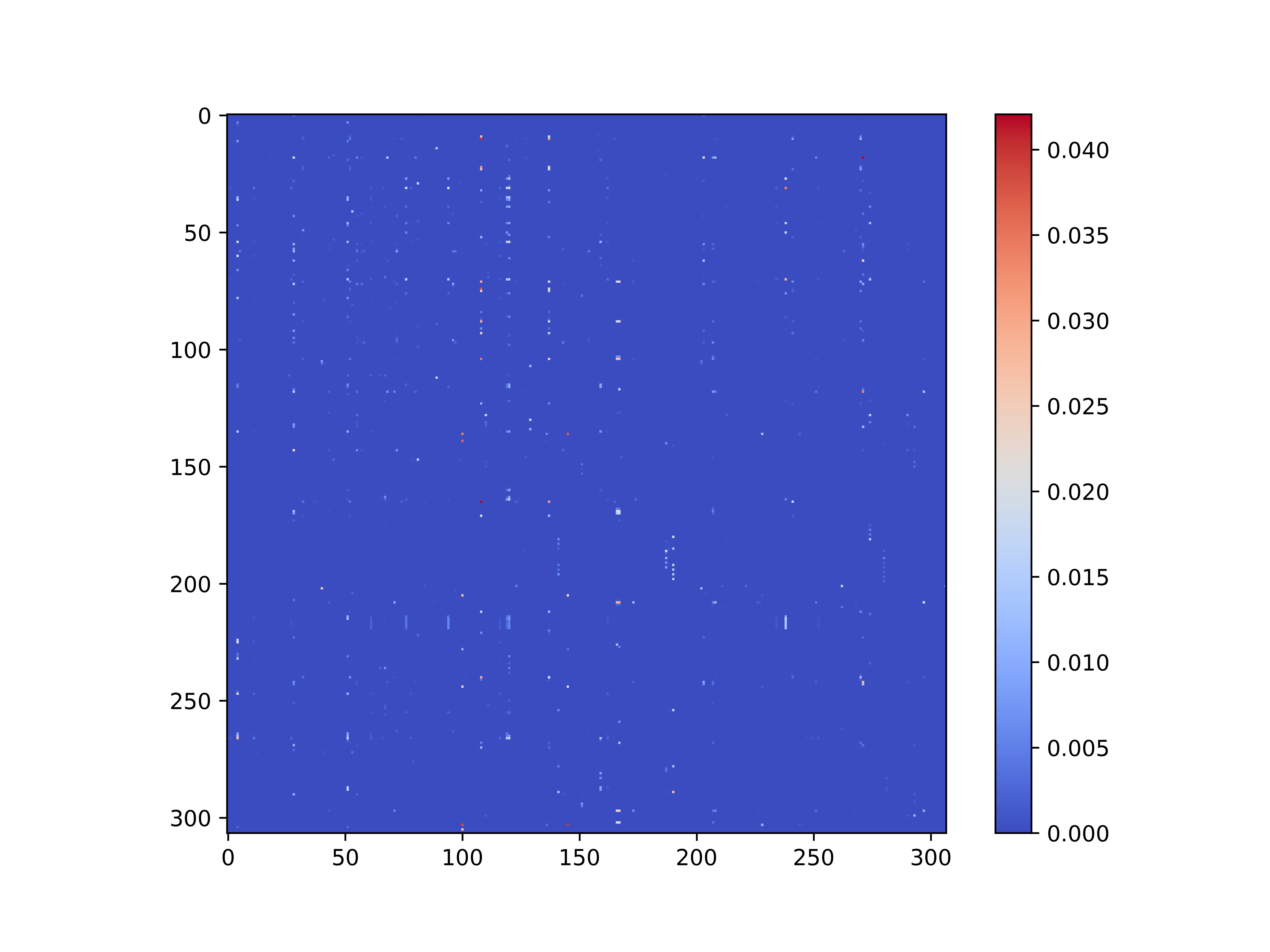}
    \label{fig:sub60}
  }
  \hfill
  \subfloat[Stage 2: H3, L3]{%
    \includegraphics[width=0.24\textwidth]{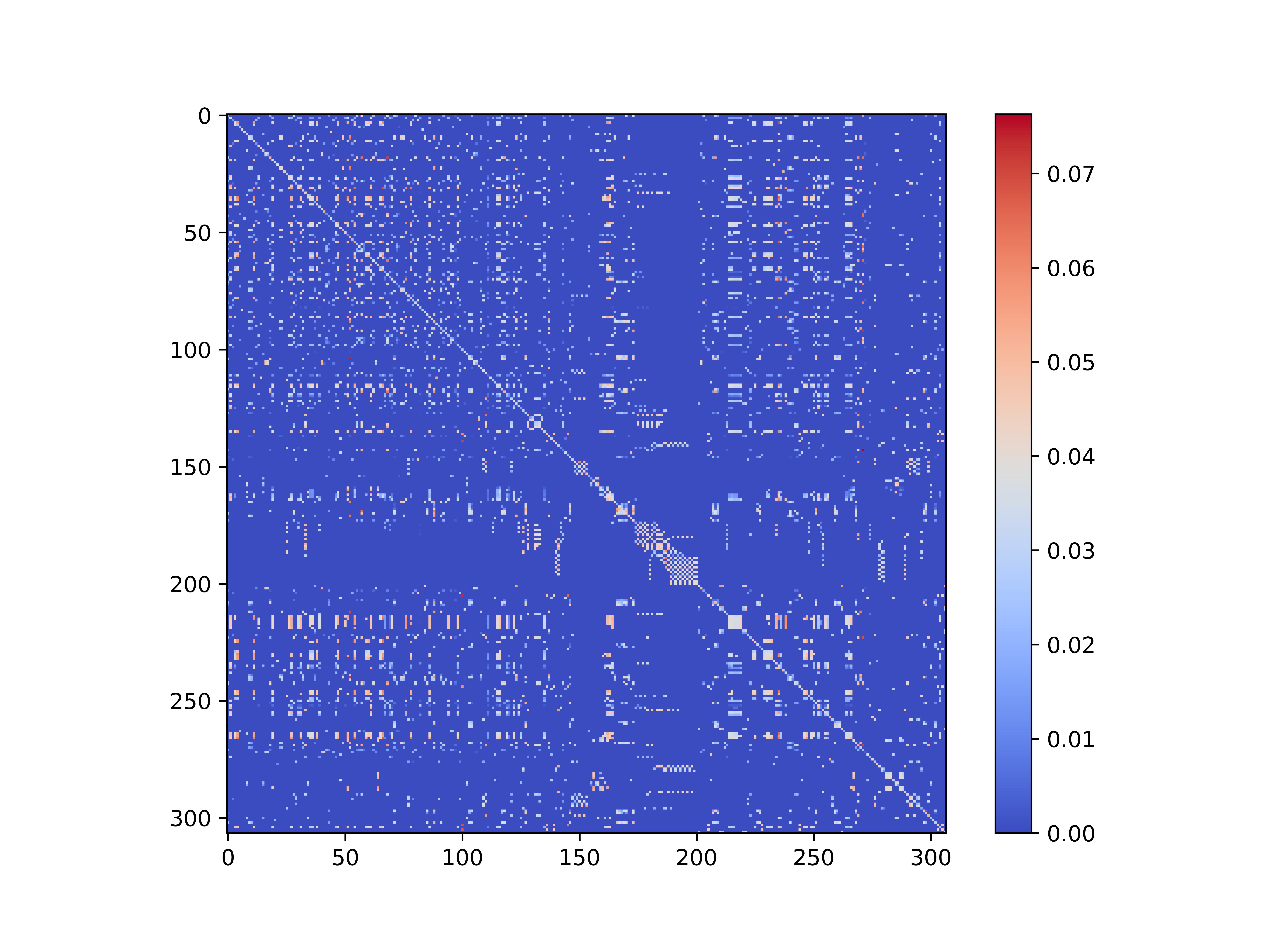}
    \label{fig:sub61}
  }
  \hfill
  \subfloat[Stage 2: H4, L3]{%
    \includegraphics[width=0.24\textwidth]{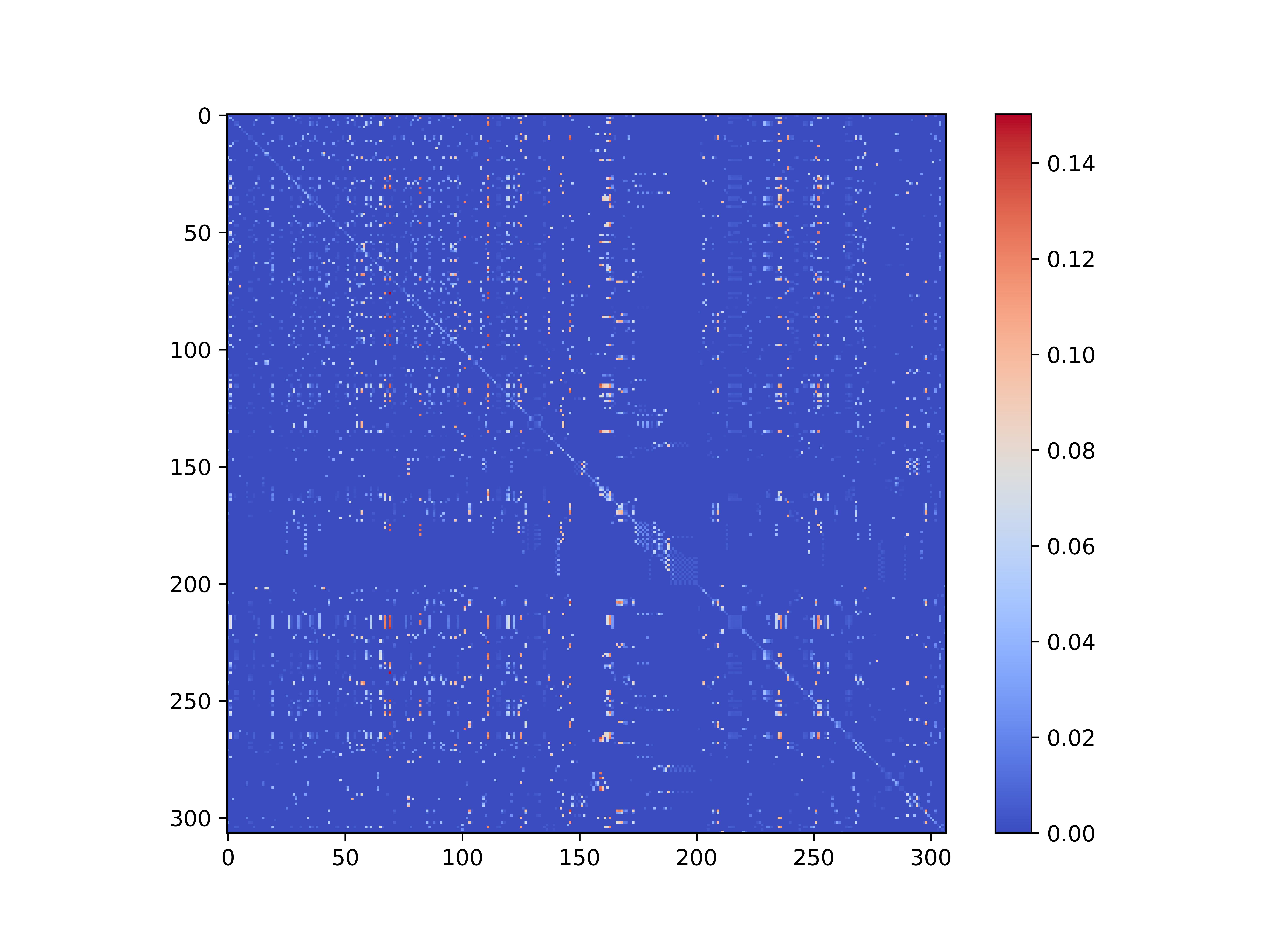}
    \label{fig:sub62}
  }
  \\[-6pt] 
  \subfloat[Stage 1: H1, L4]{%
    \includegraphics[width=0.24\textwidth]{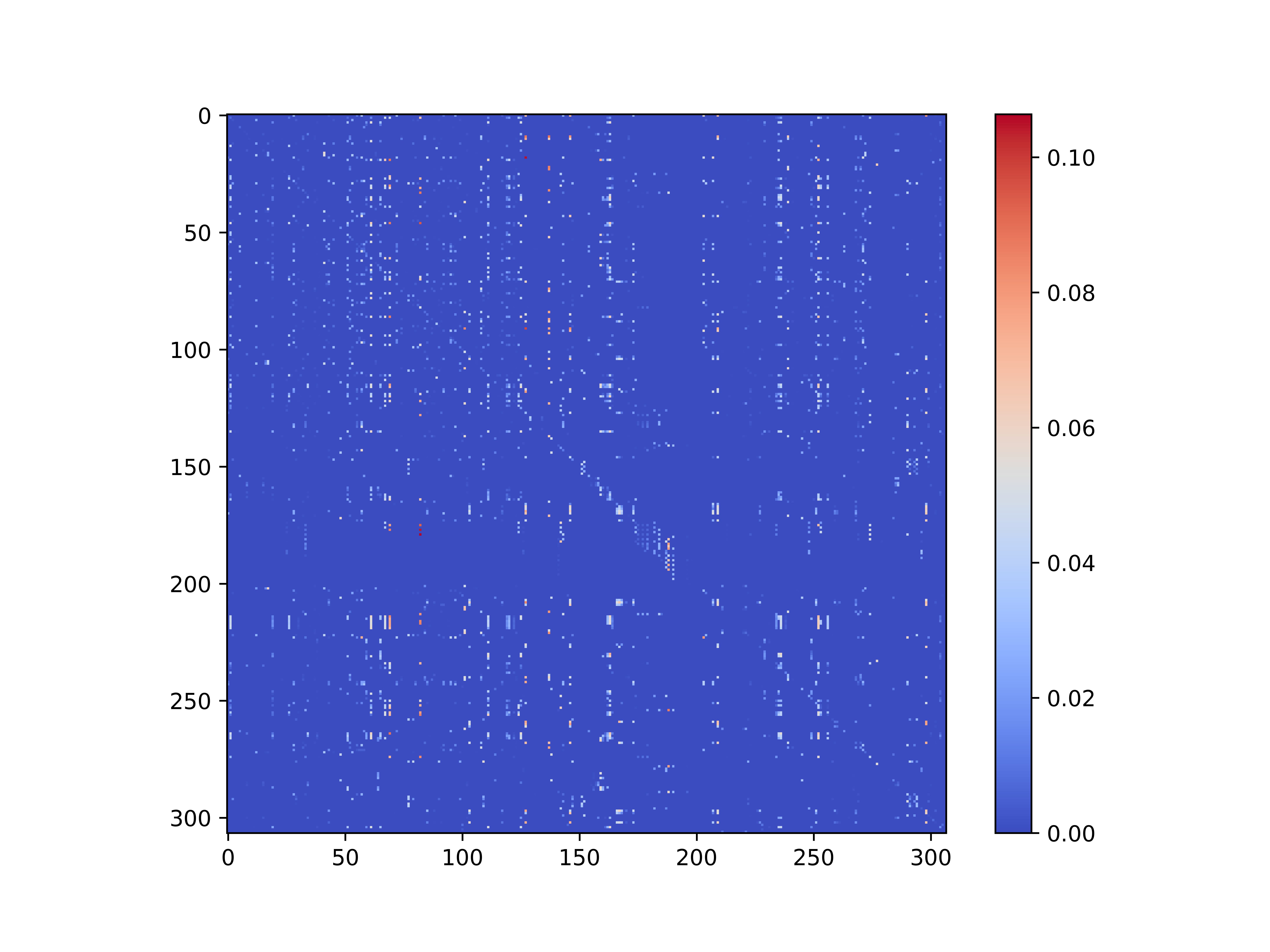}
    \label{fig:sub63}
  }
  \hfill
  \subfloat[Stage 1: H2, L4]{%
    \includegraphics[width=0.24\textwidth]{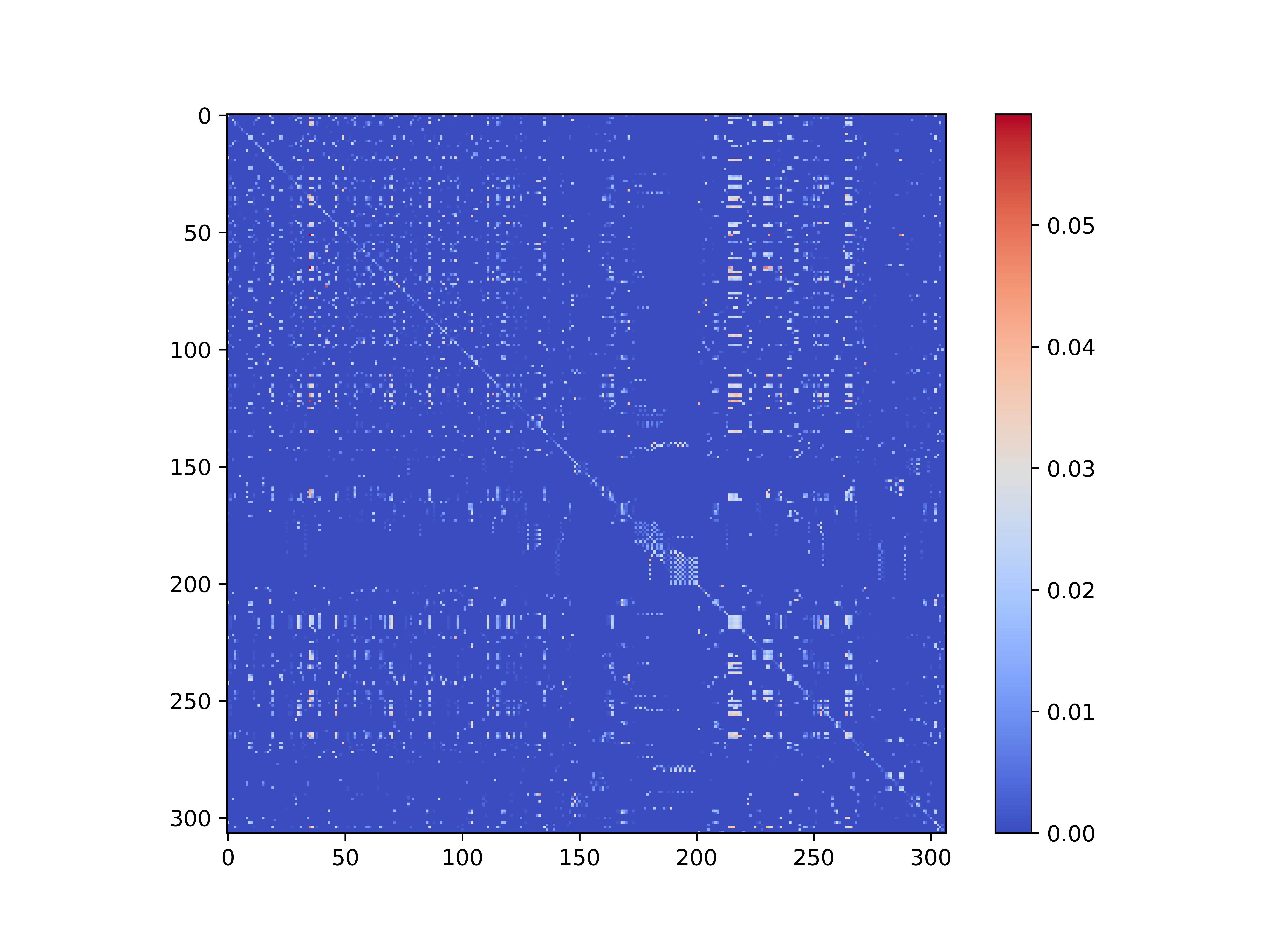}
    \label{fig:sub64}
  }
  \hfill
  \subfloat[Stage 2: H3, L4]{%
    \includegraphics[width=0.24\textwidth]{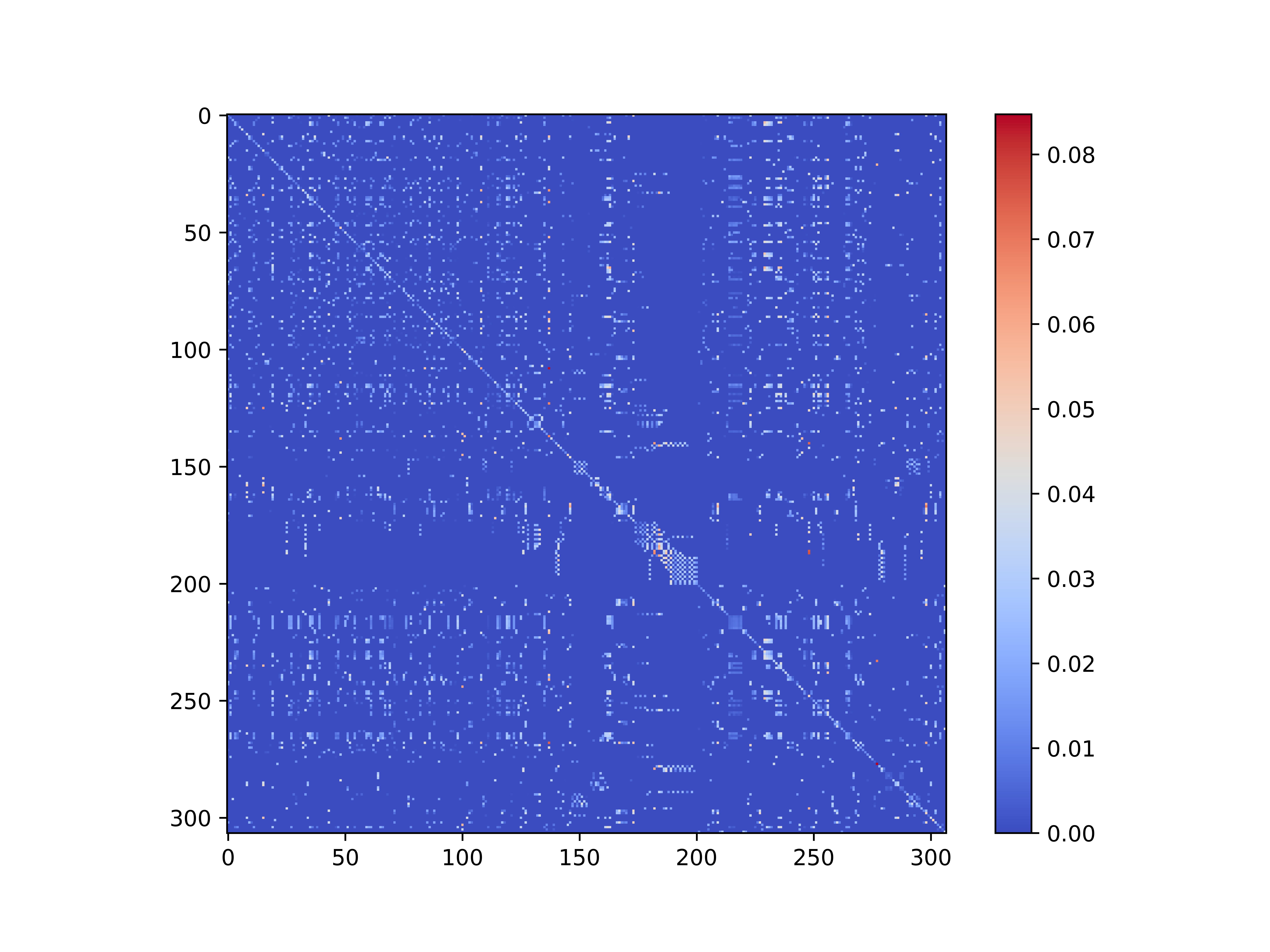}
    \label{fig:sub65}
  }
  \hfill
  \subfloat[Stage 2: H4, L4]{%
    \includegraphics[width=0.24\textwidth]{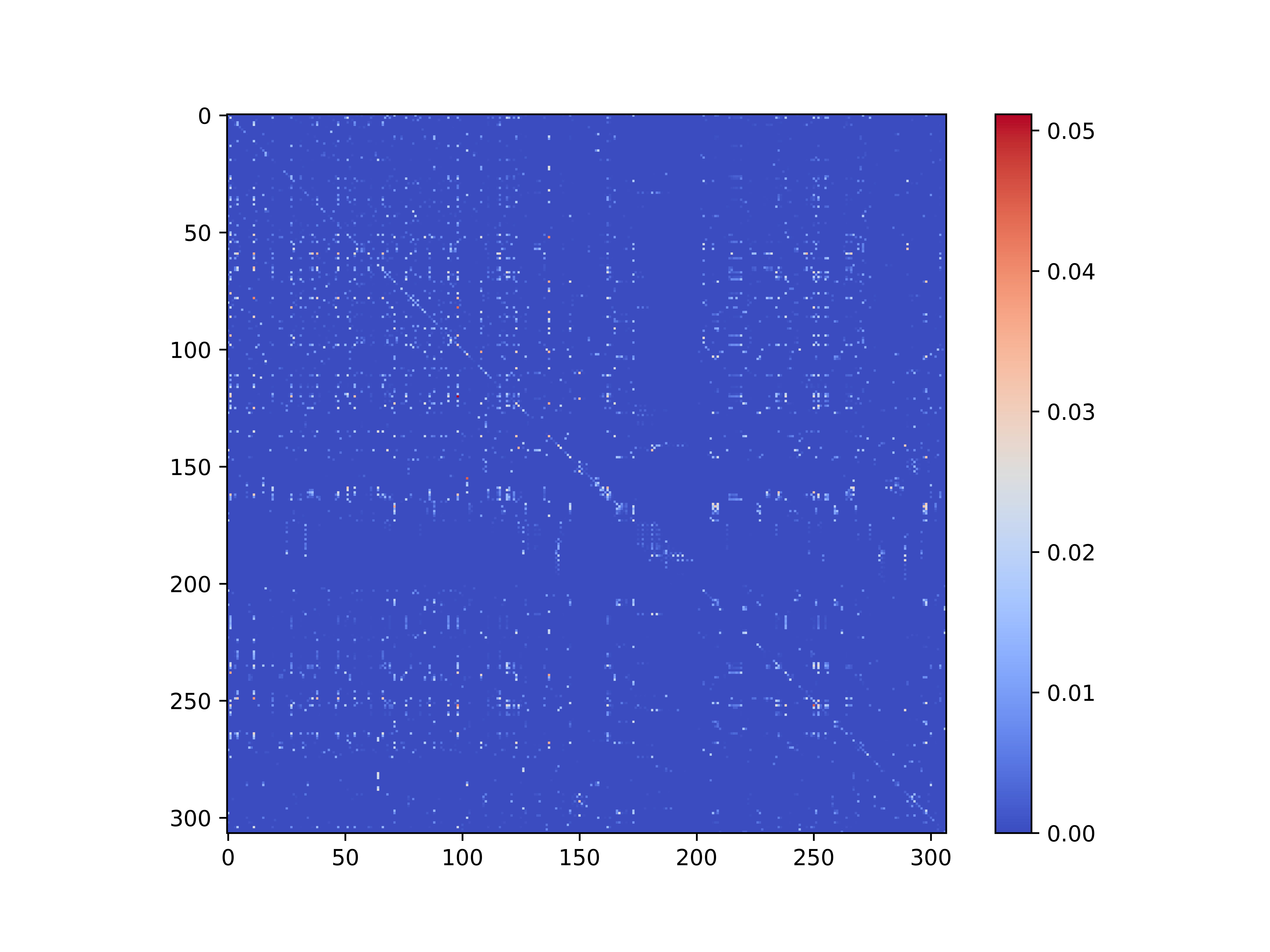}
    \label{fig:sub66}
  }
  \\[-6pt] 
  \subfloat[Stage 1: H1, L5]{%
    \includegraphics[width=0.24\textwidth]{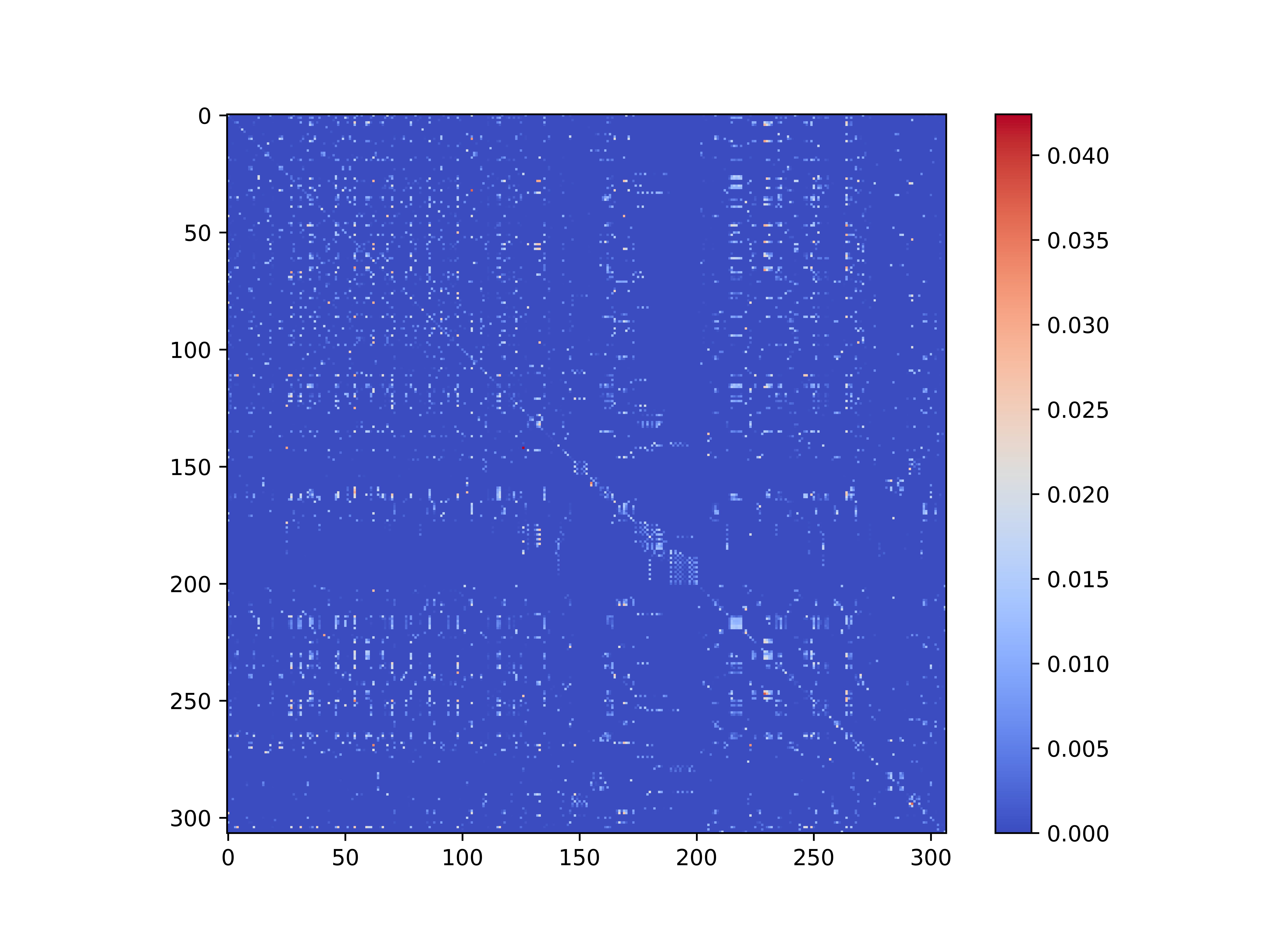}
    \label{fig:sub67}
  }
  \hfill
  \subfloat[Stage 1: H2, L5]{%
    \includegraphics[width=0.24\textwidth]{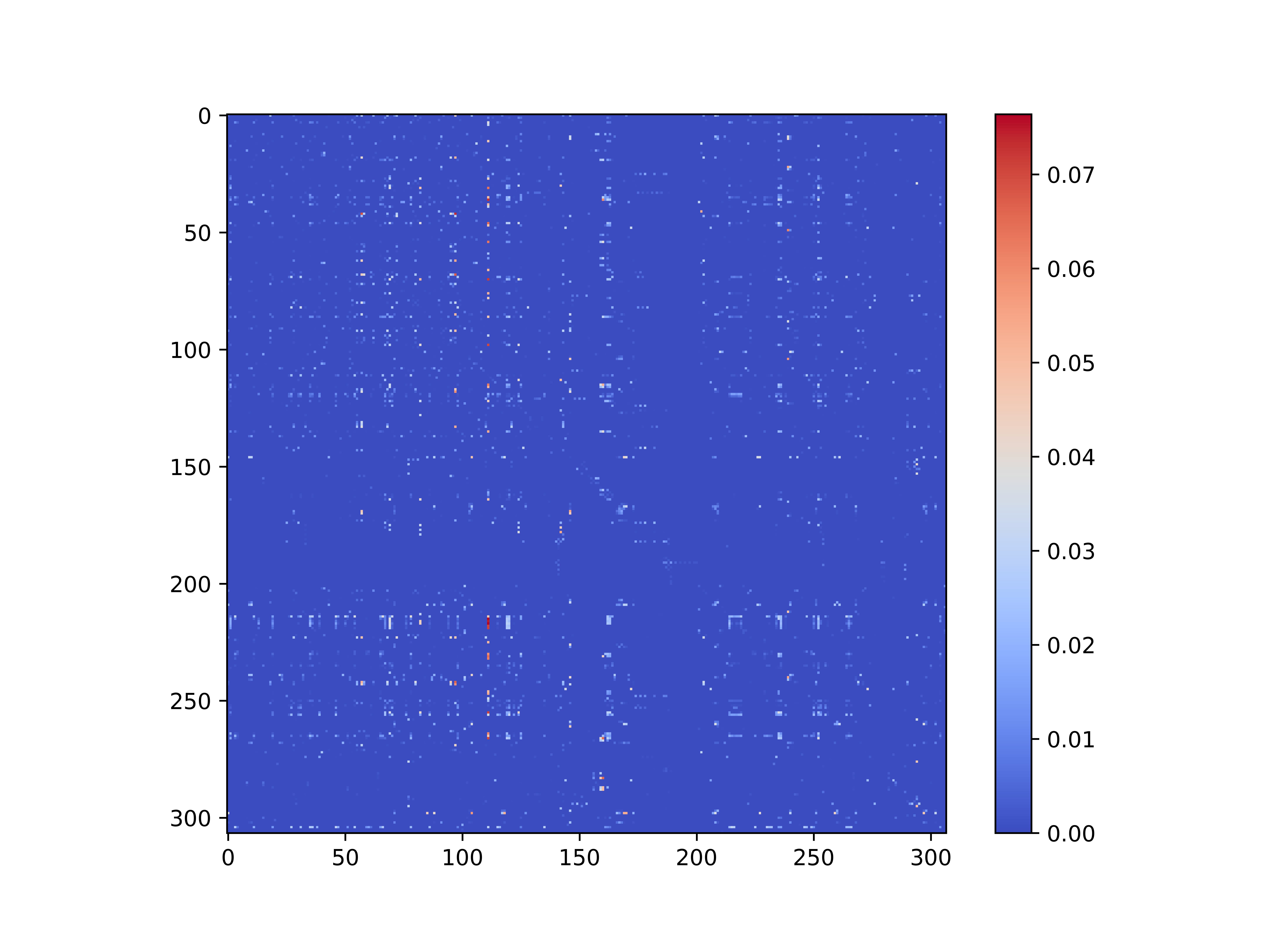}
    \label{fig:sub68}
  }
  \hfill
  \subfloat[Stage 2: H3, L5]{%
    \includegraphics[width=0.24\textwidth]{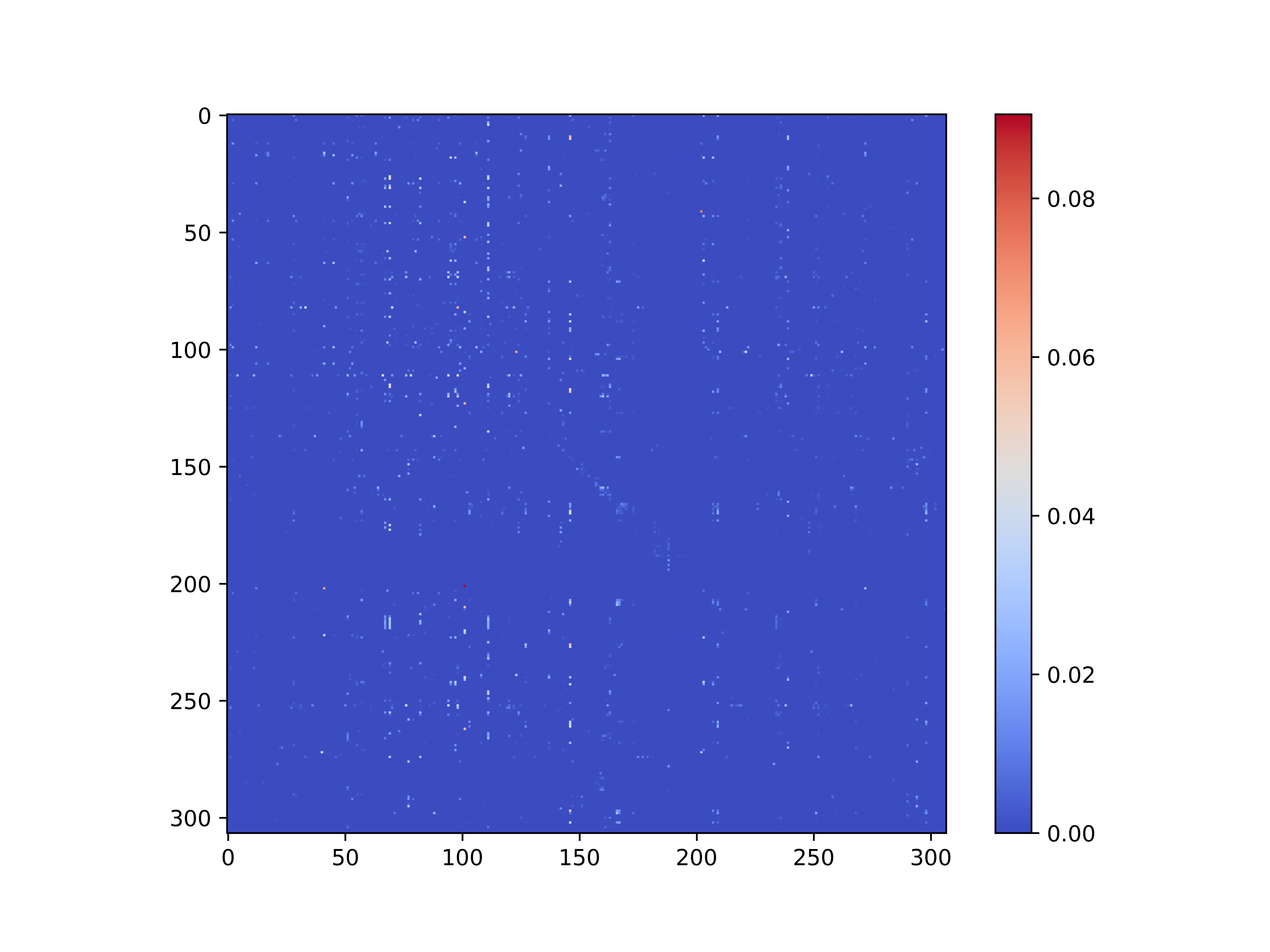}
    \label{fig:sub69}
  }
  \hfill
  \subfloat[Stage 2: H4, L5]{%
    \includegraphics[width=0.24\textwidth]{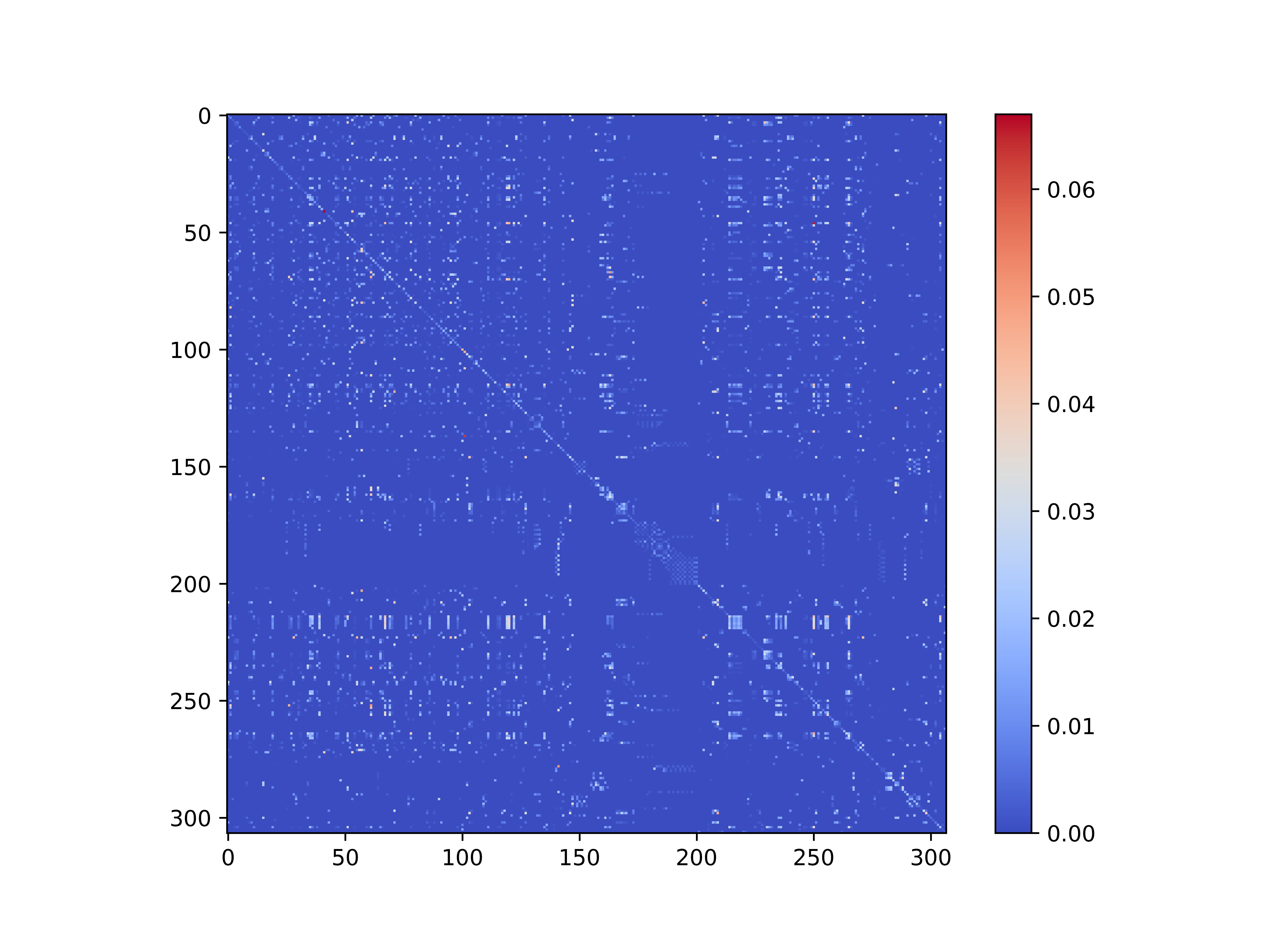}
    \label{fig:sub70}
  }
  \\[-6pt] 
  \subfloat[Stage 1: H1, L6]{%
    \includegraphics[width=0.24\textwidth]{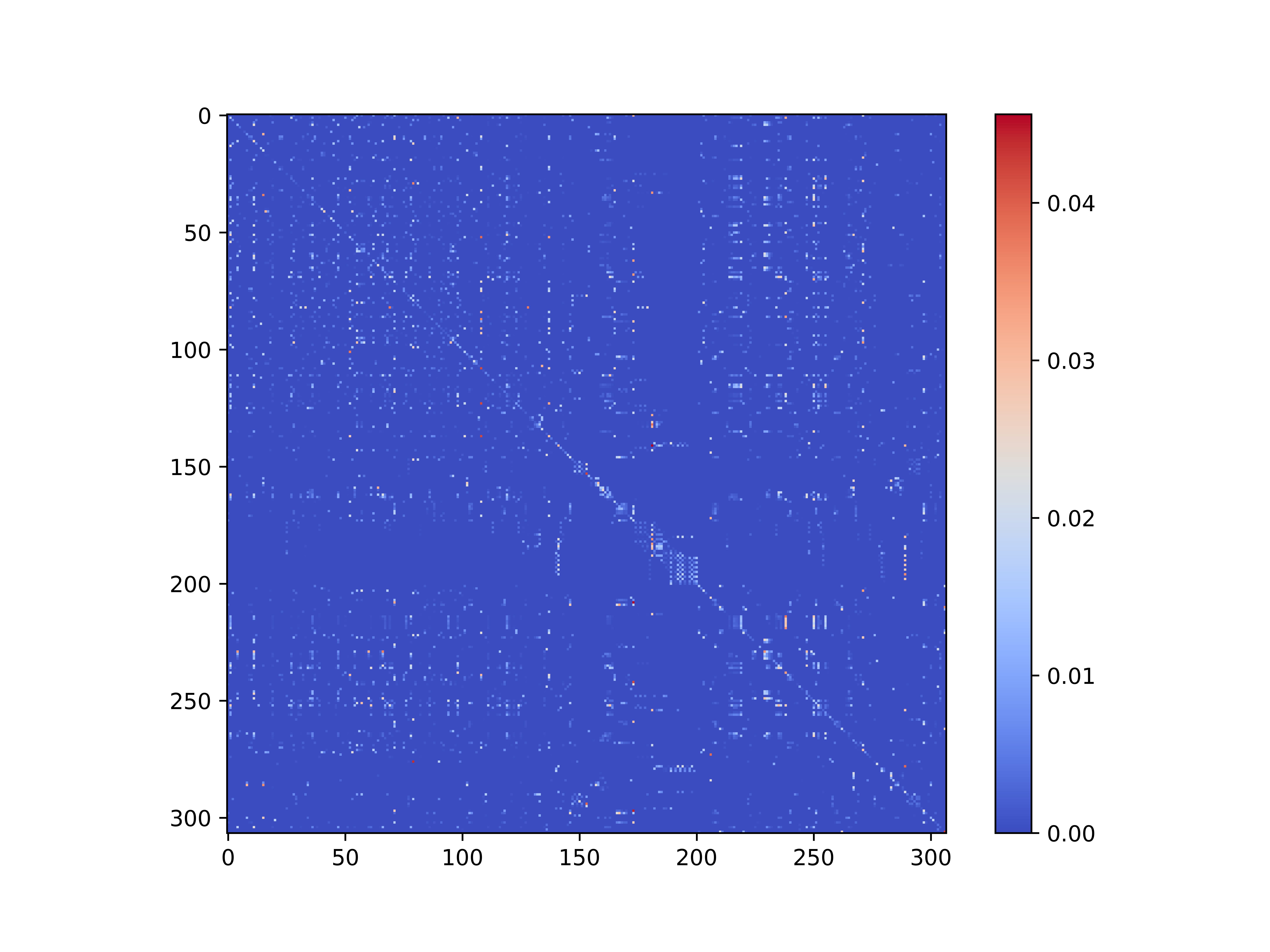}
    \label{fig:sub71}
  }
  \hfill
  \subfloat[Stage 1: H2, L6]{%
    \includegraphics[width=0.24\textwidth]{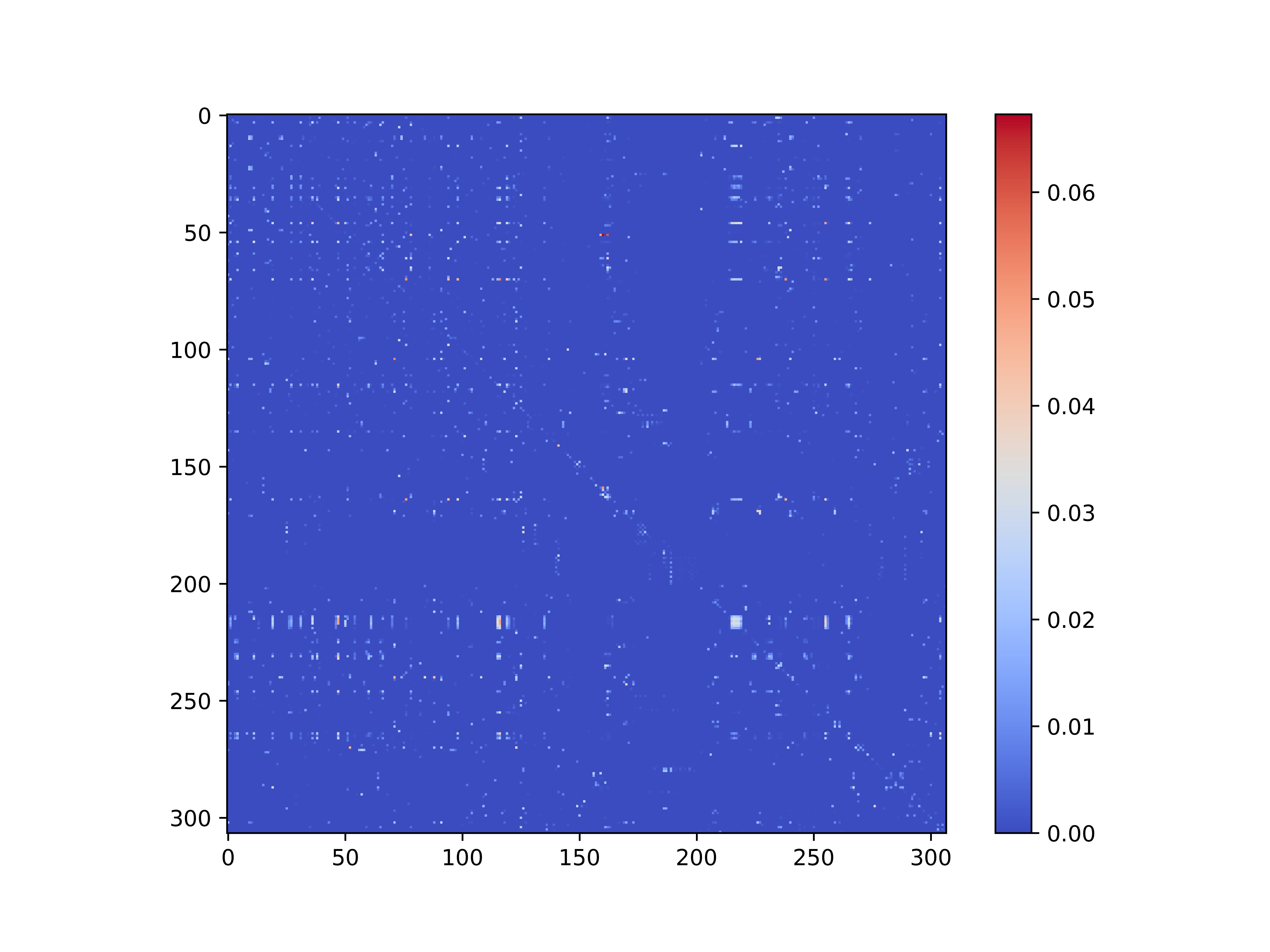}
    \label{fig:sub72}
  }
  \hfill
  \subfloat[Stage 2: H3, L6]{%
    \includegraphics[width=0.24\textwidth]{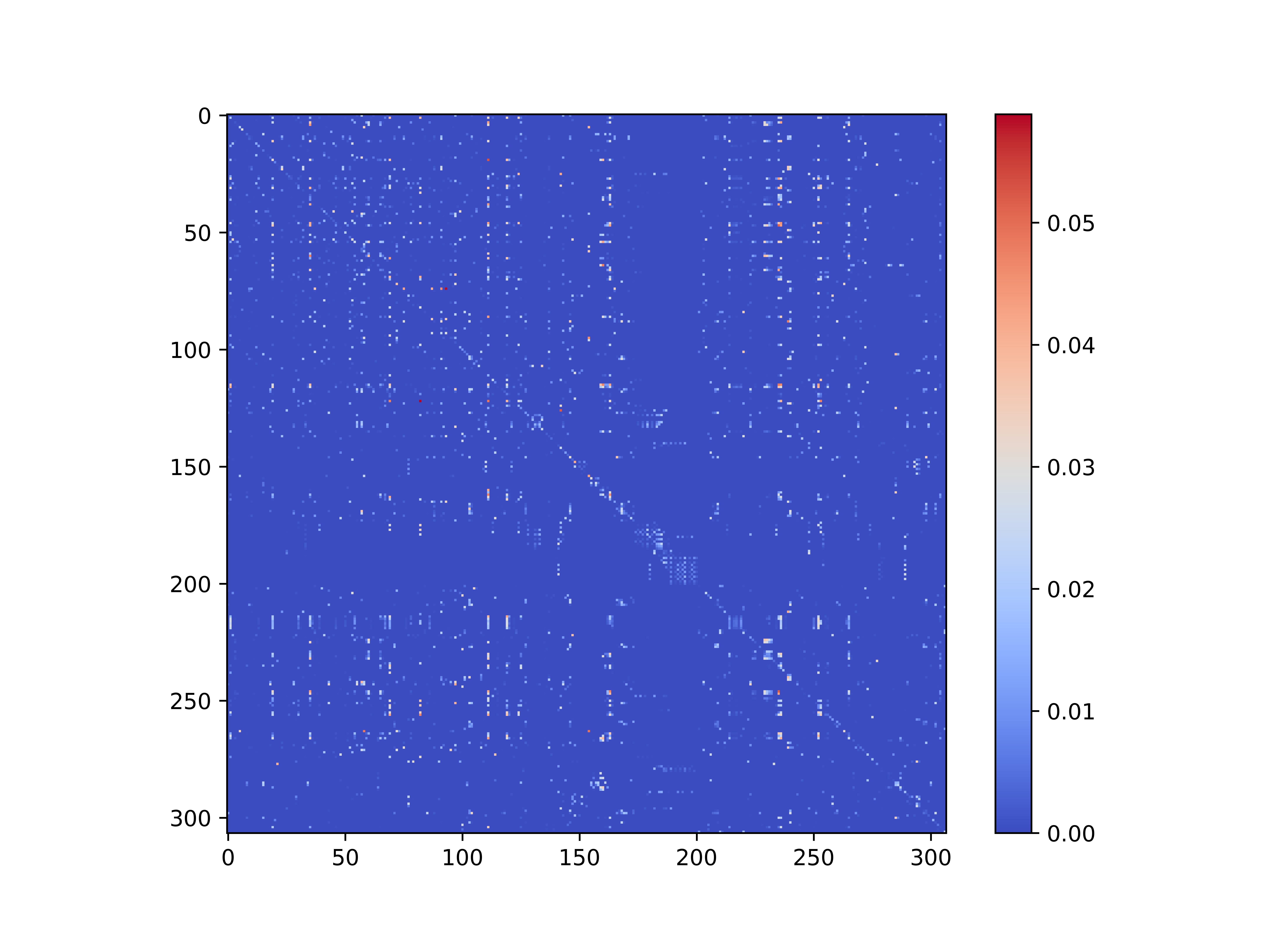}
    \label{fig:sub73}
  }
  \hfill
    \subfloat[Stage 2: H4, L6]{%
    \includegraphics[width=0.24\textwidth]{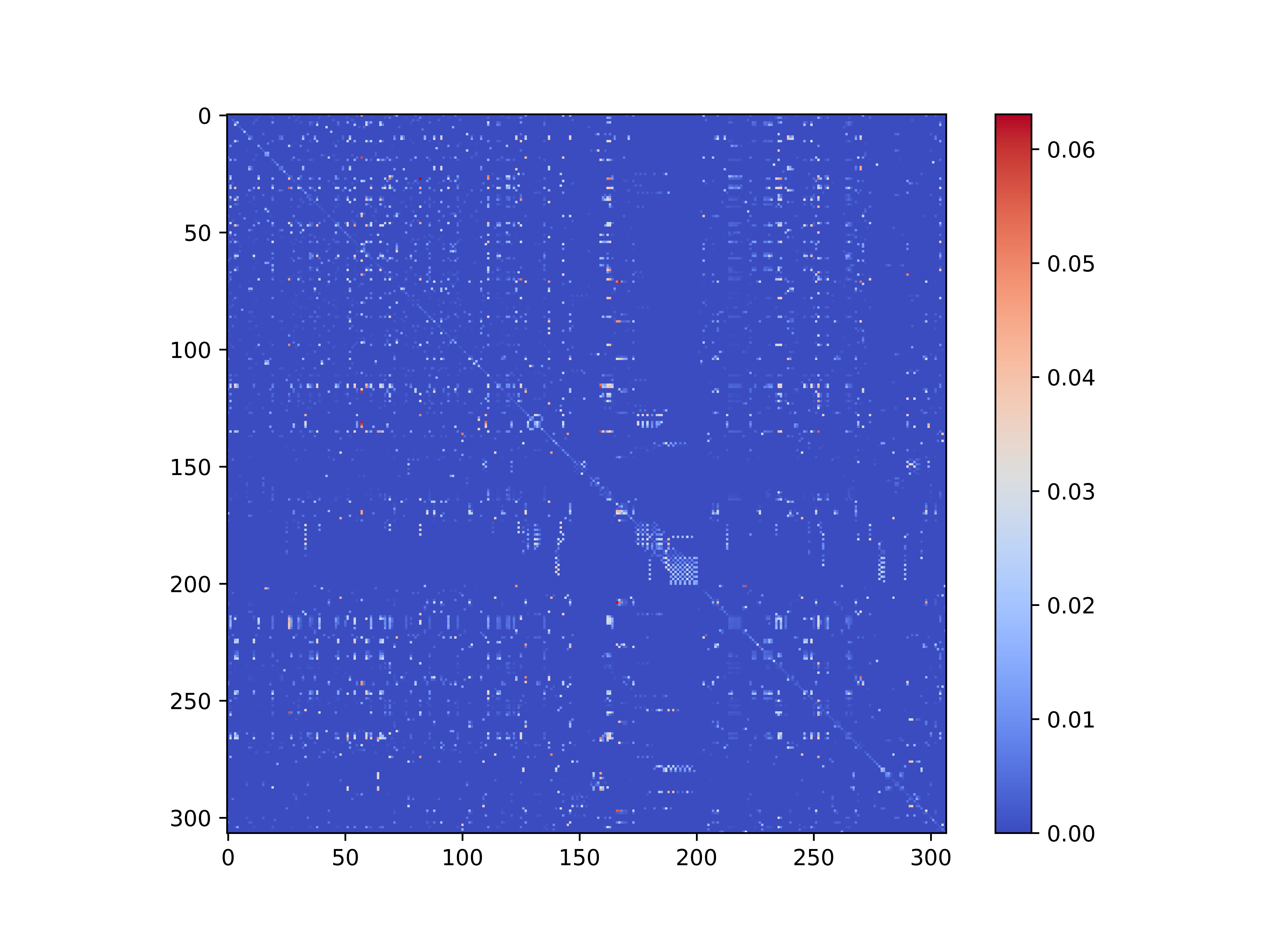}
    \label{fig:sub74}
  }
\end{figure*}

To gain a deeper understanding of the functionalities of \textbf{CCDSReFormer}, we further explore how attention mechanisms are applied to different types of information - spatial, temporal, integration of both and delay-aware attention. The structure of each attention module is divided into two distinct stages which include \textbf{ReSSA} and \textbf{ReTSA} as depicted in Figure~\ref{fig:reformer}. One given example for the spatial information \textbf{ReSSA} on the PeSM04 dataset is shown in Figure~\ref{fig:attentionmap_ressa}. The first stage is characterized by the implementation of \textbf{ReSSA}, while the second stage involves criss-cross learning or what we call integration of temporal information using \textbf{ReTSA}. For \textbf{ReDASA}, we generate visualizations that directly correspond to the increase in heads and layers, aligning closely with the experimental settings described in Section~\ref{Experim_set}.
By examining the attention score visualizations, as displayed in  Figures~\ref{fig:attentionmap_ressa},~\ref{fig:attentionmap_retsa} and \ref{fig:attentionmap_redasa}, we can draw several insightful conclusions as below:

\paragraph{Rectified self-attention with increasing of stages:}
The self-attention mechanism, when applied in increasing stages (or Criss-Cross learning), evenly distributes attention across integrated spatial and temporal information. This phenomenon is clearly visible when contrasting Figure~\ref{fig:sub1} with Figure~\ref{fig:sub4}. Initially, in \textbf{ReSSA} (Figure~\ref{fig:sub1}), the heatmap displays attention scores from 0.005 to 0.025, predominantly accentuating spatial details with elevated attention scores. However, upon integrating \textbf{ReTSA}, the heatmap in Figure~\ref{fig:sub4} exhibits a more evenly distribution of attention, with scores approximating 0.2, including both spatial and temporal aspects. 

A similar trend is observable in the initial \textbf{ReTSA} attention scores (Figure~\ref{fig:sub27}), where an increase in stage (as compared with Figure~\ref{fig:sub29}) leads to an average trend in attention scores. Additional examples are provided in Figures~\ref{fig:sub31}, \ref{fig:sub33}, \ref{fig:sub35}, and \ref{fig:sub37}, among others. Such observations suggest that the integration of the Criss-Cross methodology equips the \textbf{CCDSReFormer} with an enhanced capability to capture and delineate both spatial and temporal information with refined attention.

\paragraph{Rectified self-attention in different number of heads:}
As the number of attention heads increases from 1 to 2, each head can potentially focus on distinct segments or attributes of the input data. This concept is illustrated through a comparison on \textbf{ReSSA} (as in Figure~\ref{fig:attentionmap_ressa}) between Figure~\ref{fig:sub1} and Figure~\ref{fig:sub2}. Notably, Figure~\ref{fig:sub2} displays higher attention scores compared to Figure~\ref{fig:sub1}, indicating a more comprehensive capture of data features and relationships. 

Furthermore, in \textbf{ReTSA}, as illustrated in Figure~\ref{fig:attentionmap_retsa}, an increase in the number of heads during stage 1 (as shown in the first two columns, e.g., Figures~\ref{fig:sub27} and \ref{fig:sub28}) results in significant changes in attention scores compared to stage 2 (as depicted in the last two columns, e.g., Figures~\ref{fig:sub29} and \ref{fig:sub30}). 

Similarly, in \textbf{ReDASA} (see Figure~\ref{fig:attentionmap_redasa}), setting the number of attention heads to 4 reveals that, with an increasing number of heads, the attention scores in \textbf{ReDASA} closely resemble those when the head count is one. This can be specifically observed in comparisons such as in Figures~\ref{fig:sub51} and \ref{fig:sub54}; however, there are notable exceptions with significant differences, as seen when comparing Figures~\ref{fig:sub55} and \ref{fig:sub58}. This diversity in attention distribution enhances our model's ability to analyze input data more effectively, allowing for the recognition of a wider range of patterns and connections.

\paragraph{Rectified self-attention with increasing of layers:}
As layers increase, attention scores are dynamically adjusted. This is evident in \textbf{ReSSA} (as in Figure~\ref{fig:attentionmap_ressa}) stage 1 and head 1, where the scores slightly rise from a ceiling of 0.04 to 0.05 across layers, as shown from Figure~\ref{fig:sub1} to Figure~\ref{fig:sub21}. The change becomes slighter at higher layers, such as layer 5 (Figure~\ref{fig:sub17}) and layer 6 (Figure~\ref{fig:sub21}). Similarly, in stage 2 and head 2, the maximum attention score rises from 0.5 to 0.8, as illustrated in Figure~\ref{fig:sub4} and Figure~\ref{fig:sub26} which means even with Criss-Cross learning, with the increasing of layers can further enhance the score of attention distribution. 

Additionally, with \textbf{ReTSA} (referenced in Figure~\ref{fig:attentionmap_retsa}), a progression is evident when observing the visualization from the top row (Figures~\ref{fig:sub51}, \ref{fig:sub52}, \ref{fig:sub53}, and \ref{fig:sub54}) to the bottom row (Figures~\ref{fig:sub71}, \ref{fig:sub72}, \ref{fig:sub73}, and \ref{fig:sub74}). As we examine the subsequent layers, from Layer 1 through Layer 6, there is a noticeable trend towards a more uniform distribution of attention. This suggests that as the layers deepen, the model may be gaining a more nuanced perception of the data's interrelations.

\section{Limitation}
The current limitation is that the computational results do not surpass all the baselines, primarily due to the significant computational load of the entire model framework. However, we have demonstrated that even in models with substantial computational requirements, our method can effectively reduce computational costs. Any attention mechanism can utilize the \textbf{EnReLSA} attention to decrease computational expenses.

\section{Conclusion}
\label{Conclusion}
In this study, we introduce the Dual-Stream Criss-Cross Enhanced Rectified Transformer (\textbf{CCDSReFormer}), an innovative model designed for accurate and computationally efficient traffic flow prediction. The model uniquely combines spatial and temporal information through a dual Criss-Cross stream, effectively capturing the intricate dynamics of traffic patterns. A locally enhanced convolution within the attention mechanism is implemented, sharpening the model's focus on local spatial-temporal features and nuanced traffic dynamics influenced by localized conditions. This ensures a nuanced exploration of traffic patterns influenced by localized conditions, contributing to a more accurate prediction. Additionally, we incorporate Rectified Linear Attention (ReLA) into the traffic forecasting domain, which adapts dynamically to the specific spatial-temporal characteristics of traffic data while simultaneously reducing computational demands. This novel approach introduces dynamism and adaptability, responding effectively to the unique spatial-temporal characteristics. Comparative analysis of various \textbf{CCDSReFormer} configurations demonstrates each component's positive impact on prediction accuracy. Our method has undergone extensive testing across six diverse real-world datasets, establishing its superiority over existing state-of-the-art models in terms of both performance and computational efficiency. Additionally, our model exhibits robust parameter tuning capabilities, further emphasizing its versatility and applicability in the dynamic domain of traffic flow prediction. Looking ahead, we plan to explore the application of the \textbf{CCDSReFormer} model to other spatial-temporal datasets, such as weather forecasting.
\bibliographystyle{IEEEtran}
\bibliography{ref}

\end{document}